\documentclass[10pt,twocolumn,letterpaper]{article}

\usepackage{cvpr}              

\usepackage{graphicx}
\usepackage{amsmath}
\usepackage{amssymb}
\usepackage{booktabs}

\usepackage{multirow}
\usepackage{color}
\usepackage{colortbl}
\usepackage{xcolor}
\usepackage{lineno}
\usepackage[thicklines]{cancel}

\definecolor{forestgreen}{RGB}{47, 159, 87}

\usepackage[pagebackref,breaklinks,colorlinks]{hyperref}

\usepackage[capitalize]{cleveref}
\crefname{section}{Sec.}{Secs.}
\Crefname{section}{Section}{Sections}
\Crefname{table}{Table}{Tables}
\crefname{table}{Tab.}{Tabs.}

\def\cvprPaperID{2843} 
\def\confName{CVPR}
\def\confYear{2023}

\begin{document}

\title{Progressive Learning without Forgetting}
\author{Tao Feng$^1$, \quad Hangjie Yuan$^{2}$, \quad Mang Wang$^3$, \quad Ziyuan Huang$^4$, \quad Ang Bian$^1$, \quad Jianzhou Zhang$^1$\\
$^1$Sichuan University \quad $^2$Zhejiang University \quad $^3$ByteDance Inc. \quad $^4$National University of Singapore\\
{\tt\small fengtao.hi@gmail.com, hj.yuan@zju.edu.cn, wangmang@bytedance.com}\\
{\tt\small ziyuan.huang@u.nus.edu, bian@scu.edu.cn,  zhangjz@scu.edu.cn}
}

\maketitle

\begin{abstract}
Learning from changing tasks and sequential experience without forgetting the obtained knowledge is a challenging problem for artificial neural networks. 
In this work, we focus on two challenging problems in the paradigm of Continual Learning (CL) without involving any old data: \textbf{(i)} the accumulation of catastrophic forgetting caused by the gradually fading knowledge space from which the model learns the previous knowledge; \textbf{(ii)} the uncontrolled tug-of-war dynamics to balance the stability and plasticity during the learning of new tasks. In order to tackle these problems, we present \textbf{Progressive Learning without Forgetting} (PLwF) and a credit assignment regime in the optimizer. PLwF densely introduces model functions from previous tasks to construct a knowledge space such that it contains the most reliable knowledge on each task and the distribution information of different tasks, while credit assignment controls the tug-of-war dynamics by removing gradient conflict through projection.
Extensive ablative experiments demonstrate the effectiveness of PLwF and credit assignment. In comparison with other CL methods, we report notably better results even without relying on any raw data.
\end{abstract}

\vspace{-3mm}

\section{Introduction}
\label{sec:intro}

Continual Learning (CL) remains a long-standing challenge for Artificial Neural Networks (ANNs) \cite{hadsell2020embracing,cichon2015branch,DBLP:journals/corr/abs-2209-01814}. During CL, the model is prone to suffer from catastrophic forgetting, where the deep learner only performs well on the most recent tasks while no longer recalls the knowledge learned in earlier tasks.
A naive strategy for avoiding catastrophic forgetting would be training a new model on data for all the existing tasks.
However, this is impractical due to data inaccessibility. Hence, a line of work \cite{guo2020improved,lopez2017gradient,chaudhry2018efficient} focuses on constructing special subspace of old tasks instead of adopting data to mitigate forgetting.

\begin{figure}
\centering
\includegraphics[width=0.47\textwidth]{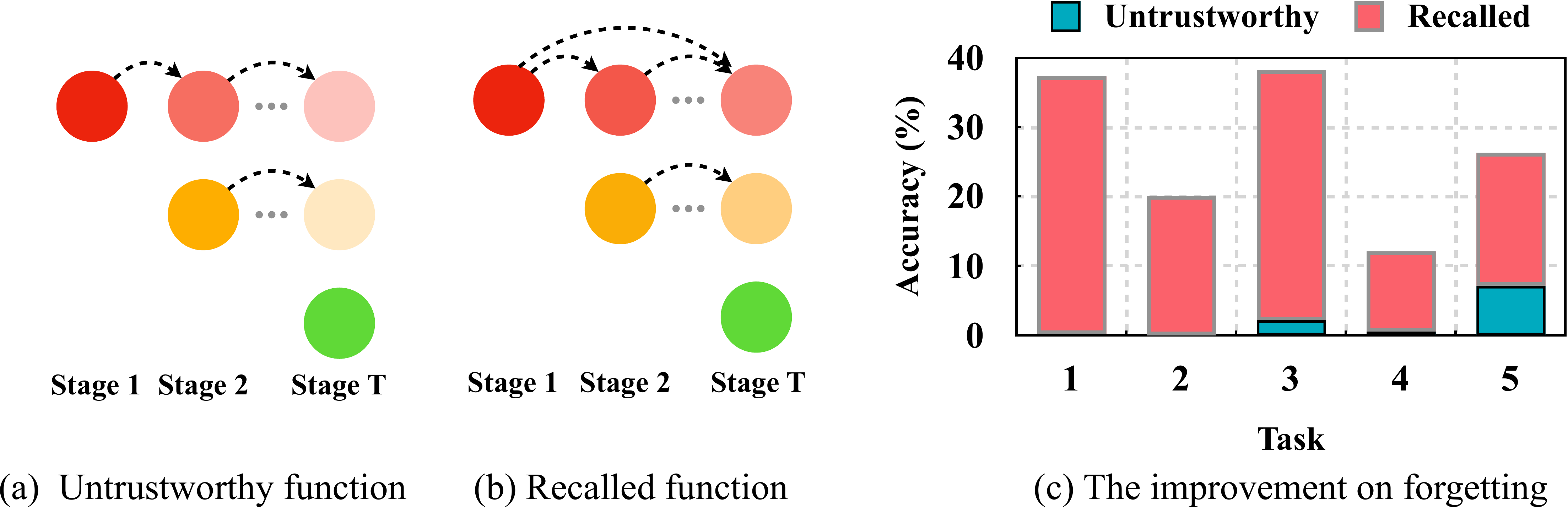}
\caption{\textbf{(a)} The illustration of the accumulation of forgotten knowledge caused by gradually fading knowledge space when learning in a new stage. 
Dotted lines denote the use of previous knowledge space in new tasks. The degree of opacity denotes the reliability of knowledge in current tasks.
\textbf{(c)} The figure indicating results for the first 5 tasks of our PLwF compared to a vanilla method (LwF~\cite{DBLP:journals/pami/LiH18a}).}
\label{figure_accumulation}
\vspace{-7mm}
\end{figure}

Another popular solution~\cite{aljundi2018memory,kirkpatrick2017overcoming,zenke2017continual,DBLP:conf/cvpr/FengWY22} is to impose regularization on the deep learner in the battle against forgetting.
Under this paradigm, there exists two key problems. First, although the passing of knowledge of previous tasks on to the model during the learning of a new task through regularization~\cite{DBLP:journals/pami/LiH18a} can reduce the amount of knowledge that is lost, it can hardly preserve all the knowledge of the old tasks. Hence, during CL over a long sequence of tasks, the model becomes less and less certain of its knowledge over early tasks (as in Figure~\ref{figure_accumulation}c, Teal column), \textit{i.e.,} the reliability of a certain model as the knowledge container of early tasks gradually dwindles~\cite{buzzega2020dark}. This means the amount of forgotten knowledge is accumulating (as in Figure~\ref{figure_accumulation}a), which we term as the \textit{accumulation of catastrophic forgetting}. Second, when learning a new task, the model is required to balance the learning of the coming task and the fight against forgetting. From the perceptive of gradient-based optimization~\cite{goodfellow2016deep,lecun2015deep}, this creates a \textit{tug-of-war game}~\cite{hadsell2020embracing}, where the gradients for achieving two objectives conflict and impede the learning of both aspects. In this work, we focus on mitigating these two problems.

In a sense, overcoming forgetting by regularization can be understood as the process of function matching~\cite{DBLP:journals/corr/abs-2106-05237} between the current model function and the previous model function(s) that contain knowledge on the old tasks~\cite{DBLP:journals/pami/LiH18a}. Hence, the essential reason for the accumulation of catastrophic forgetting is the fading reliability of the learnable knowledge space constructed by the previous model function~\cite{buzzega2020dark}. In light of this, we propose to densely exploit the model functions produced in previous tasks to construct the knowledge space.
Since this space is based on an ample number of past experiences~\cite{benna2016computational,fusi2005cascade}, the learning process is more progressive and hence we dub our method Progressive Learning without Forgetting (PLwF).
This has two merits: \textbf{(i)} the knowledge space now contains the most precise and fresh knowledge over all the previous tasks, which means the reliability of the knowledge space over the old tasks will not fade with increasingly more new tasks; \textbf{(ii)} the full distribution over labels of different tasks can now be inferred from the knowledge space that contains all previous model functions, which is proven effective for learning classification models by the strategy of label smoothing~\cite{muller2019does,zheng2022neurons}. 

Since the tug-of-war game arises due to the conflicting gradients of learning the new task and overcoming forgetting the early tasks~\cite{hadsell2020embracing}, we resort to assigning different credits to the gradients contributed by these two parts such that the combined gradient can update the model without hindering the optimization of either objective. 
For the credit assignment, we take inspiration from~\cite{yu2020gradient,hadsell2020embracing} and remove the conflicting part in one of the gradients by projecting it to the orthogonal direction of the other gradient. 
Since we have densely introduced all the previous model functions in the knowledge space, we enumerate all the possible conflict pairs and repeat the conflict removal operation in our credit assignment algorithm. 
Our approach is also partially inspired by the biological studies~\cite{yang2014sleep,hayashi2015labelling} which suggest that during the learning process of new knowledge, the human brain reduces the rate of synaptic plasticity to avoid the disturbances caused by the new knowledge~\cite{cichon2015branch}, so as to preserve the learned knowledge.

The contribution of this paper can be summarized as follows. \textbf{(i)} In order to overcome the problem of fading reliability of the knowledge space, we propose \textit{PLwF}, where the previous model functions are densely introduced to the knowledge space. \textbf{(ii)} We establish the \textit{credit assignment} regime in optimization to reconstruct the tug-of-war dynamics in CL, which better balances learning and overcoming forgetting by determining the degree of stability-plasticity of individual parameters. \textbf{(iii)} We perform extensive experiments which show the effectiveness of PLwF and credit assignments. Under the paradigm of CL without involving any old data, our method achieves notable performance improvement on CIFAR-10, CIFAR-100 and Tiny-ImageNet compared with state-of-the-art methods. 

\begin{figure*}
\centering
\includegraphics[width=0.85\textwidth]{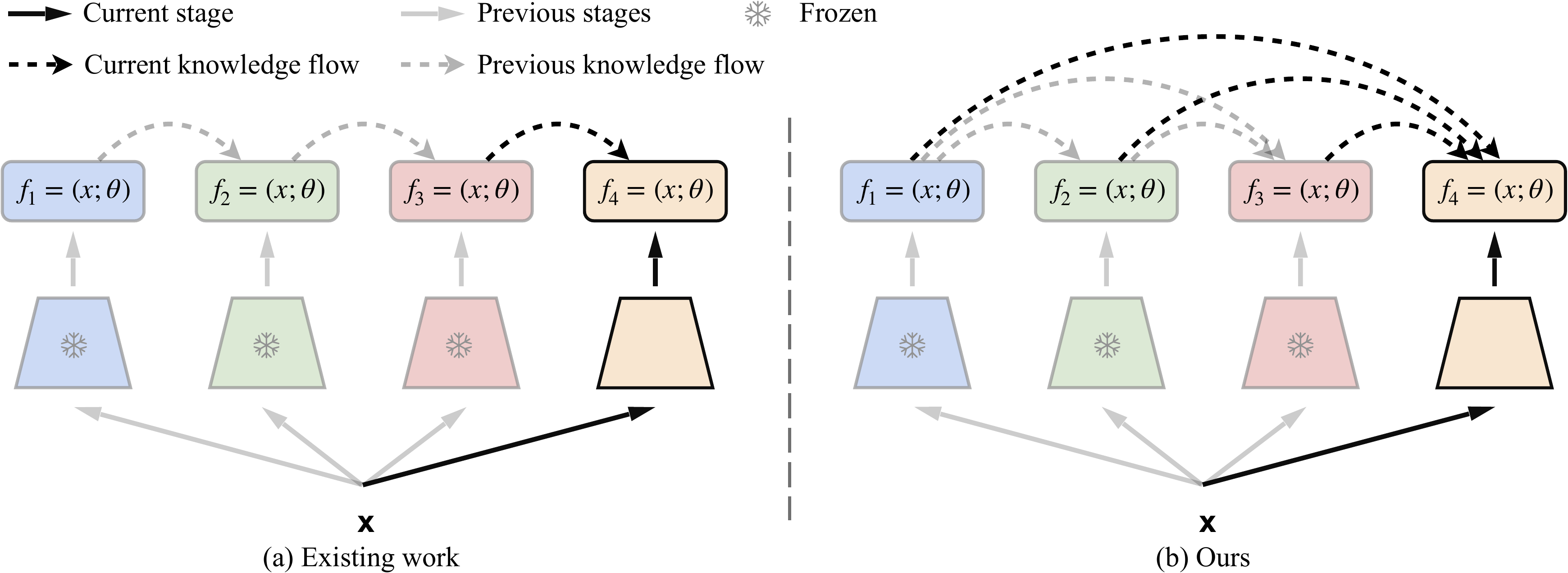}
\vspace{-2mm}
\caption{Comparison between the existing work and the proposed method. (\textbf{Note that both methods do not get access to data from previous tasks.}) (a) Existing regularization-based approaches that only leverages the model function in the last task as the knowledge space for overcoming forgetting. (b) Our PLwF approach that constructs the knowledge space by densely introducing the model functions in the previous tasks. We use \textbf{single-headed layout}, all tasks share the final classifier layer and inference is performed \textbf{without task identity}.}
\label{figure_framework}
\vspace{-4mm}
\end{figure*}

\section{Related Work}
\label{sec:related}

\textbf{Regularization-based Continual Learning.} These methods attempt to realize consolidation of the previously acquired knowledge by extending additional regularization terms. For example, both EWC~\cite{kirkpatrick2017overcoming} and EWC++~\cite{DBLP:conf/icml/Schwarz0LGTPH18} adopts second-order derivatives to measure the sensitivity of each task parameter and penalize changes in important parameters specific to previous tasks Likewise, IMM~\cite{DBLP:conf/nips/LeeKJHZ17} estimates Gaussian posteriors for the task parameters. MAS~\cite{aljundi2018memory} redefines the parameter importance measure as unsupervised settings. SI~\cite{zenke2017continual} computes path integrals on the optimization trajectory. Besides, R-walk~\cite{DBLP:conf/eccv/ChaudhryDAT18}, as a generalized version of \cite{kirkpatrick2017overcoming} and \cite{zenke2017continual}, introduces episode memory. Moreover, LwF~\cite{DBLP:journals/pami/LiH18a} utilizes knowledge distillation to retain the learned knowledge from previous tasks, which signifies that probability functions outputted by the current task are constrained. EBLL~\cite{DBLP:conf/iccv/TrikiABT17} promotes \cite{DBLP:journals/pami/LiH18a} and encourages the maintenance of low-dimensional important feature representations of previous tasks. However, plain thoughts of distillation~\cite{DBLP:journals/corr/abs-2106-05237} do not mean that no information loss occurs. 
Different from those methods described above, this work reformulate long-sequence learning tasks as a progressive matching problem for functions to minimize forgetting.

\textbf{Gradient-based Continual Learning.} This category forces the current task gradient to stay aligned with the gradient from previously learned tasks to achieve forgetting minimization. For instance, both GEM~\cite{lopez2017gradient} and A-GEM~\cite{chaudhry2018efficient} restrict the update direction of the current task gradient by calculating gradients depending on previous samples of episode memory. GPM~\cite{saha2020gradient} stores the bases from a gradient subspace of old samples into the memory in a form of gradient projection memory and updates the gradient in the orthogonal directions. Similarly, OGD~\cite{farajtabar2020orthogonal} takes gradient steps in the orthogonal direction of new task and past task gradients to minimize forgetting. FS-DGPM~\cite{deng2021flattening} further raises flattening sharpness to improve the gradient projection memory. RGO~\cite{liu2022continual} adopts an iteratively updated optimizer to modify the gradient, thus providing the model with the capability of continuous learning. Regarding these successful gradient solutions, a core element is that a gradient subspace of the previous tasks can be constructed to satisfy the memory mechanism. 
Unlike previous methods that require optimization-based gradient projection along with raw data, credit assignment is a heuristic method to remove gradient conflict without old data.

\textbf{Other approaches.} \emph{(i) Expansion-based methods.} Under this category, DEN~\cite{DBLP:conf/iclr/YoonYLH18} and HAT~\cite{DBLP:conf/icml/SerraSMK18} dynamically extend extra components to reduce the interference of new and old tasks. PNN~\cite{DBLP:journals/corr/RusuRDSKKPH16} is immune to forgetting and can leverage prior knowledge via lateral connections to previously learned features. RCL~\cite{DBLP:conf/nips/XuZ18} utilizes reinforcement learning to find an optimum structure for sequential tasks. By extending particular task parameters, APD~\cite{DBLP:conf/iclr/YoonKYH20} is able to minimize the increase in network complexity. \emph{(ii) Rehearsal-based methods}. This category attempts to alleviate the forgetting via using subsets of previous task examples as memory cells, such as iCaRL~\cite{rebuffi2017icarl} and GSS~\cite{aljundi2019gradient}. Based on different sampling strategies, they establish limited budgets in a memory buffer for rehearsal. And GEM families~\cite{lopez2017gradient,chaudhry2018efficient} proposes episode memory as the buffer. To get rid of the need to directly store raw data, recently a series of works elaborately construct special subspace of old tasks as the memory~\cite{deng2021flattening,lin2022trgp,saha2020gradient} and have shown remarkable performance.

\section{Preliminaries}
\label{sec:preliminaries}

\textbf{Continual Learning.} We consider the CL problem is comprised of $T$ tasks, denoted by $t=1, \ldots, T$, with $N$ examples sampled for each task. We assume that training instance $x_{t}$, observed in task $t$, is an i.i.d sample from distribution $\mathbb{P}_{x_t}$. 
The key of CL is how the targets $y_{t}$ is chosen, where $y_{t} \in  \mathcal{Y}_t$ is the targets space. We assume the targets for $t$th task is $\{y_{1}, y_{2}, ... , y_{t}\}$. Our goal is to learn a prediction function $f(\cdot; \theta): \mathbb{R}^{D} \longmapsto \mathbb{R}$, 
such that $f(\cdot; \theta)$ not only learns towards ground-truth targets of the current stage $y_{t}$, 
but also targets $\{y_{1}, y_{2}, ... , y_{t-1}\}$ received from the early tasks. Formally, at $t$th task, we seek to minimize the following objective:
\begin{equation}
\min _{\theta} \sum\nolimits_{t=1}^{T} \mathbb{E}_{x \sim \mathbb{P}_{x_t}}\left[\mathcal{L}\left(f(x; \theta), {y}_{t}\right)\right]
\end{equation}
\textbf{Learning without Forgetting.} 
In LwF~\cite{DBLP:journals/pami/LiH18a}, for each new task, the goal is to learn a function that maps input $x$ to the corresponding label $y_{t}$, \emph{i.e.} $f_t \left(x_{t} ; \theta_t, \vartheta_t \right)=y_{t}$, where $\theta_t$ and $\vartheta_t$ denotes parameters of the feature encoder and predictor. 
Formally, the optimization objective is defined as:
\begin{equation}
\min _{\theta_t, \vartheta_t} \mathbb{E}_{x \sim \mathbb{P}_{x_t}}\left[\mathcal{L}_{CE} \left(f_{t}\left(x ; \theta_t, \vartheta_t \right), y_{t}\right)\right]
\end{equation}
where $\mathcal{L}_{CE}$ denotes the Cross-Entropy loss.

For the old task, LwF forces the output probabilities for each image to be close to the recorded output from the last task. The KL Divergence is used to encourage the outputs of one network to approximate the outputs of another network. This procedure is also referred as function matching~\cite{DBLP:journals/corr/abs-2106-05237}. Formally, the optimization is defined as
\begin{equation}\label{eqn:lwf}
\min _{\theta_t, \vartheta_t} \mathbb{E}_{x \sim \mathbb{P}_{x_t}}\left[\mathcal{L}_{KL} \left(f_{t}\left(x ; \theta_t, \vartheta_t \right), f_{t-1}\left(x ; \theta_{t-1}, \vartheta_{t-1} \right)\right)\right]
\end{equation}
Where $\theta_{t-1}$ and $\vartheta_{t-1}$ denotes the optimal set of parameters in the last task and  $f_{t-1}(x;\theta_{t-1},\vartheta_{t-1}) \in \mathcal{Y}_1 \times \mathcal{Y}_2 \times ... \times \mathcal{Y}_{(t-1)}$ . As shown in Figure~\ref{figure_framework}a, given its alignment with the prediction function of the last task, LwF provides a knowledge space closely correlated to the current task.

Nevertheless, limited by catastrophic forgetting, the last knowledge space has suffered the evaporation of knowledge from previous tasks. 
This results in a deep learner that does not recall well the knowledge learned in earlier tasks.

\section{Progressive Learning without Forgetting}
\label{sec:plwf}

In the protocol of regularization-based methods~\cite{DBLP:journals/corr/abs-1912-03049}, determining how to progressively cumulate experience in long task sequences is not intuitive due to the lack of raw data. To tackle this problem, we propose PLwF.

In CL consisting of $T$ tasks, for naturally accumulating experience as shown in Figure~\ref{figure_framework}b, we need to learn a function $f_{t}(\cdot; \theta_{t}, \vartheta_{t})$ shared by the majority of parameters across different tasks. Therefore, the function space $\mathcal{F}_{t}$
covering more knowledge is the core. This space carries rich training signals thanks to the distribution information of labels over all the previous functions, which is defined as follows.

\textbf{Definition 1.} Suppose there exists a prediction function $f(\cdot)$ for each task, then we define a function set $\mathcal{F}_{t}=\{f_{1}\left(\cdot; \theta_{1}, \vartheta_{1}\right), f_{2}\left(\cdot; \theta_{2}, \vartheta_{2}\right), \ldots f_{t-1}\left(\cdot; \theta_{t-1}, \vartheta_{t-1}\right) \}$ 
where each function $f(\cdot)\in \mathcal{F}_{t}$ has an encoder $\theta$ and a predictor $\vartheta$ for a given task and is frozen during $t$th task.

Intuitively, when $t$ is larger than 2, the set benefits form richer prior knowledge thanks to functions of early tasks. We describe the set as a space with less vanishing knowledge for previous tasks. When optimizing for PLwF, for the new task, we follow LwF~\cite{DBLP:conf/icml/Schwarz0LGTPH18} to learn towards the label $y_t$ as formulated in Equation~\ref{eqn:lwf}. For the old tasks, we formulate the optimization objective at $t$th task as:
\begin{equation}\label{eq:pl}
\min _{\theta_{t}, \vartheta_{t}} \sum\nolimits_{i=1}^{t-1} \mathbb{E}_{x \sim \mathbb{P}_{x_t}}\left[\mathcal{L}_{KL} \left(f_t \left(x ; \theta_t, \vartheta_t \right), f_{i}\left(x ; \theta_{i}, \vartheta_{i}\right)\right)\right]
\end{equation}
where function $f_{t}(\cdot; \theta_{t}, \vartheta_{t})$ pushes its output probability to be close to the recorded output from every function in $\mathcal{F}_{t}$.
Similarly, we use KL Divergence to encourage matching of probability distributions among all functions. 
Through PLwF, we encourage the functions to progressively encode the similarities between the distribution information over labels in long sequence of tasks.
Merely learning from the outputs of the last task (Figure~\ref{figure_framework}a) can lead to a loss of knowledge of the earlier tasks, while PLwF (Figure~\ref{figure_framework}b) retains knowledge of earlier tasks, thus reducing the space of knowledge vanishing.

\textbf{Relaxation of PLwF.}
In \textit{Definition 1}, we define the function set to contain all previous functions, as illustrated in Figure~\ref{figure_framework}b. 
However, this appears to be a strong assumption to build on. 
Thus, we propose a relaxation of PLwF to fit in more scenarios. 
To be more specific, a relaxed function set can be $\Tilde{\mathcal{F}}_{t} \subset \mathcal{F}_{t}$, which includes fewer functions from previous stages. 
When applying Equation~\ref{eq:pl}, we would encourage the output probability of $f_{t}(\cdot; \theta_{t}, \vartheta_{t})$ to be close to the recorded output from every function in $\Tilde{\mathcal{F}}_{t}$.
Note that under this definition, LwF~\cite{DBLP:conf/icml/Schwarz0LGTPH18} appears to be a special case of the relaxed PLwF.

\section{Credit Assignment in PLwF}
\label{sec:ca_in_plwf}
The concept of credit assignment was proposed by~\cite{hadsell2020embracing} in CL and reflects how different parameters are responsible for expected network behaviors. The standard gradient method takes a small step along the descending direction to update the network. In such a process, the gradient independently determines whether a change in each parameter reduces the loss. At this point, credit assignment is entirely subject to plasticity. Changing settings, such as extending regularization term and using it as a gradient proxy, may help strengthen stability, but naive credit assignment can cause a gradient update with forced tug-of-war dynamics. Therefore, it is of value to recreate refined tug-of-war dynamics through proper credit assignment.



Since we densely introduced all the previous model functions through PLwF, the credit assignment regime needs to handle more complex tug-of-war dynamics. Suppose an essence of optimization problem in the PLwF comes from the tug-of-war of gradients. 
In an SGD~\cite{DBLP:conf/icml/Zinkevich03} optimizer, parameters are updated as follows:
\begin{equation}
\theta_{i}:=\theta_{i-1}-\eta \sum_{t} \nabla \mathcal{L}_{t}(\theta)
\end{equation}
However, such a solution is unable to develop reliable credit assignments in learning process. To this end, we recreate a regime by modifying the gradient in PLwF, thereby minimizing forgetting. We allow positive interactions between the gradients of different tasks to overcome the stability-plasticity dilemma. Inspired by~\cite{yu2020gradient,hadsell2020embracing}, we define the following conditions to explore credit assignment in PLwF.

\textbf{Definition 2.} The credit assignment measure $\phi$ is given by the cosine similarity $\phi_{g_{a}, g_{b}}=\frac{\left\langle g_{a}, g_{b}\right\rangle}{\left\|g_{a}\right\| \cdot\left\|g_{b}\right\|}$, where $g_{a}$ and $g_{b}$ are gradients from two arbitrary tasks.

\textbf{Definition 3.} If credit assignment measure between two tasks $\phi_{g_{a},g_{b}} < 0$, then the tug-of-war dynamics exists, otherwise it does not.

To better determine the degree of stability-plasticity of each parameter and avoid negative changes of gradients, we define a gradient assignment matrix $\mathcal{M}$ in each iteration.

\textbf{Definition 4.} Suppose there exists a gradient set $G=\left\{\mathrm{g}_{1}, \mathrm{g}_{2}, \ldots, \mathrm{g}_{t}, \forall t \in T \right\}$. For two arbitrary elements $g_{a}$, $g_{b}$ in the set, there is a measure $\phi_{g_{a}, g_{b}}$ with the symmetry condition satisfied. Then there exists a matrix $\mathcal{M}$ of size $T * T$ as the gradient assignment matrix, reflecting the credit assignment of two arbitrary elements. 

The set $G$ comes from the optimization process of PLwF. If considering symmetry, $\phi_{g_{a}, g_{b}}=\phi_{g_{b}, g_{a}}$.
Thereby, we focus on the entries in the upper or lower triangle of $\mathcal{M}$ and the entries on the diagonal of $\mathcal{M}$.
Throughout the entire process, \emph{Definition 4} investigates the assignment conditions of each gradient across all tasks. Particularly, the matrix $\mathcal{M}$ could capture the tug-and-war dynamics in each iteration.

\begin{table}
\centering
\caption{Gradient assignment matrix example following the notations in \emph{Definition 4}. $\phi$ denotes credit assignment measure. Pink denotes $\phi_{g_{a}, g_{b}} < 0$.}
\vspace{-2mm}
\label{table_gradient_matrix}
\begin{tabular}{@{}l|lllll@{}}
\toprule
$\phi$       & $g_{1}$                     & $g_{2}$     & ... & $g_{t-1}$     & $g_{t}$     \\ \midrule
$g_{1}$   & $\phi_{1,1}$                   & \cellcolor{pink}$\phi_{1,2}$   & ... & $\phi_{1,t-1}$   & $\phi_{1,t}$   \\
$g_{2}$   & \bcancel{\cellcolor{pink}$\phi_{2,1}$}   & $\phi_{2,2}$   & ... & \cellcolor{pink}$\phi_{2,t-1}$   & $\phi_{2,t}$   \\
...       & ...                         & ...         & ... & ...           & ...         \\
$g_{t-1}$ & $\phi_{t-1,1}$                 & \bcancel{\cellcolor{pink}$\phi_{t-1,2}$} & ... & $\phi_{t-1,t-1}$ & \cellcolor{pink}$\phi_{t-1,t}$ \\
$g_{t}$   & $\phi_{t,1}$                   & $\phi_{t,2}$   & ... & \bcancel{\cellcolor{pink}$\phi_{t,t-1}$}   & $\phi_{t,t}$   \\ \bottomrule
\end{tabular}
\vspace{-4mm}
\end{table}

\begin{figure*}
\centering
  \begin{subfigure}{0.3\textwidth}
      \includegraphics[width=2.in]{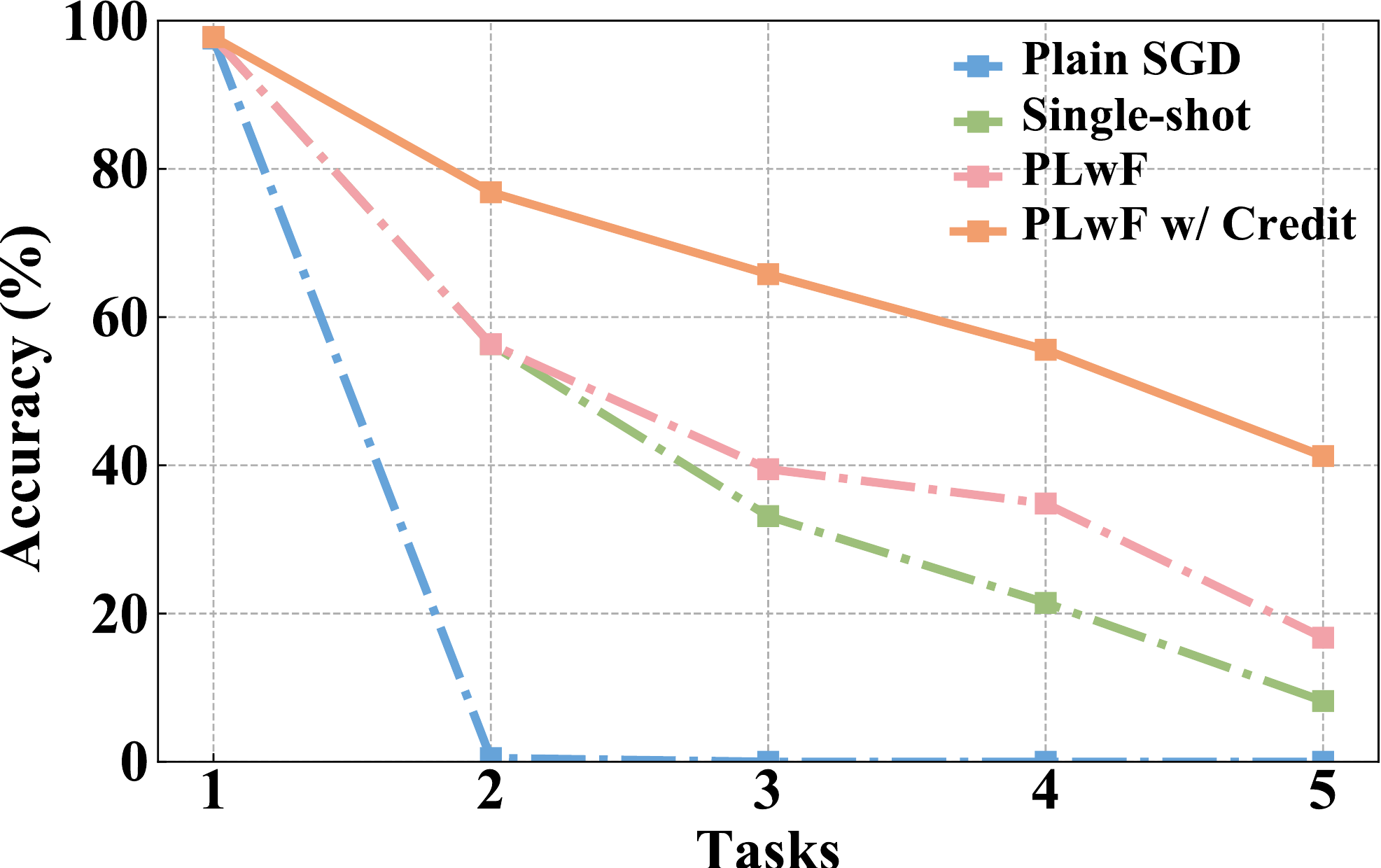}
        \caption{5-Split CIFAR-10}
    \end{subfigure}
  \hspace{.3cm}
  \begin{subfigure}{0.3\textwidth}
      \includegraphics[width=2.in]{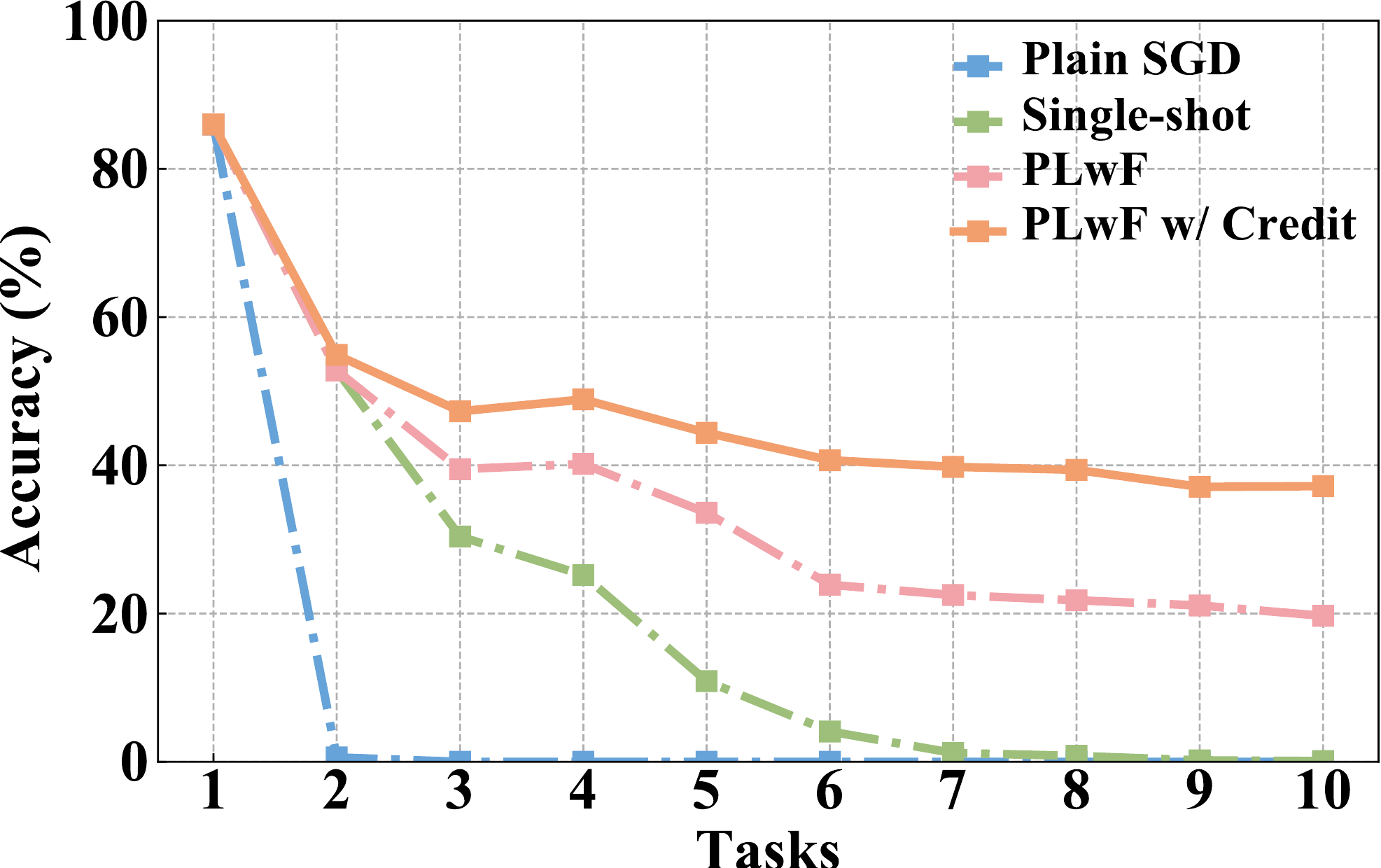}
        \caption{10-Split CIFAR-100}
    \end{subfigure}
  \hspace{.3cm}
  \begin{subfigure}{0.3\textwidth}
      \includegraphics[width=2.in]{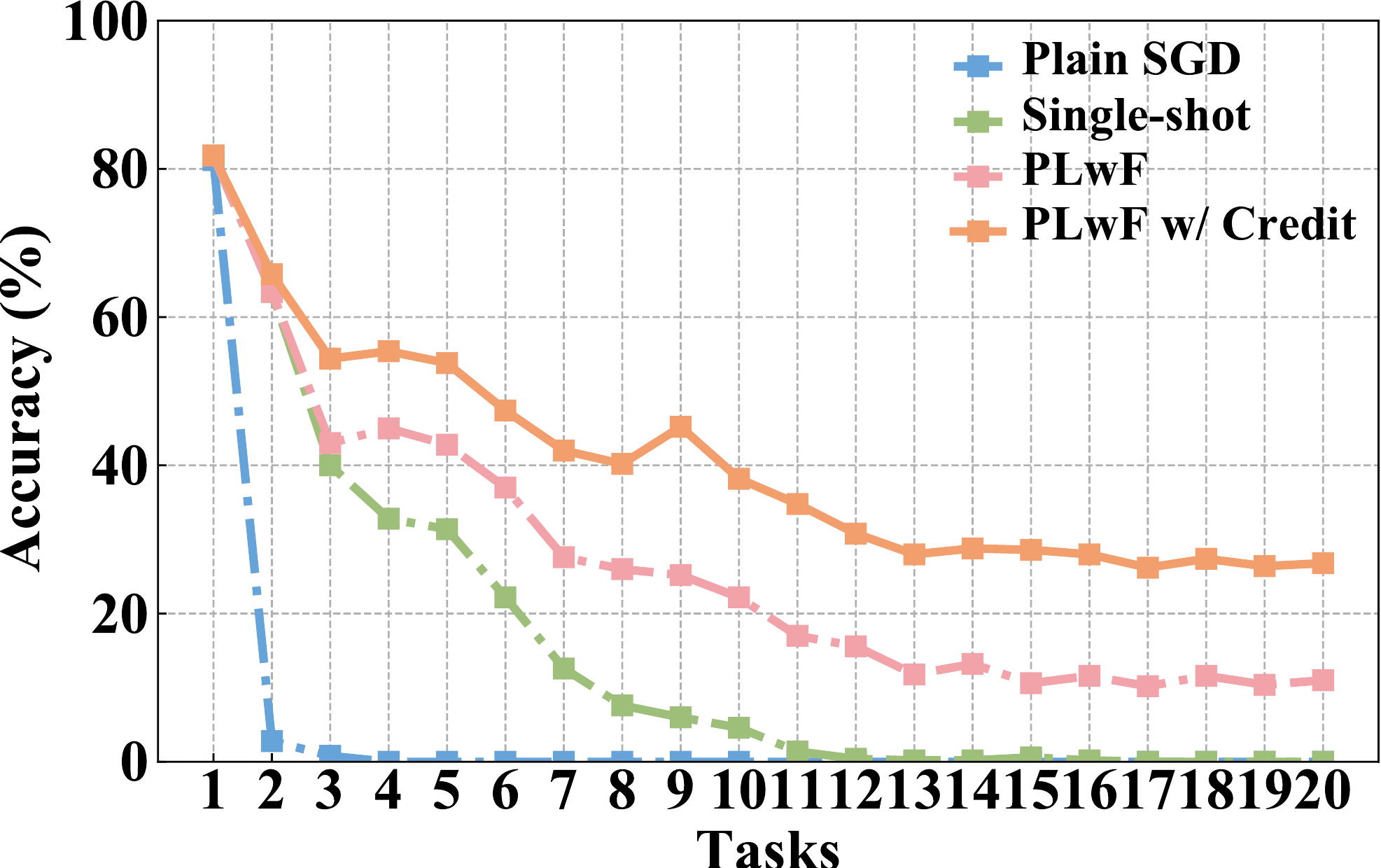}
        \caption{20-Split CIFAR-100}
    \end{subfigure}
\vspace{-3mm}
\caption{Performance variation on the first task when trained over 5 tasks, 10 tasks and 20 tasks on CIFAR-10 and CIFAR-100.}
\vspace{-6mm}
\label{figure_controlling_forgetting}  
\end{figure*}

To generate reliable credit assignment regime in the optimization process, the following steps are performed: \textbf{(i)} Map the credit assignment measure $\phi_{g_{a},g_{b}}$ in the set $G$ and calculate the credit assignment matrix $\mathcal{M}$. \textbf{(ii)} Extract the pairs of the gradients located above (or below) the main diagonal with $\phi_{g_{a},g_{b}} < 0$ (Pink in Table 1) in the matrix $\mathcal{M}$ as a subset $\mathcal{H}$. \textbf{(iii)} For each pair $\left(g_{a}, g_{b}\right)$ of the gradients in $\mathcal{H}$, replace $g_{a}$ by its projection onto the normal plane of $g_{b}$: $g_{a}=g_{a}-\frac{g_{a} \cdot g_{b}}{{\left\|g_{b}\right\|}^2} g_{b}$.

The above procedure is executed for every iteration during optimization. 
This reduces the degree of disturbance of gradient applied in per batch of each task towards other tasks in the batch, thereby reducing the tug-of-war dynamics. After this, the parameter is updated as: 
\begin{equation}
\theta_{i}:=\theta_{i-1}-\eta \sum_{t} (\nabla \mathcal{L}_{t}(\theta))^{credit}
\end{equation}
Where the \emph{credit} term indicates the regime operator for the gradient. Through the simple reset, our solution replaces the original gradients with updated ones and passes them to the respective optimizer. The experimental results validate the hypothesis of recreating a careful credit assignment regime and the improvement in CL capacity intuitively demonstrates the alleviation of the stability-plasticity dilemma.

\section{Experimental Setup}
\label{sec:exp_set}

\textbf{Datasets.} 
Following typical research on class incremental learning~\cite{krizhevsky2009learning,delange2021continual,zenke2017continual}, we evaluate the performance of PLwF on CIFAR-10, CIFAR-100 and Tiny-ImageNet. 
Specifically, we split CIFAR-10 into 5 tasks with 2 classes per task, split CIFAT-100 into 10/20 tasks with 10/5 classes per task and split Tiny-ImageNet into 10 tasks with 20 classes per task.
Specially, we construct short task sequences 2-Split CIFAR-10, 2-Split CIFAR-100, and 2-Split Tiny-ImageNet for empirical analysis of credit assignment, in which they divide 10, 100, and 200 classes of CIFAR-10 and CIFAR-100, and Tiny-ImageNet into 2 tasks with 5, 50, and 100 classes per task. For a fair comparison with different methods, we use all the classes in the same order to perform experiments.

\textbf{Methods for Comparison.} PLwF strictly follows the regularization method, we do not store any old samples from the raw data when learning new classes. Therefore, we first compare our method with several regularization-based methods: EWC~\cite{kirkpatrick2017overcoming}, EWC++~\cite{DBLP:conf/icml/Schwarz0LGTPH18}, MAS~\cite{aljundi2018memory}, and SI~\cite{zenke2017continual}. Additionally, we also compare several rehearsal-based methods: ER~\cite{NEURIPS2019_fa7cdfad}, GEM~\cite{lopez2017gradient}, A-GEM~\cite{chaudhry2018efficient}, FDR~\cite{DBLP:conf/iclr/BenjaminRK19}, GSS~\cite{aljundi2019gradient}, HAL~\cite{DBLP:conf/aaai/ChaudhryGDTL21}, and PODNET~\cite{DBLP:conf/eccv/DouillardCORV20}. The buffer size for all rehearsal-based methods is set to 500 following \cite{buzzega2020dark}. For the compared methods, we follow the open-source implementations~\cite{buzzega2020dark,Hsu18_EvalCL} with accompanying best hyper-parameters to perform CL.

\textbf{Architectures and Training Details.} Similar to~\cite{rebuffi2017icarl,buzzega2020dark}, we employ the ResNet-18~\cite{he2016deep} in CIFAR-10 experiments. In CIFAR-100 and Tiny-ImageNet experiments, we use the ResNet-32 similar to \cite{zhou2021pycil}. To ensure a fair comparison, all models are trained with the vanilla-SGD~\cite{DBLP:conf/icml/Zinkevich03} optimizer. For experiments on CIFAR-10 and CIFAR-100, the network is trained with the batch size of 128 while the batch size is set to 32 in Tiny-ImageNet. For all datasets, we train the initial task with 200 epochs and the remaining task with 250 epochs. All the experiments are performed on 1 NVIDIA TITAN GPU. Finally, we report average accuracy over all tasks and the last-task accuracy, the latter of which indicates the degree of forgetting. For all experiments, we evaluate and compare our method in \textbf{single-headed layout} where all tasks share the final classifier layer and inference is performed \textbf{without task identity}.


\begin{table*}
\centering
\small
\caption{Evaluation results (\%) on different datasets. `Reg.' and `Reh.' indicates the regularization-based and rehearsal-based methods. `Avg' and `Last' indicates the average accuracy over all tasks, the last-task accuracy. `-' indicates experiments cannot be performed due to intractable training time (\textit{e.g.} GEM on Tiny-ImageNet).}
\vspace{-2mm}
\label{table_main}
\begin{tabular}{@{}ccccccccccc@{}}
\toprule
                         & \multicolumn{2}{c}{Techniques} & \multicolumn{2}{c}{5-Split CIFAR-10} & \multicolumn{2}{c}{10-Split CIFAR-100} & \multicolumn{2}{c}{20-Split CIFAR-100} & \multicolumn{2}{c}{10-Split Tiny-ImageNet}                          \\ \cmidrule(l){2-11} 
\multirow{-2}{*}{Method} & Reg.           & Reh.          & Avg           & Last          & Avg            & Last          & Avg                          & Last     & Avg          & Last           \\ \midrule
EWC~\cite{kirkpatrick2017overcoming}                      &\checkmark      &               & 41.00            & 11.24         & 30.73          & 13.98    & 18.25      & 6.17    & 25.36                        & 12.40                         \\
EWC++~\cite{DBLP:conf/icml/Schwarz0LGTPH18}                    &\checkmark      &               & 42.02         & 18.67         & 25.27          & 8.93   & 13.39       & 3.17      & 20.81                        & 7.06                         \\
MAS~\cite{aljundi2018memory}                      &\checkmark      &               & 44.25          & 22.40    & 34.42           & 15.88      & 18.85         & 6.98    & 26.57           & 10.01                \\
SI~\cite{zenke2017continual}                       &\checkmark      &               & 44.15         & 19.60     & 25.78          & 9.24     & 16.22    & 4.71    & 20.13                        & 7.03                         \\ \midrule
GEM~\cite{lopez2017gradient}                      &\checkmark      &\checkmark     & 50.12         & 30.92         & 28.57          & 12.25     & 19.87    & 9.09    & -                            & -                            \\
A-GEM~\cite{chaudhry2018efficient}                    &\checkmark      &\checkmark     & 47.38         & 19.68         & 25.81          & 9.32     & 16.16   & 4.75     & 22.58                        & 7.99                         \\
ER~\cite{NEURIPS2019_fa7cdfad}        &         &\checkmark     & 62.54    & 45.7      & 39.31     & 19.61    & 25.43    & 9.38     & 26.60   & 10.19                     \\
FDR~\cite{DBLP:conf/iclr/BenjaminRK19}                      &                &\checkmark     & 54.34         & 31.34         & 40.51          & 23.07    & 29.20         & \textbf{15.63}     & 29.67                        & 10.89                        \\
GSS~\cite{aljundi2019gradient}             &                &\checkmark     & \textbf{68.63}         & 45.07         & 32.90           & 12.81     & 22.85    & 7.19    & 26.30                         & 9.29                         \\
HAL~\cite{DBLP:conf/aaai/ChaudhryGDTL21}          &                &\checkmark     & 62.63         & 45.05         & 30.15          & 12.80      & 25.46     & 11.90    & 18.75                        & 5.99                         \\ 
PODNET~\cite{DBLP:conf/eccv/DouillardCORV20}      &                &\checkmark     & 67.87        & 45.20    & 44.18     & 21.44    & 28.49         & 11.51    & 29.16            & 13.6                \\ \midrule
Ours                     &\checkmark      &               & 66.36         & \textbf{51.15} & \textbf{44.72}          & \textbf{28.79}     & \textbf{30.65}     & 15.09     & \textbf{30.95}                        & \textbf{17.02} \\ \bottomrule
\end{tabular}
\vspace{-4mm}
\end{table*}

\section{Results and Discussions}
\label{sec:exp_result}

\textbf{Controlling Forgetting.} 
As learning continues, the performance of the initial task tends to undergo the most thorough forgetting. 
To illustrate how PLwF ameliorates forgetting, we show the changes in performance on the first task as learning continues on 5-Split CIFAR-10, 10-Split CIFAR-100 and 20-Split CIFAR-100 in Figure \ref{figure_controlling_forgetting}.
The same trend is observed in all datasets: \textbf{(i)} Plain-SGD demonstrates utter catastrophic forgetting due to the standard learning protocol, and its performance on the first task collapses to zero when learning the second task; 
\textbf{(ii)} Specifically, the single-shot setting represents a naive learning method which only adopts the function at the last task as a basic learning space. This setting benefits from knowledge transferred from the last task, preserving some performance; 
\textbf{(iii)} In comparison, PLwF shows a considerably gentler slope of the forgetting curve, which shows intransigence towards earlier tasks;
\textbf{(iv)} After credit assignment mediates the tug-of-war dynamics, forgetting is further controlled, which is reflected in the gentler slope of the forgetting curve than PLwF. To sum up, we control forgetting by using functions before they become untrustworthy, and by removing gradient conflict.

\textbf{Main Results.} Table \ref{table_main} reports results on all datasets, which sheds light on the following observations: \textbf{(i)} \emph{Fair comparison with regularization-based methods:} In this setup, any raw data from the past stages are prohibited, the proposed method achieves state-of-the-art performance on all datasets. When compared to EWC~\cite{kirkpatrick2017overcoming}, EWC++~\cite{DBLP:conf/icml/Schwarz0LGTPH18}, MAS~\cite{aljundi2018memory}, and SI~\cite{zenke2017continual}, the gap appears unbridgeable. From our perspective, this category of methods has its root in the important parameters from the earlier tasks, whose reliability dwindles at subsequent stages of learning, thereby resulting in limited performance. \textbf{(ii)} \emph{Comparison with rehearsal-based methods:} The proposed method presents an outstanding performance towards all previous methods, despite no raw data from any past tasks being stored. GEM~\cite{lopez2017gradient} and A-GEM~\cite{chaudhry2018efficient} resort to gradients likewise, but their episode memory present a less satisfactory effect when evaluating the distribution information of classes. \textbf{(iii)} Particularly, GSS~\cite{aljundi2019gradient} is slightly better than our method on CIFAR-10 on Avg, which is attributed to its efficient buffer strategy. However, our method takes a 6.08\% lead on the Last task. Moreover, on Split CIFAR-100 and Split Tiny-ImageNet with greater difficulty, our method significantly exceeds GSS. GSS presents a decreased performance when evaluating on challenging tasks. To sum up, the performance of the proposed method surpasses the most advanced method under the protocol of regularization-based methods. Besides, it also presents better or comparable performance than rehearsal-based methods even without relying on any raw data. Remarkably, our method performs more stably on more complex datasets due to the distribution information over labels from all the previous functions.

\begin{figure}
   \begin{subfigure}{0.47\linewidth}
       \includegraphics[width=1.6in]{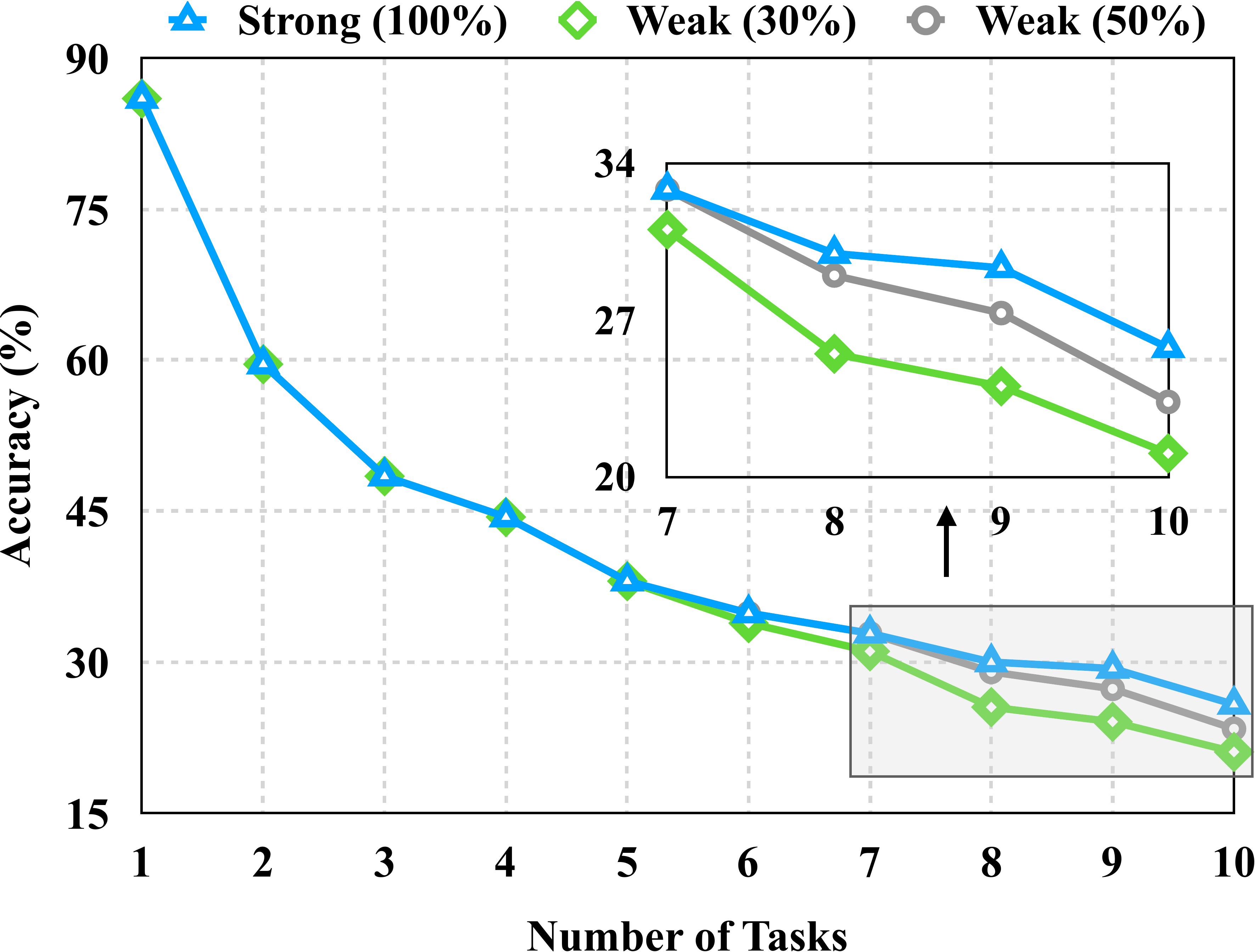}
        \caption{10-Split CIFAR-100.}
    \end{subfigure}
    \hspace{.2cm}
   \begin{subfigure}{0.47\linewidth}
       \includegraphics[width=1.6in]{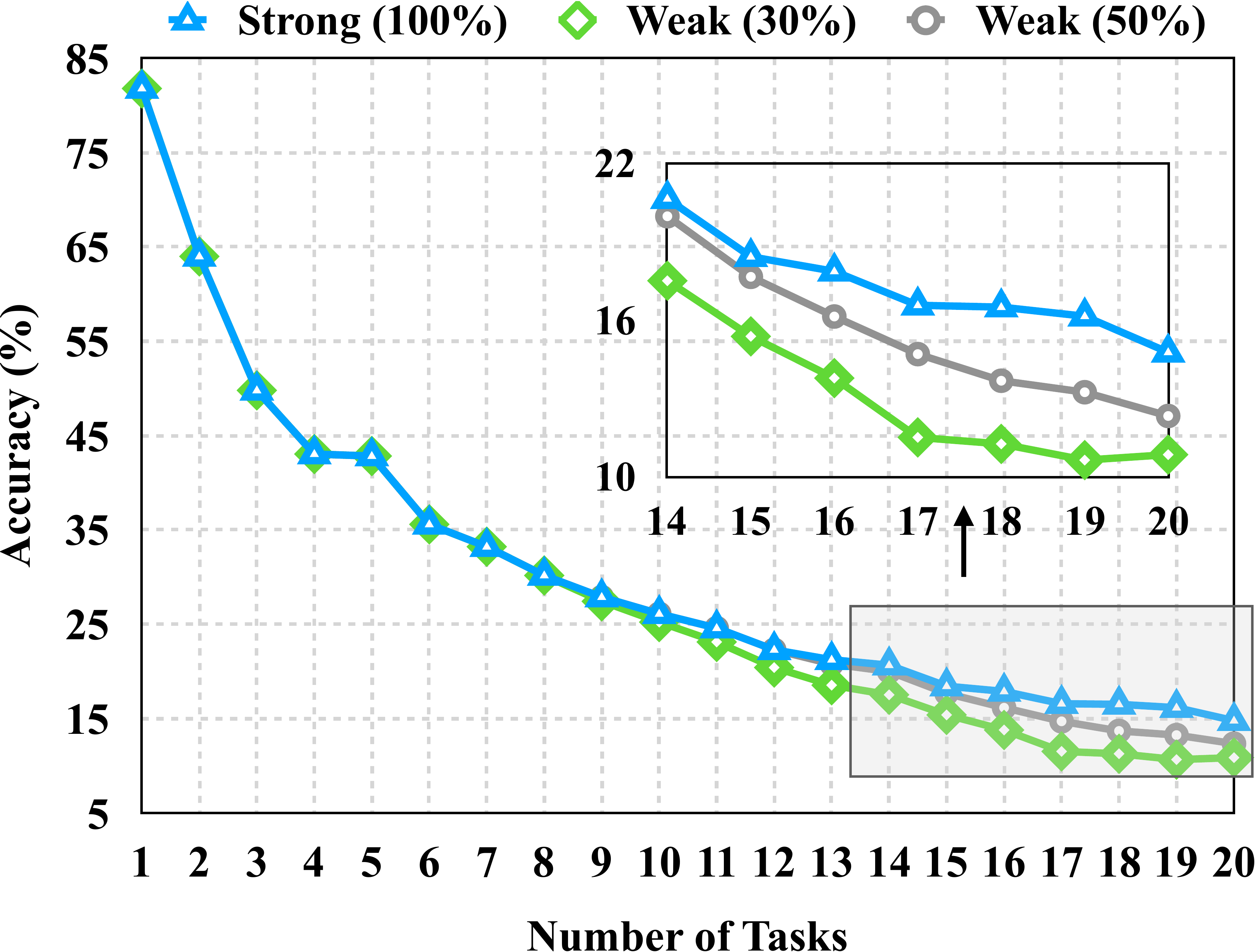}
        \caption{20-Split CIFAR-100}
    \end{subfigure}
    \vspace{-.3cm}
\caption{The average accuracy measured by the end of each task for the relaxed PLwF on Split CIFAR-100.}
\vspace{-7mm}
\label{fig:rex_assump}  
\end{figure}




\begin{table*}
\centering
\small
\caption{Ablation study (\%) on different datasets. `Avg.' and `Last' indicates average accuracy over all tasks, last-task accuracy.}
\vspace{-2mm}
\label{table_ablation}
\begin{tabular}{@{}ccccccc@{}}
\toprule
\multirow{2}{*}{Method} & \multicolumn{2}{c}{5-Split CIFAR-10}             & \multicolumn{2}{c}{10-Split CIFAR-100}            & \multicolumn{2}{c}{10-Split Tiny-ImageNet}        \\ \cmidrule(l){2-7} 
                        & Avg                   & Last                  & Avg                   & Last                  & Avg                   & Last                  \\ \midrule
Single-shot             & 57.56               & 39.51                  & 33.04                     & 15.94                     & 23.36               & 8.43                  \\
vanilla PLwF    & 64.38(\textcolor{forestgreen}{+9.82})           & 48.07(\textcolor{forestgreen}{+8.56})         & 42.94(\textcolor{forestgreen}{+9.9})           & 25.79(\textcolor{forestgreen}{+9.85})           & 27.69(\textcolor{forestgreen}{+7.59})     & 15.12(\textcolor{forestgreen}{+8.59})         \\
PLwF w/ Credit     & \textbf{66.36}(\textcolor{forestgreen}{+1.98}) & \textbf{51.15}(\textcolor{forestgreen}{+3.08}) & \textbf{44.72}(\textcolor{forestgreen}{+1.78}) & \textbf{28.79}(\textcolor{forestgreen}{+3.00}) & \textbf{30.95}(\textcolor{forestgreen}{+3.26}) & \textbf{17.02}(\textcolor{forestgreen}{+1.90}) \\ \bottomrule
\end{tabular}
\vspace{-2mm}
\end{table*}

\begin{figure*}
\begin{minipage}[c]{0.7\textwidth}
\footnotesize
\captionof{table}{The effects (\%) of credit assignment on a short task sequences (2-steps).}
\vspace{-2mm}
\label{table_short}
\begin{tabular}{@{}ccccccc@{}}
\toprule
                         & \multicolumn{2}{c}{2-Split CIFAR-10}          & \multicolumn{2}{c}{2-Split CIFAR-100}         & \multicolumn{2}{c}{2-Split Tiny-ImageNet}      \\ \cmidrule(l){2-7} 
\multirow{-2}{*}{Method} & Avg                                          & Last         & Avg                   & Last                  & Avg                   & Last                  \\ \midrule
w/o Credit               & 80.53                                        & 70.59        & 61.55                 & 46.84                 & 33.68                 & 25.82                 \\
w/ Credit        & \textbf{81.52}(\textcolor{forestgreen}{+0.99}) & \textbf{72.58}(\textcolor{forestgreen}{+1.99}) & \textbf{62.21}(\textcolor{forestgreen}{+0.66}) & \textbf{48.15}(\textcolor{forestgreen}{+1.31}) & \textbf{34.42}(\textcolor{forestgreen}{+0.74}) & \textbf{27.17}(\textcolor{forestgreen}{+1.35}) \\ \bottomrule
\end{tabular}
\end{minipage}
\hspace{.5cm}
\begin{minipage}[c]{0.3\textwidth}
\footnotesize
\captionof{table}{Credit assignment on EWC.}
\vspace{-2mm}
\label{table_EWC}
\begin{tabular}{@{}ccc@{}}
\toprule
\multirow{2}{*}{Method} & \multicolumn{2}{c}{5-Split CIFAR-10}                \\ \cmidrule(l){2-3} 
                        & Avg                   & Last                  \\ \midrule
EWC                     & 41.00                 & 11.24                 \\
w/ Credit                & \textbf{43.98}(\textcolor{forestgreen}{+2.98}) & \textbf{19.74}(\textcolor{forestgreen}{+8.50}) \\ \bottomrule
\end{tabular}
\end{minipage}
\vspace{-4mm}
\end{figure*}


\textbf{Relaxing PLwF.}
As indicated in Section~\ref{sec:plwf}, we show a more relaxed version of PLwF by adopting a subset of previous functions $\Tilde{\mathcal{F}}_{t}$.
To validate the impact of the relaxed PLwF, we set up the following experiments: 
\textbf{(i)} Strong assumptions (to use all previous functions in Equation \ref{eq:pl}). 
\textbf{(ii)} Weak assumptions (to use the $(t-1)$th function and the first 30\% or 50\% functions from earlier stages in Equation \ref{eq:pl}).
The results in Figure \ref{fig:rex_assump} reveal that when adopting the $(t-1)$th and the first 50\% functions, PLwF suffers a performance drop by only 0.54\% (Avg) and 0.69\% (Avg) on 10 steps and 20 steps setting while reducing computational overhead by about 40\% compared to strong assumptions; when adopting the $(t-1)$th and the first 30\% functions, PLwF suffers a performance drop by only 1.73\% (Avg) and 1.18\% (Avg) on 10 steps and 20 steps setting while reducing computational overhead by about 60\% compared to strong assumptions.
Similar ideas \cite{DBLP:arxiv/corr/gdm} are adopted in~\cite{DBLP:arxiv/corr/gdm} to boost the sampling efficiency of diffusion models by up to 256 times. 
This performance-speed trade-off brought by adaptively changing the matching complexity could potentially benefit the application of PLwF.


\begin{figure}
   \begin{subfigure}{0.47\linewidth}
       \includegraphics[width=1.6in]{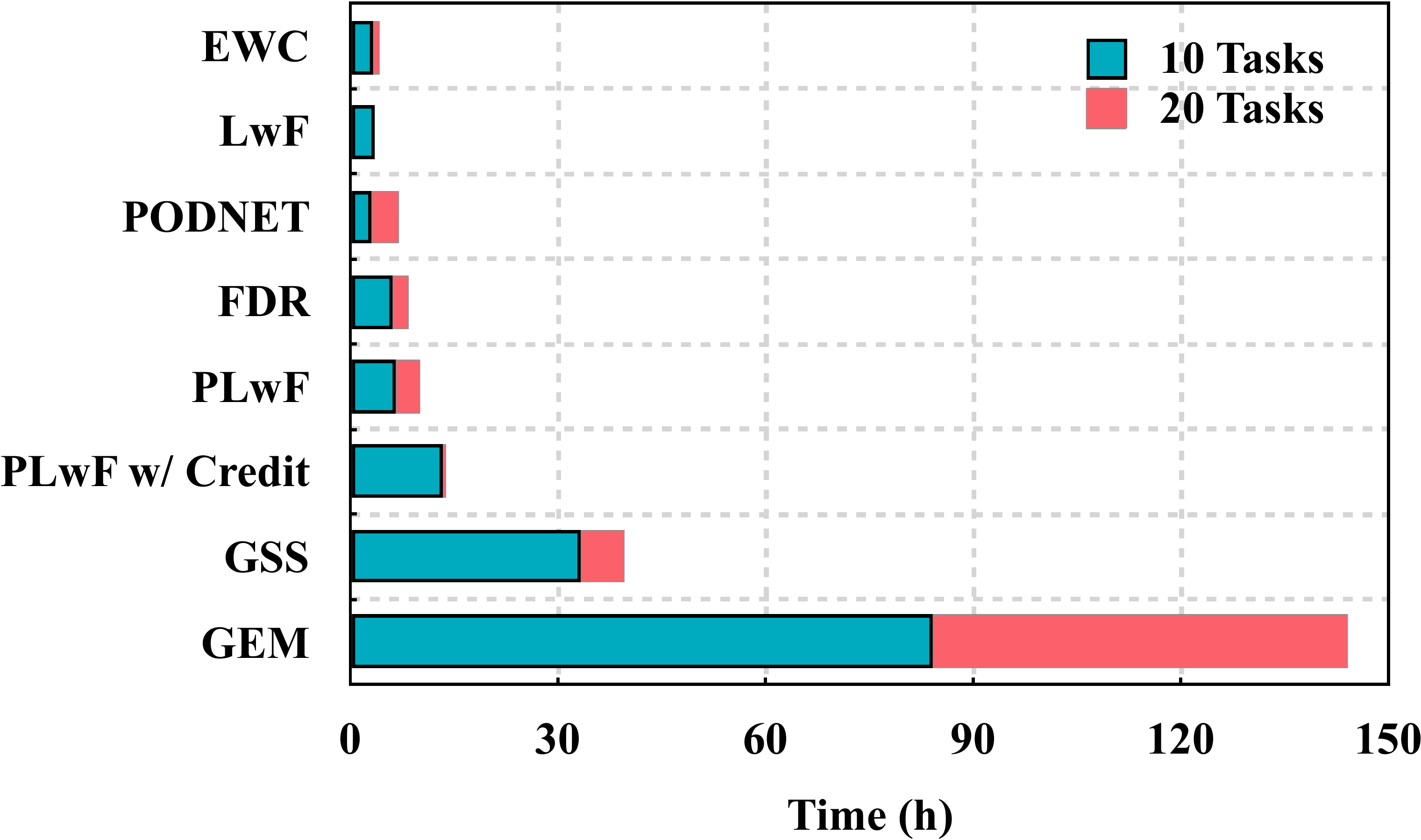}
        \caption{Training time}
    \end{subfigure}
    \hspace{.2cm}
   \begin{subfigure}{0.47\linewidth}
       \includegraphics[width=1.6in]{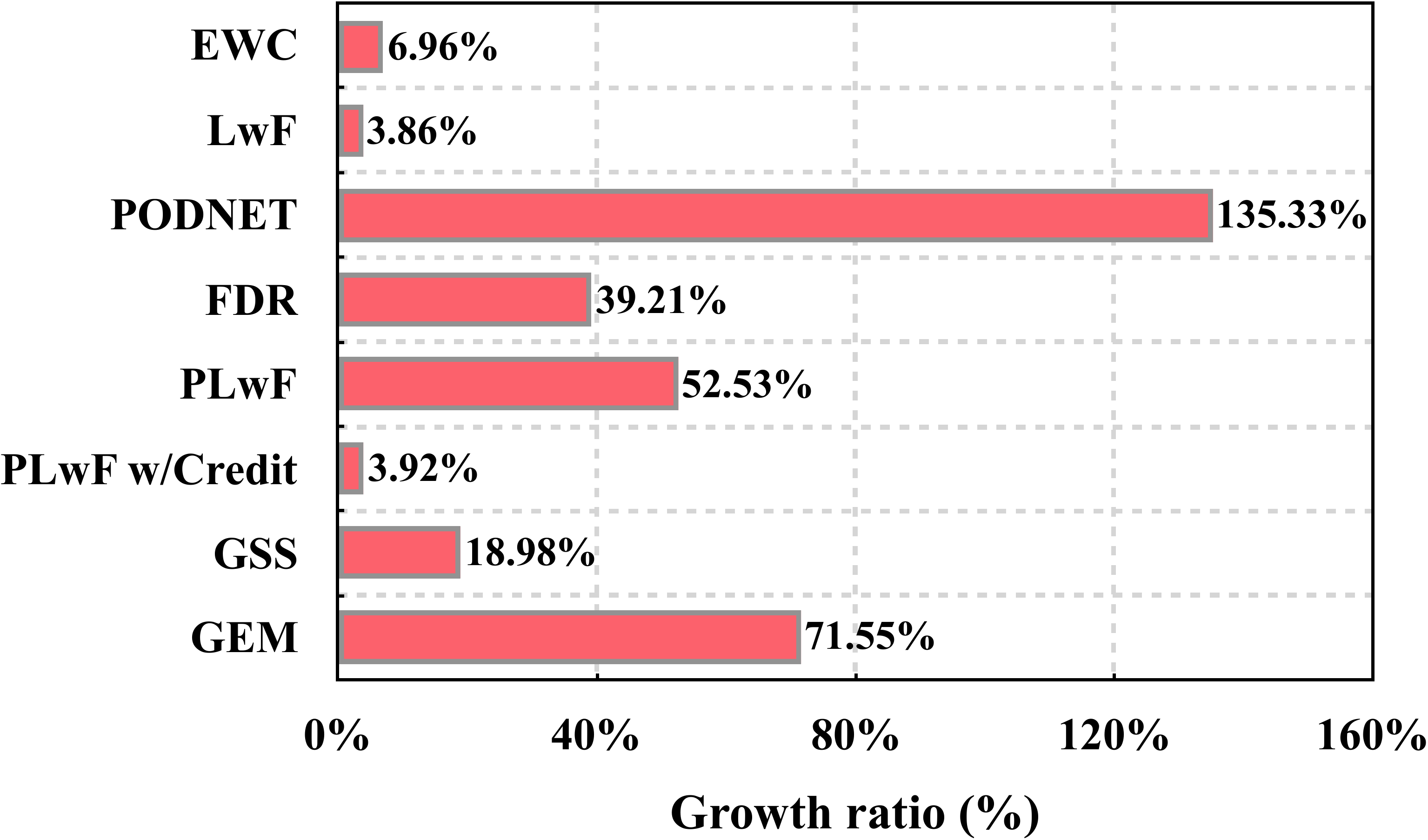}
        \caption{Increasing ratio}
    \end{subfigure}
    \vspace{-.3cm}
\caption{Model efficiency on CIFAR-100 (10 steps and 20 steps).}
\vspace{-3mm}
\label{fig:times}  
\end{figure}

\textbf{Model Efficiency.} In this subsection, we assess the performance of PLwF, PLwF w/ credit and typical CL methods in terms of model efficiency. 
All experiments are executed on a server with one NVIDIA TITAN Graphic Card. 
Figure \ref{fig:times} reports the training time and the increasing ratio of training 20 steps over training 10 steps on CIFAR-100.
We draw the following observations about the training time: \textbf{(i)} In both benchmarks, plain PLwF has a comparable running time to FDR (Figure \ref{fig:times}a). 
\textbf{(ii)} The computational overhead of our method is significantly lower than rehearsal-based methods like GSS and GEM. 
We draw the following observations about the increasing ratio: 
\textbf{(i)} As the number of tasks increases from 10 to 20, the increasing ratio of PLwF in computation is less than PODNET (135.33\%) and GEM (71.55\%) (Figure \ref{fig:times}b).
\textbf{(ii)} PLwF w/ credit imposes trivial time overhead (3.92\%) even if trained on 20 tasks.
These observations reveal the practicality of PLwF.

\textbf{Limitations.} 
PLwF progressively transfers knowledge from previous functions to current stage, while this can result in the growth of computation when we have increasing number of tasks. 
We provide brief discussions here on how to potentially reduce computation, and more can be found in the Appendix.
First, vanilla function matching involves cumbersome repetitive forwarding, and our method exacerbates the process due to the increase in the number of matching functions. This suggests that growth can be ameliorated by reducing repeated forwarding, \textit{e.g.} \textbf{Fast Knowledge Distillation}~\cite{shen2021afast} that can speed up 2$\sim$4x without compromising accuracy in a single-shot matching.
In our case, we could largely speed up the training due to a large number of tasks. 
Textbrewer~\cite{yang-etal-2020-textbrewer} provides a plug-and-play support for this solution to enable a fast instantiation.
Second, \textbf{Prune, then Distill}~\cite{DBLP:journals/corr/abs-2109-14960} reduces computational overhead in the matching process by pruning models, which can even further improve the performance in a low-cost manner.
In-depth explorations are left as future work.



\begin{table*}
\centering
\small
\caption{Results (\%) of order-agnostic behavior on 10-Split CIFAR-100, averaged across 4 random seeds. `-' is similar to Table \ref{table_main}}
\vspace{-2mm}
\label{table_main_seed}
\begin{tabular}{@{}c|ccccccccccc|c@{}}
\toprule
Method & EWC   & EWC++ & MAS   & SI    & GEM & A-GEM & ER    & FDR   & GSS   & HAL   & PODNET & Ours           \\ \midrule
Avg    & 30.77 & 24.03 & 32.99  & 25.87 & -   & 25.63 & 39.95 & 40.77 & 32.98 & 29.88 & 45.22  & \textbf{46.69} \\
Last   & 15.11 & 7.09  & 15.25  & 9.29  & -   & 9.30  & 20.13 & 23.56 & 12.93 & 12.79 & 21.08  & \textbf{27.77} \\ \bottomrule
\end{tabular}
\vspace{-3mm}
\end{table*}

\section{Empirical Analysis of Method}
\label{sec:emp}

\textbf{An Intuitive Explanation of PLwF.}
To examine the benefit of adapting previous functions, we experiment on a special case of the relaxed PLwF --- adopting the first function of all previous functions. 
More experiments on other choices of functions are detailed in the Appendix.
Figure \ref{fig:1st_distance} reveals the following observations: 
\textbf{(i)} The results on the first ten classes remains decent throughout ten incremental steps (the orange line in Figure \ref{fig:1st_distance}(a)).
\textbf{(ii)} At the $10$th step, the result of the first ten classes is substantially higher than other ten classes (the orange column in Figure \ref{fig:1st_distance}(b)). 
We conclude that a deep learner can recall the knowledge learned in an early task by adopting the corresponding function, which makes the current function more trustworthy.


\begin{figure}
   \begin{subfigure}{0.46\linewidth}
       \includegraphics[width=1.5in]{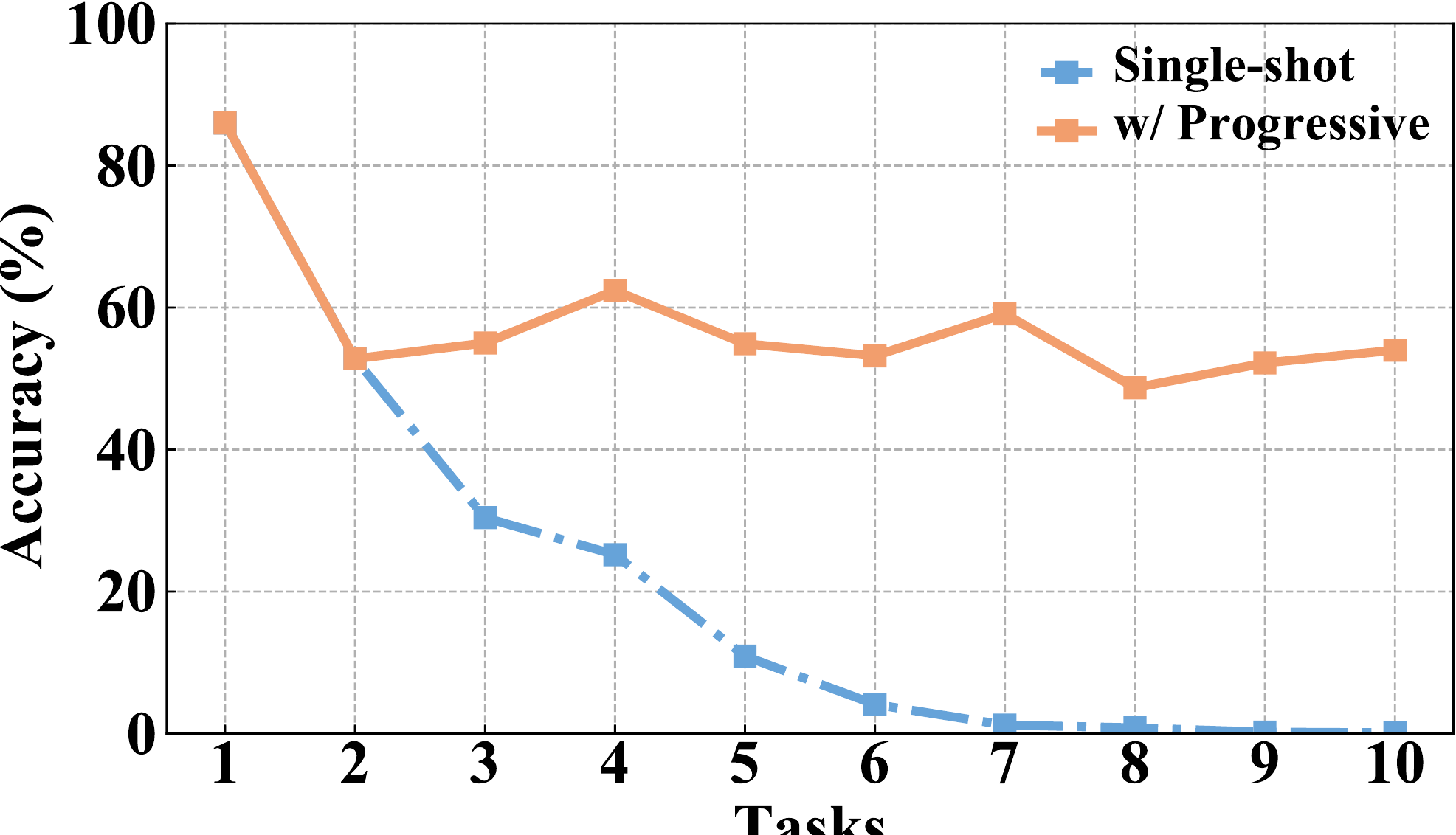}
        \caption{Performance variation on the 1st task for 10-step.}
    \end{subfigure}
    \hspace{.3cm}
   \begin{subfigure}{0.46\linewidth}
       \includegraphics[width=1.5in]{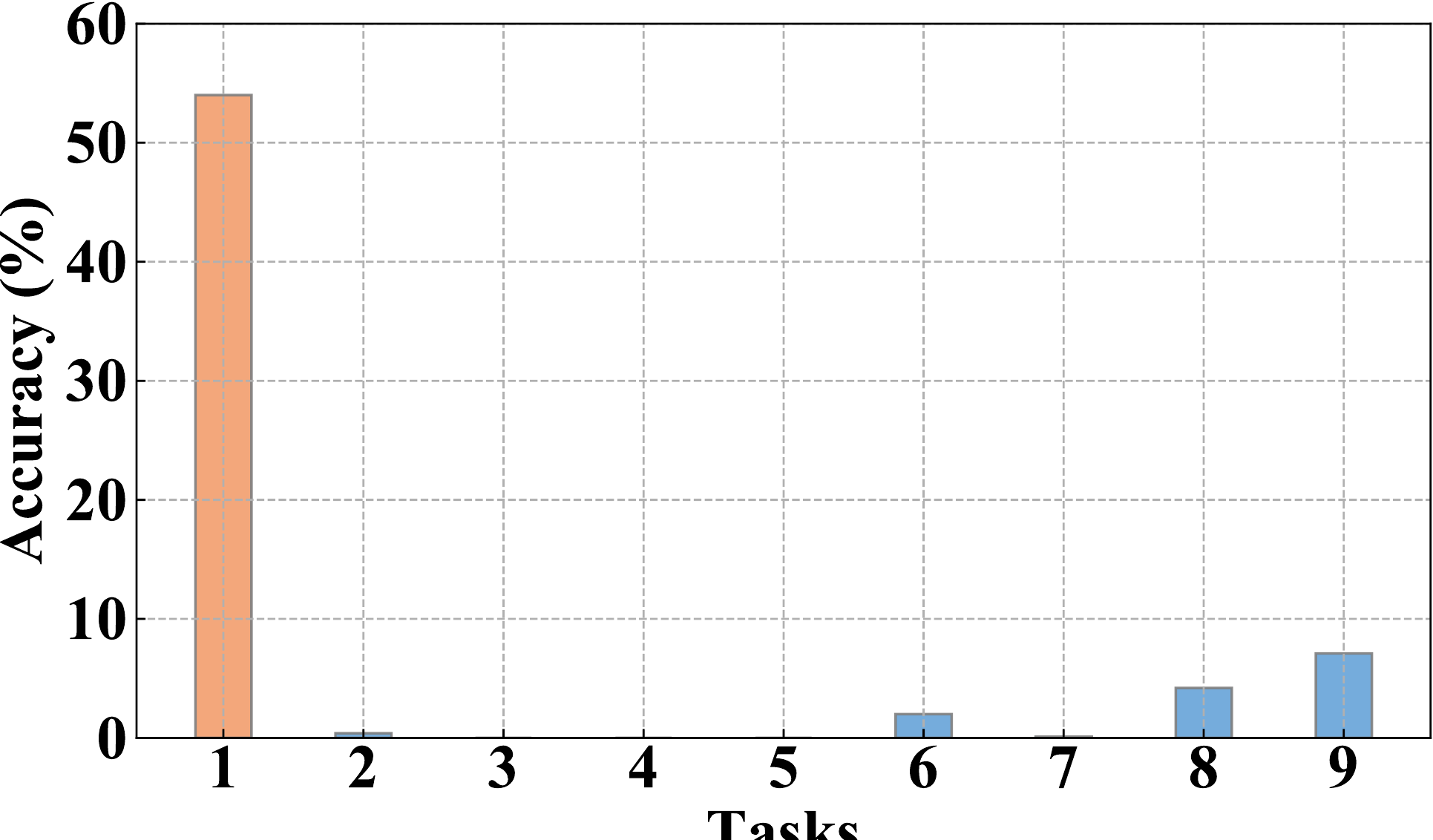}
        \caption{Results of all tasks at $10$th step.}
    \end{subfigure}
    \vspace{-.3cm}
\caption{The effects of adopting the first function in PLwF when trained over 10 tasks on CIFAR-100.}
\vspace{-6mm}
\label{fig:1st_distance}  
\end{figure}

\begin{figure}
   \begin{subfigure}{0.47\linewidth}
       \includegraphics[width=1.6in]{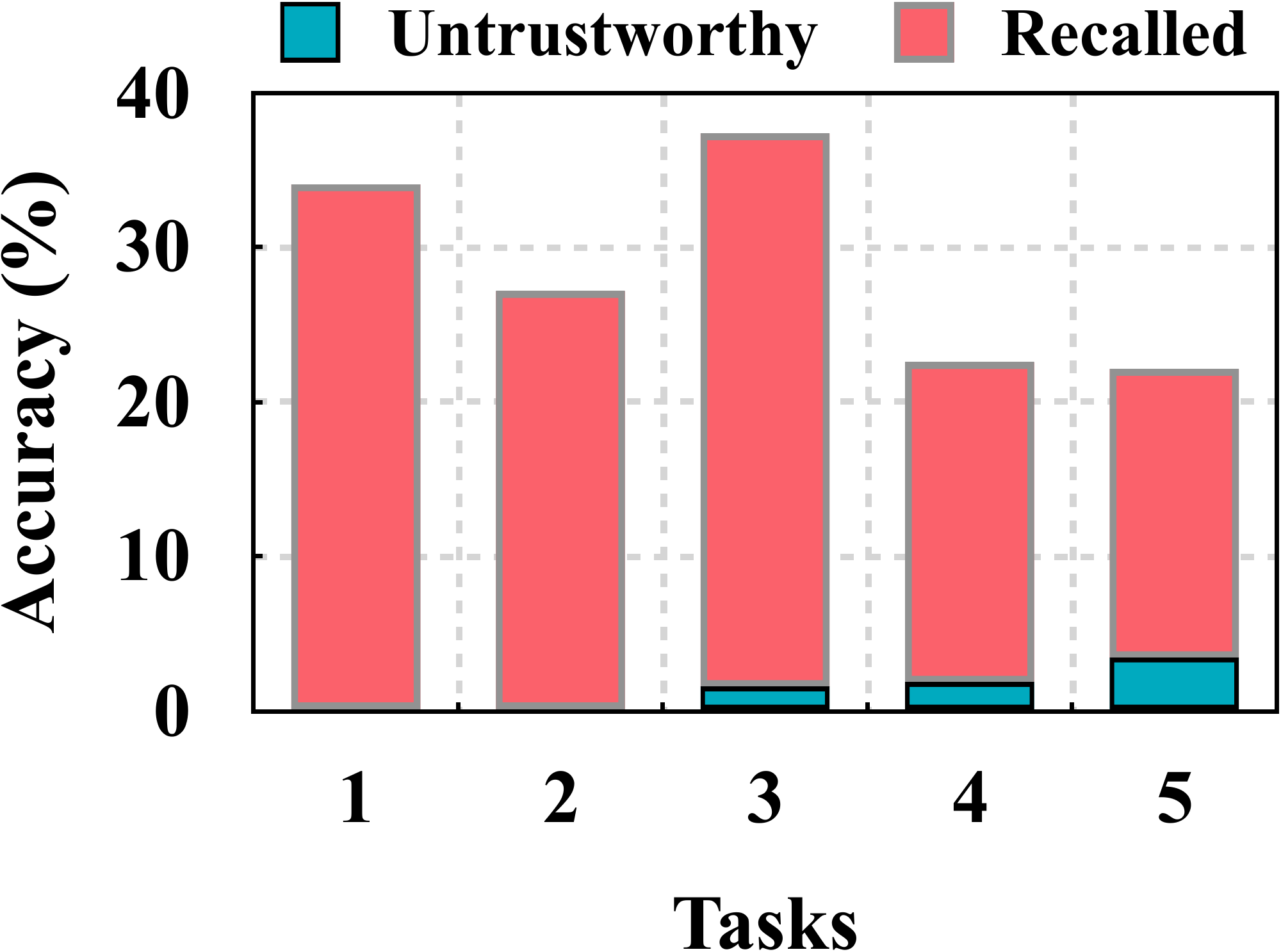}
        \caption{Random order 1}
    \end{subfigure}
    \hspace{.2cm}
   \begin{subfigure}{0.47\linewidth}
       \includegraphics[width=1.6in]{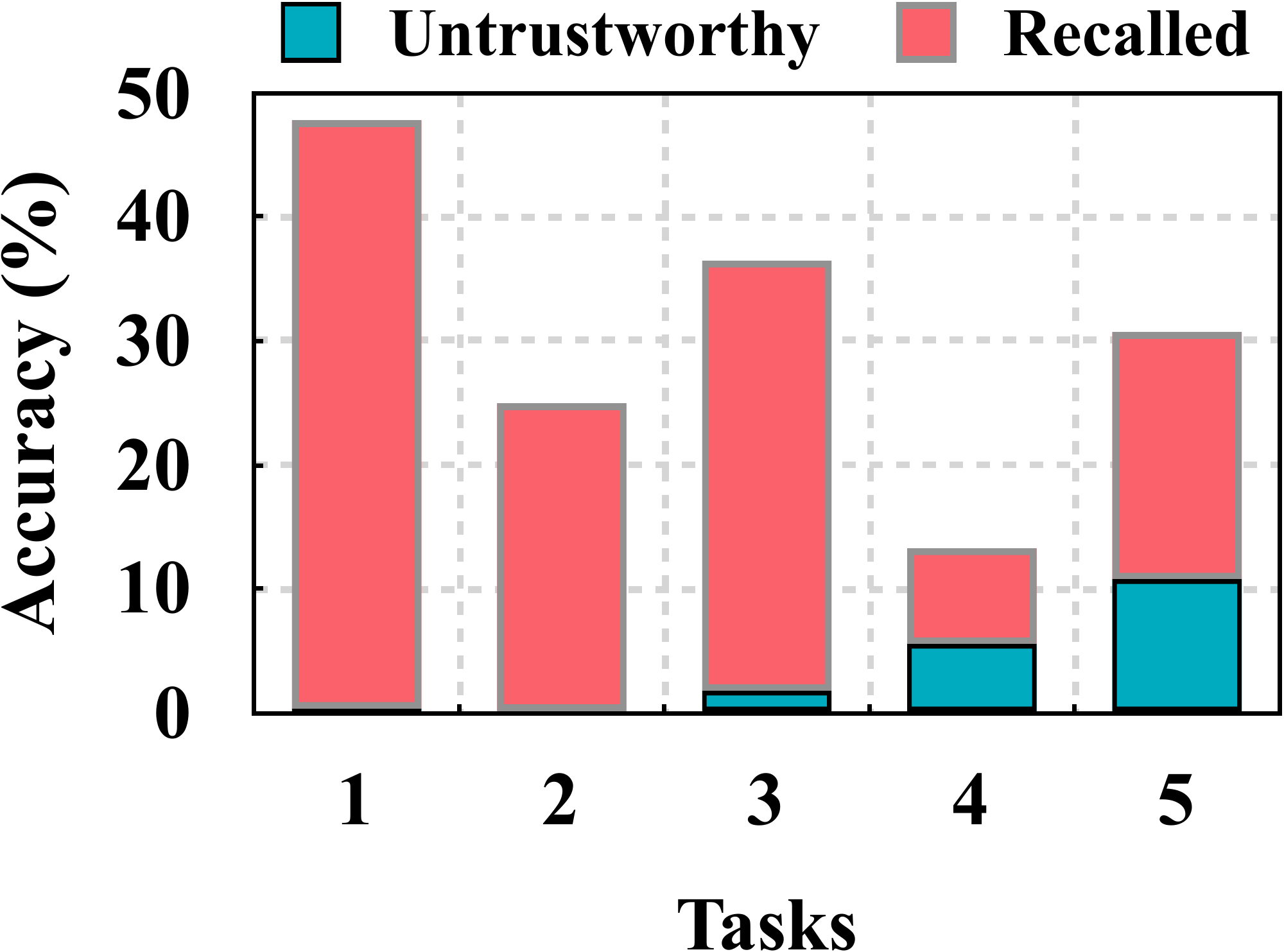}
        \caption{Random order 2}
    \end{subfigure}
    \vspace{-.3cm}
\caption{The examination of order-agnostic behaviour on 10-Split CIFAR-100. `Recalled' indicates improved performance of our method in first 5 tasks. `Untrustworthy' indicates performance of vanilla method (LwF~\cite{DBLP:journals/pami/LiH18a} in first 5 tasks).}
\vspace{-7mm}
\label{fig:ordering}  
\end{figure}

\begin{figure}
   \begin{subfigure}{0.46\linewidth}
       \includegraphics[width=1.5in]{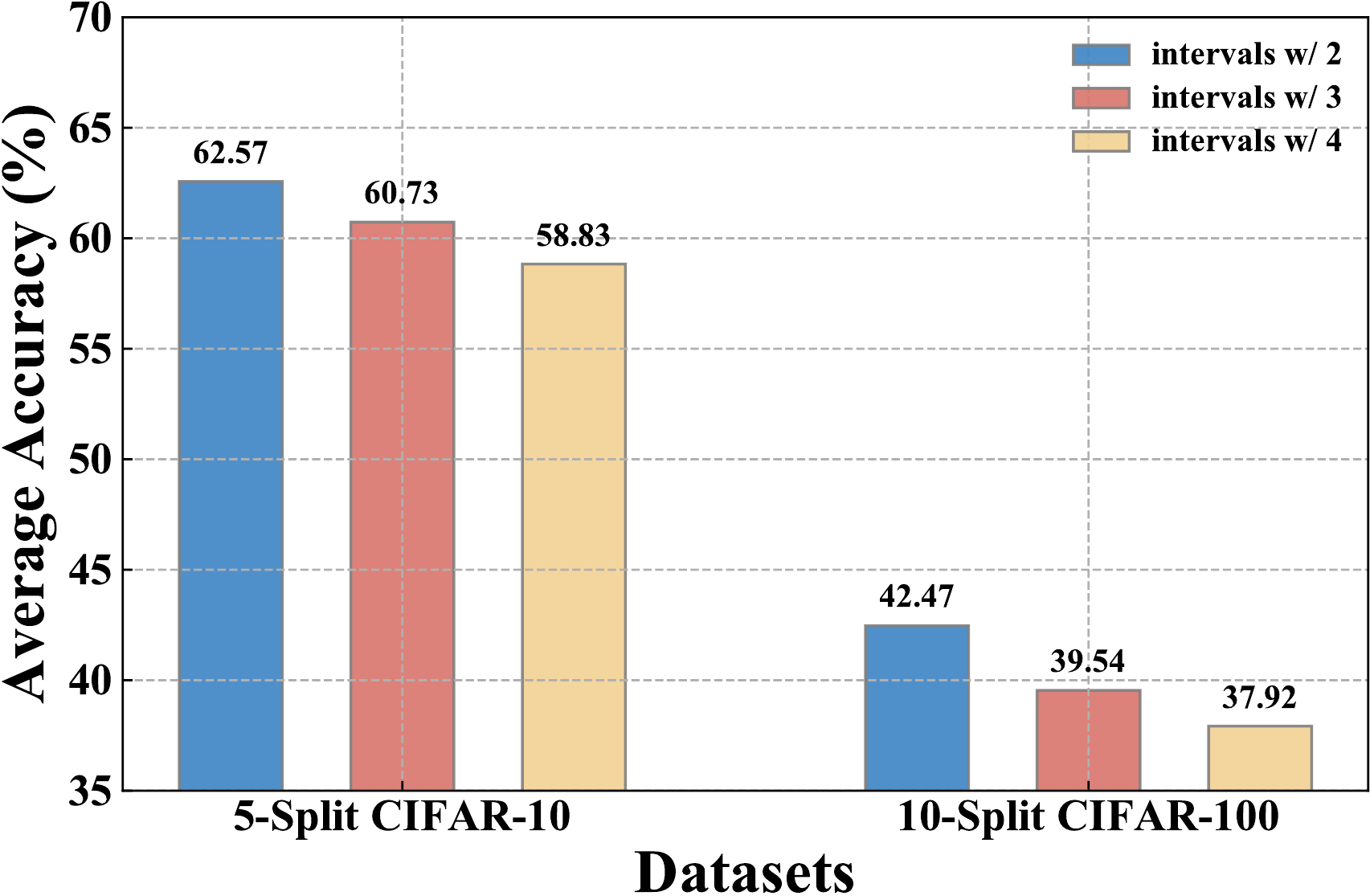}
        \caption{Regular sampling}
        \label{fig:subfigure21}
    \end{subfigure}
    \hspace{.3cm}
   \begin{subfigure}{0.46\linewidth}
       \includegraphics[width=1.5in]{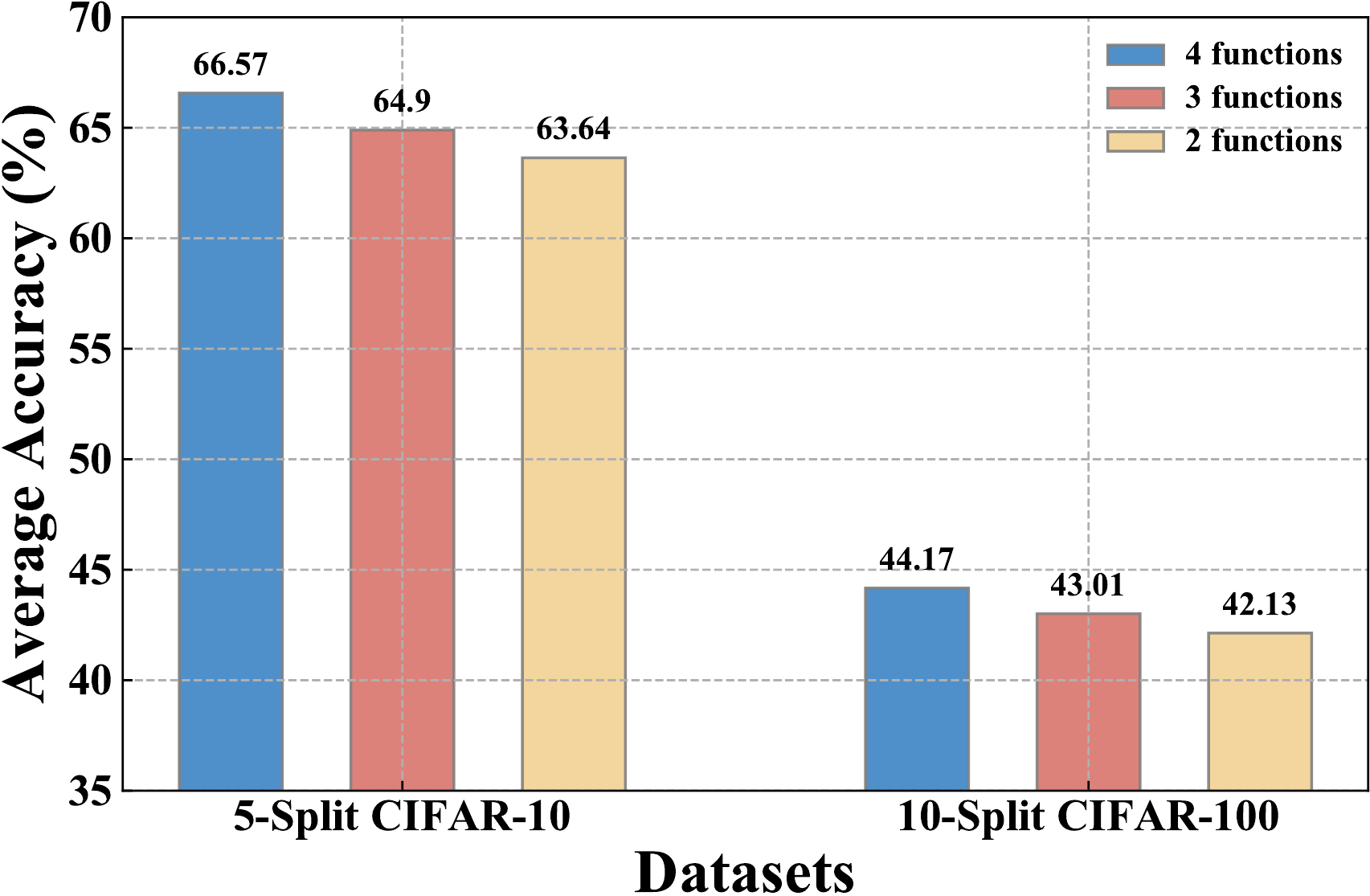}
        \caption{Irregular sampling}
        \label{fig:subfigure22}
    \end{subfigure}
    \vspace{-.2cm}
\caption{Empirical analysis of PLwF using (a) regular sampling with fixed step sizes and (b) irregular sampling to random sample 2/3/4 previous model functions into the knowledge space.}
\vspace{-8mm}
\label{fig:Assumption1}  
\end{figure}

\textbf{Support for PLwF} In \emph{Definition 1}, we propose that there is a space where knowledge disappears less. Firstly, we ablate the influences of the PLwF hypothesis.  Table \ref{table_ablation} reveals that the expansion in learning space significantly improves the task performance. To further examine such a hypothesis, different sampling strategies are used to introduce various function spaces in which the degree of knowledge disappearance varies.
In Figure \ref{fig:Assumption1}, we present results using different sampling strategies. Specifically, regular sampling represents sampling from the function $\mathcal{F}$ by the set of intervals ${2, 3, 4}$. Random sampling represents that 2, 3 and 4 functions are randomly selected from the function $\mathcal{F}$. Intuitively, the function space $\mathcal{F}$ refers to the better knowledge that can be learned when the function of each task is used. This is proved from the best performance shown in Table \ref{table_main} that the function space in this case has the most significant CL ability with the least forgetting. As shown in Figure \ref{fig:Assumption1}, as the space size got reduced under different sampling strategies, the result of CL showed a downward trend, reflecting the disappearance of more knowledge. These phenomena support our hypothesis. That means the space where knowledge disappears less can facilitate PLwF, since it covers a priori knowledge of more accurate label distributions. In contrast, the improvement in learning ability at each task signifies that using the PLwF training model is a good initialization for the subsequent learning. Therefore, a benign promotion is formed to maximize the expectation of accumulating experience. More details about the impact of learning space are given in the Appendix.

\textbf{Support for Credit Assignment Hypothesis.} To test the credit assignment hypothesis proposed in the method section, the experiments are conducted on the 5-Split CIFAR-10, 10-Split CIFAR-100, and 10-Split Tiny-ImageNet. As shown in Table \ref{table_ablation}, on the 5-Step and 10-Step benchmarks, the performance of CL after the incorporation of credit assignment obtains an increase of 1.98\%, 1.78\%, and 3.26\% on Avg and 3.08\%, 3.00\%, and 1.90\% on Last. This indicates that the optimizer is changing to a direction more favorable to the continual learner, and the credit assignment successfully mediates stability and plasticity. 
Moreover, it also shows that the tug-of-war dynamics in each iteration between different tasks can be well captured by the gradient assignment matrix to improve the reliability of the credit assignment. Furthermore, a short sequence of tasks containing only 2 steps, the most fundamental puzzle for the credit assignment regime, is specially designated to observe stability-plasticity. 
In this case, there is only one tug-of-war in which the regime must work to improve upon. 
Therefore, it’s the most essential scenario for the credit assignment. As shown in Table \ref{table_short}, the performance is tangibly improved on each setting, robustly mediating stability and plasticity. In summary, the performance of our method on tasks of sequences with different lengths provides good empirical evidence for the credit assignment assumption that mediates the tug-of-war dynamics.

\begin{table}
\small
\centering
\vspace{-3mm}
\caption{The benefit of credit assignment on different optimizers, which is evaluated on 5-Split CIFAR-10.}
\vspace{-2mm}
\label{table_optimizer}
\begin{tabular}{@{}cccc@{}}
\toprule
\multicolumn{2}{c}{{Optimizer}} &  Avg            & Last           \\ \midrule
\multirow{2}{*}{Adam}        & w/o Credit     & 65.77          & 49.25          \\
                              & w/ Credit       & \textbf{66.78}(\textcolor{forestgreen}{+1.01})   & \textbf{50.16}(\textcolor{forestgreen}{+0.91})   \\ \midrule
\multirow{2}{*}{Adadelta}     & w/o Credit     & 61.76          & 41.79          \\
                              & w/ Credit       & \textbf{62.00}(\textcolor{forestgreen}{+0.24})   & \textbf{42.28}(\textcolor{forestgreen}{+0.49})    \\ \midrule
\multirow{2}{*}{RMSprop}     & w/o Credit     & 47.06          & 24.80          \\
                              & w/ Credit       & \textbf{48.82}(\textcolor{forestgreen}{+1.76})   & \textbf{27.86}(\textcolor{forestgreen}{+3.06})    \\ \bottomrule
\end{tabular}
\vspace{-6mm}
\end{table}

\textbf{The Order-agnostic Behaviour.}
To scrutinize the effect of changing class orders, we define several random orders based on different random seeds to split CIFAR-100, which yields more dynamic and agnostic class distributions. 
Table \ref{table_main_seed} reveals that our method still performs well without involving any old data.
Moreover, as shown in Figure \ref{fig:ordering}, in this case, our method remains effective in controlling forgetting (Red column). Overall, the impact of the task order seems insignificant.


\textbf{Generality of Credit Assignment.} 
We analyze the generality of credit assignment from the use of different optimzers and from its usefulness on previous CL methods. 
Firstly, we analyze the performance of credit assignments on other optimizers (Adam~\cite{DBLP:journals/corr/KingmaB14}, Adadetla~\cite{DBLP:journals/corr/abs-1212-5701}, and RMSprop) apart from SGD.
We obtain three observations from Table \ref{table_optimizer}: \textbf{(i)} the tug-of-war dynamics occurs commonly in gradient-based optimization methods;
\textbf{(ii)} the credit assignment regime shows tangible improvement for continuous learners by alleviating tug-of-war dynamics; 
\textbf{(iii)} the credit assignment regime can work together with widely-adopted gradient-based optimizers;  
Secondly, we adapt credit assignment to previous methods like EWC~\cite{kirkpatrick2017overcoming}.
Remarkably, as shown in Table \ref{table_EWC}, we observe an incredible improvement by equipping the credit assignment on EWC (+2.98\% on Avg, +8.50\% on Last).
Combining these observations, we conclude that recreating carefully the tug-of-war dynamics is valuable for CL and helps us to better understand CL.


\section{Conclusion}
In this paper, we propose PLwF, which densely introduces previous functions, creating a learning space with less vanishing knowledge. Building on the newly-constructed space, we further minimize forgetting by establishing the credit assignment regime to recreate the tug-of-war dynamics when learning new tasks.
We show PLwF retains faithful knowledge while requiring neither old samples nor elaborate construction of the old task subspace.



{\small
\bibliographystyle{ieee_fullname}
\bibliography{refs}
}

\clearpage
\section*{Supplementary Material}

In this supplementary material, we first provide more observations about the credibility of the functions based on different methods (Appendix A.1). 
Next, we provide more intuitive explanations of PLwF from different previous stages (Appendix A.2). 
Then, we present detailed discussion about limitations (Appendix A.3). And we provide more versions of relaxed PLwF (Appendix A.4). 
Moreover, we analyze the GPU occupation of PLwF (Appendix A.5).
Then, we provide additional experiments and discuss the influence brought by the order-agnostic behaviour and more details about main results (Appendix A.6 and A.8).
Next, we present more discussions about the generality of credit assignment (Appendix A.7). 
Finally, we discuss trendy directions in Raw-Data-Free methods (Appendix A.9).
Code will be publicly available upon publication.


\subsection*{A.1 More observations about the credibility of the functions} 
\label{A1}
In this subsection, we conduct more extensive experiments on EWC~\cite{kirkpatrick2017overcoming} and iCaRL~\cite{rebuffi2017icarl}, aiming to prove that the problem of fading credibility of a given function is prevailing in regularization-based and memory-based methods.
This observation motivates PLwF.

\begin{figure}[h]
    \centering
   \begin{subfigure}{0.3\textwidth}
       \includegraphics[width=1.8in]{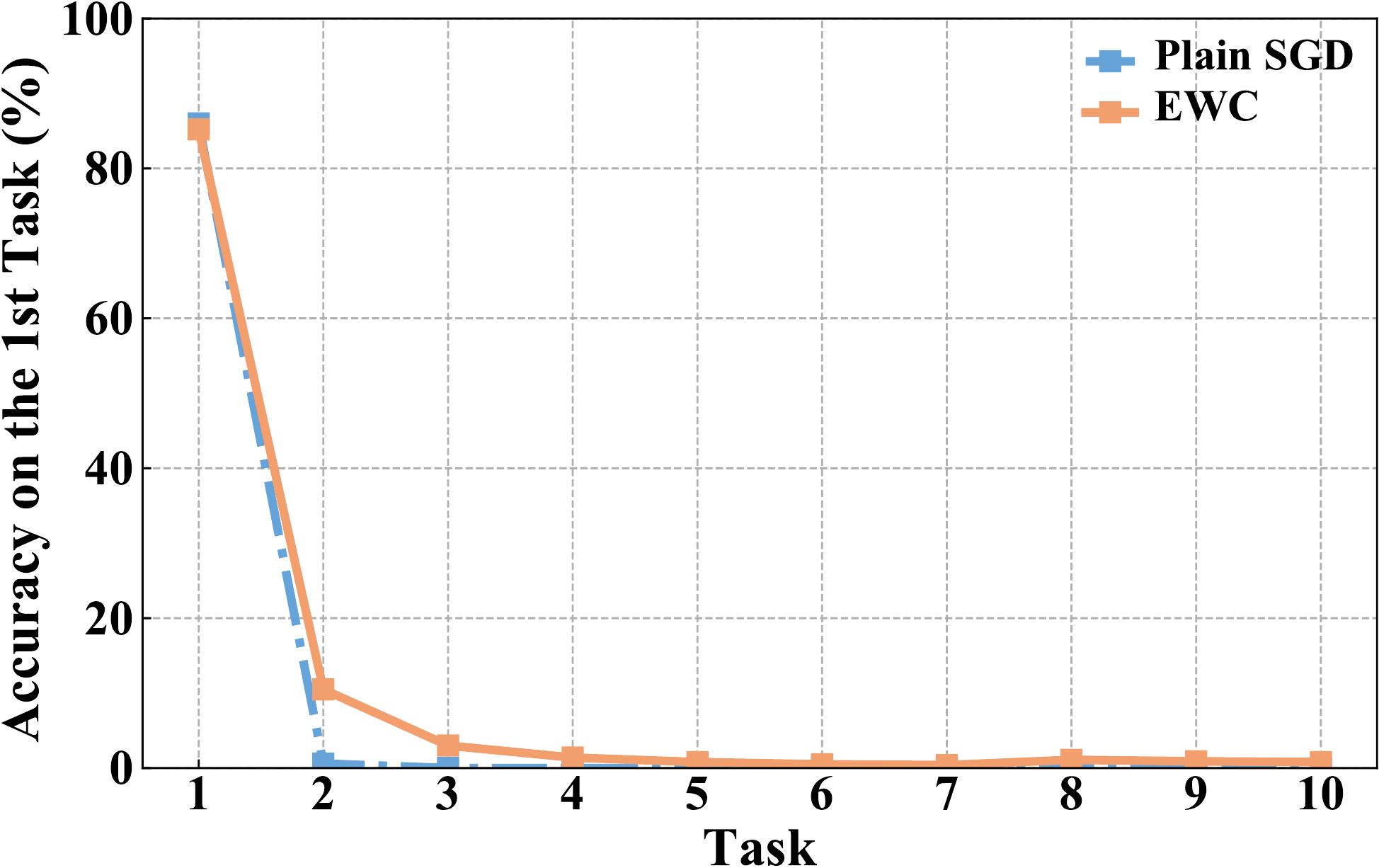}
        \caption{EWC}
        \label{fig:1st_ewc}
    \end{subfigure}
   \begin{subfigure}{0.3\textwidth}
       \includegraphics[width=1.8in]{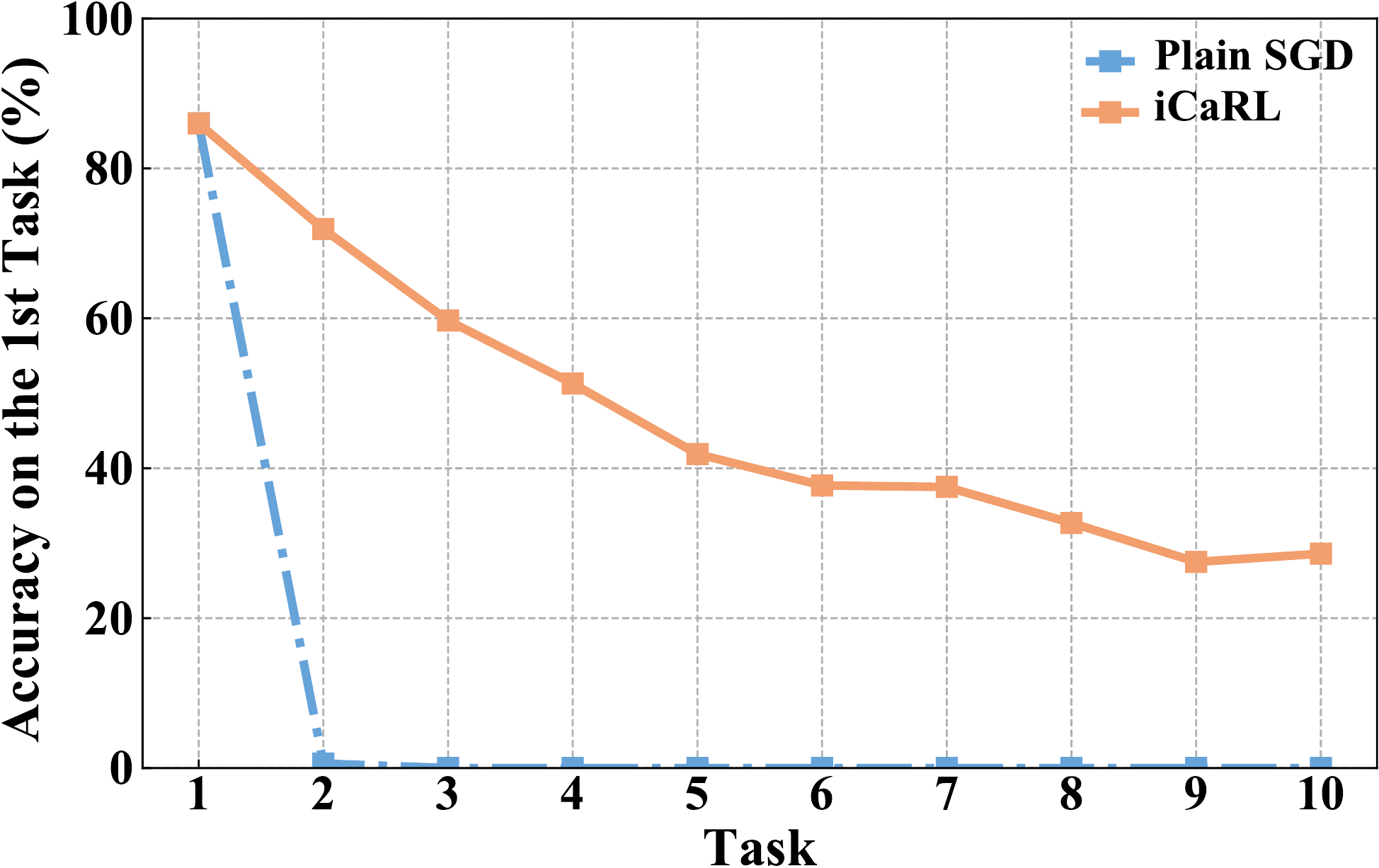}
        \caption{iCaRL}
        \label{fig:1st_icarl}
    \end{subfigure}
\caption{Performance variation of EWC and iCaRL on the first task when trained over 10 tasks on CIFAR-100. Plain SGD demonstrates utterly forgetting due to the standard learning protocol.}
\label{fig:1st_ewc_icarl}  
\end{figure}

Figure \ref{fig:1st_ewc_icarl} shows how a single function can lose credibility. 
We observe that even though EWC~\cite{kirkpatrick2017overcoming} and iCaRL~\cite{rebuffi2017icarl} prevent forgetting to different extent based on different methods, they still show the same trends: \emph{as incremental learning proceeds, the function forgets more and more about the stages they should be responsible for}. 
As shown in Figure \ref{fig:1st_ewc_icarl}, although EWC and iCaRL retain more superior performance (Orange line) compared to Plain-SGD~\cite{DBLP:conf/icml/Zinkevich03} (Blue line), forgetting is still accumulating. Compared to iCaRL, EWC is the worse one. This indicates that the function becomes untrustworthy for the first task as the sequence of tasks increases.
(Although we exemplify our idea using the first task, other tasks can show similar trends.)

\begin{figure}[h]
    \centering
   \begin{subfigure}{0.3\textwidth}
       \includegraphics[width=1.8in]{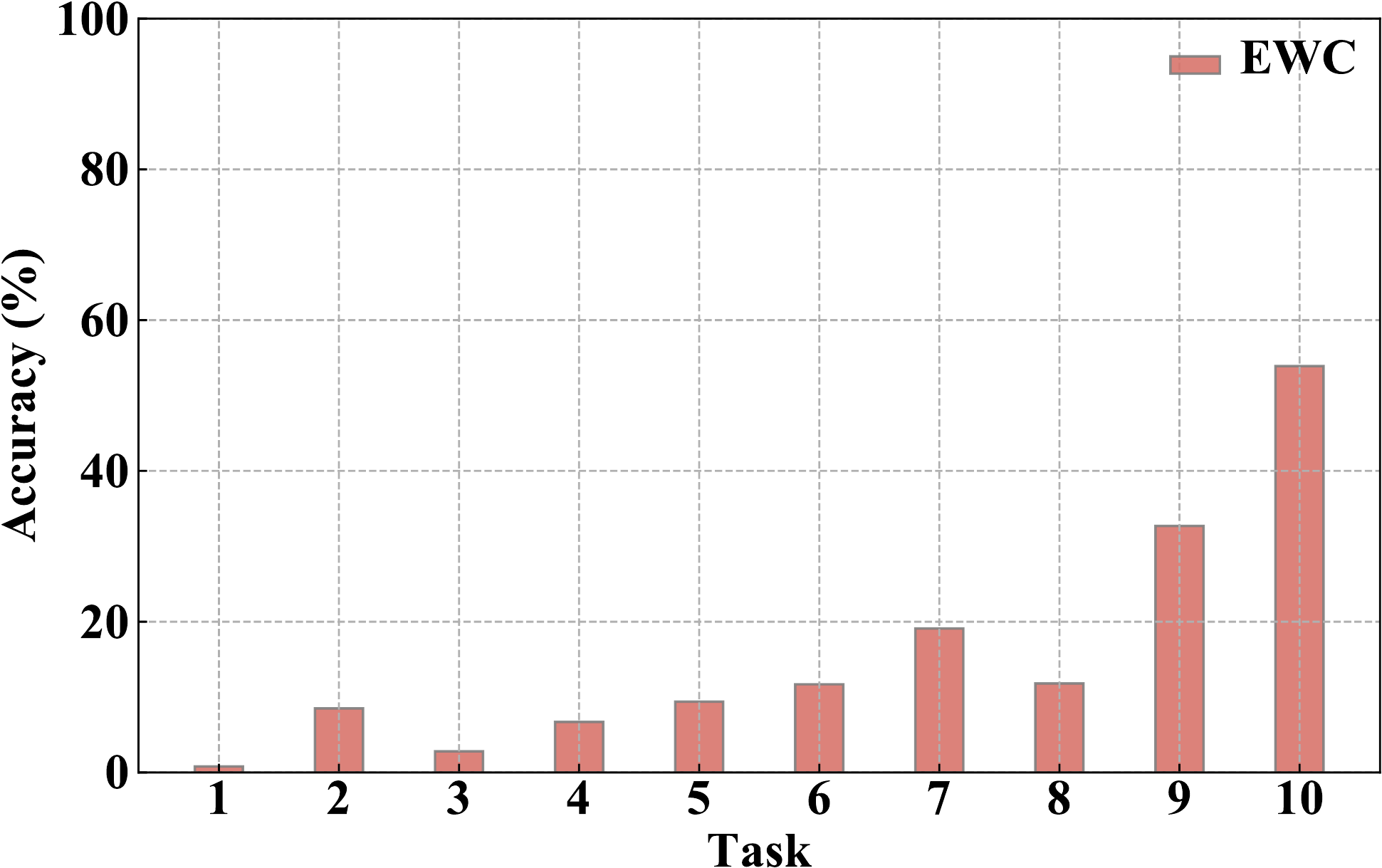}
        \caption{EWC}
        \label{fig:all_ewc}
    \end{subfigure}
   \begin{subfigure}{0.3\textwidth}
       \includegraphics[width=1.8in]{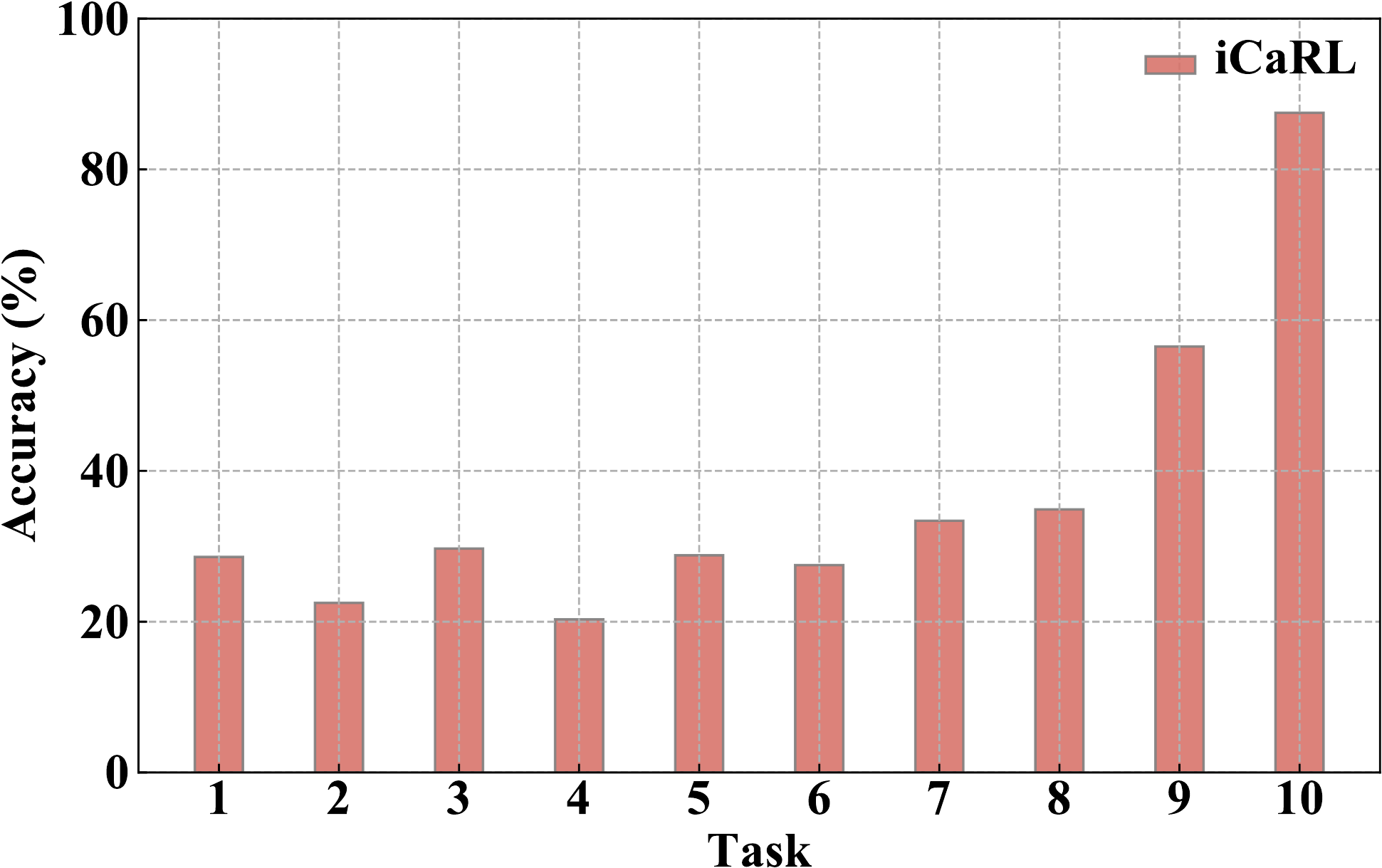}
        \caption{iCaRL}
        \label{fig:all_icarl}
    \end{subfigure}
\caption{Performance variation of EWC and iCaRL on each task when trained over 10 tasks on CIFAR-100.}
\label{fig:all_ewc_icarl}  
\end{figure}

Figure \ref{fig:all_ewc_icarl} reveals the relationship between the reliability of the function at the early task and at the current task. 
As shown in Figure \ref{fig:all_ewc_icarl}, we observe a trend that more performance is retained for the function closer to the current task (task 10 in Figure \ref{fig:all_ewc_icarl} (a) and Figure \ref{fig:all_ewc_icarl} (b)), \emph{e.g.} the performance of function from task 9 (nearest to the task 10) is the least forgotten. 
Conversely, less performance is retained for the function further away from the current task, \emph{e.g.} the performance of function from task 1 (farthest to the task 10) is the most forgotten. 
This indicates that the function of the early task gradually loses its reliability.

Based on these observations, we can conclude that changes of the knowledge space caused by the fading reliability of the function is an issue that needs to be carefully considered during CL. This provides more support for the proposed PLwF method.

\begin{figure*}
    \centering
   \begin{subfigure}{0.3\textwidth}
       \includegraphics[width=1.8in]{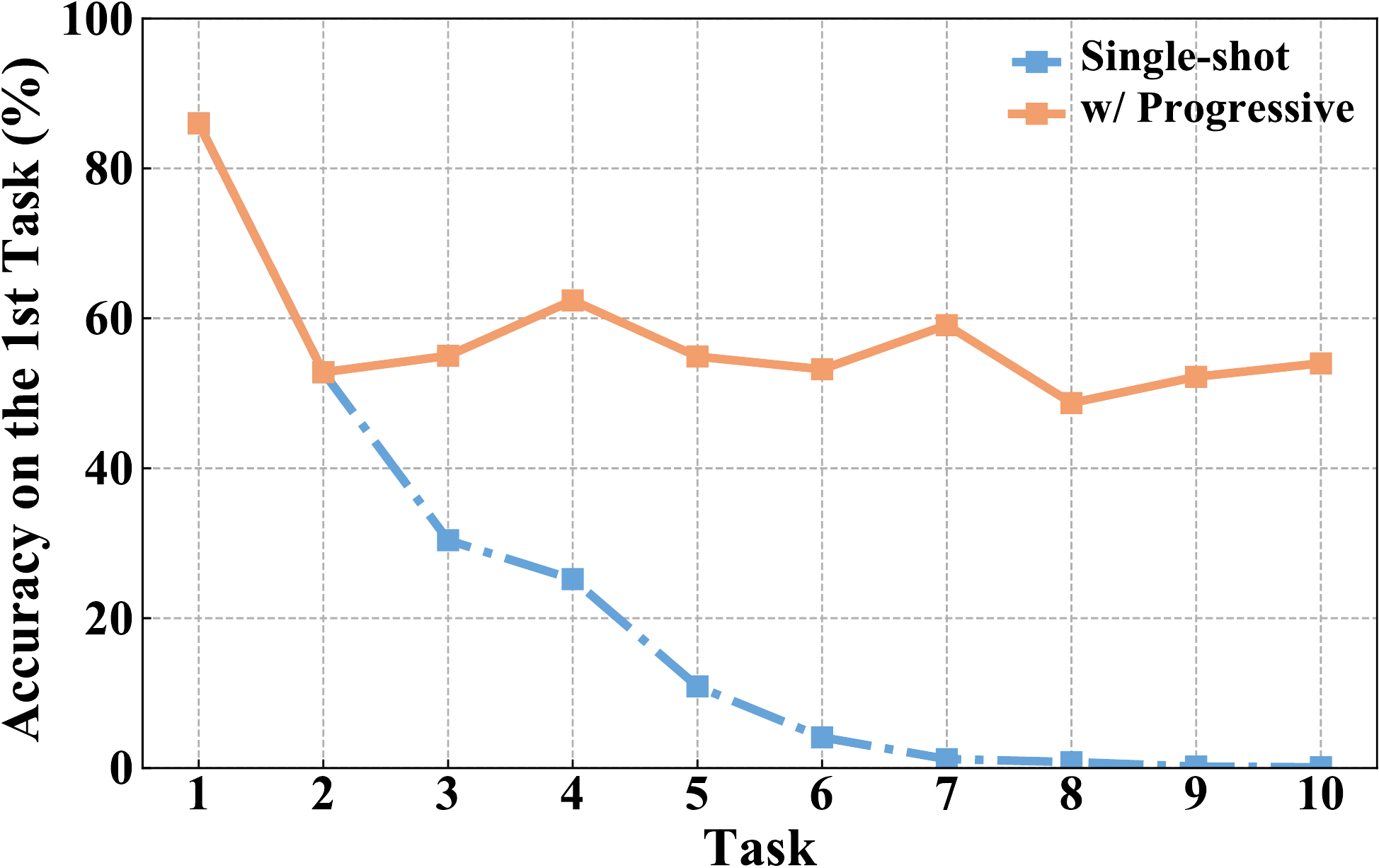}
        \caption{Performance variation on the 1st task if only matching the function from the 1st task.}
    \end{subfigure}
    \hspace{2cm}
   \begin{subfigure}{0.3\textwidth}
       \includegraphics[width=1.8in]{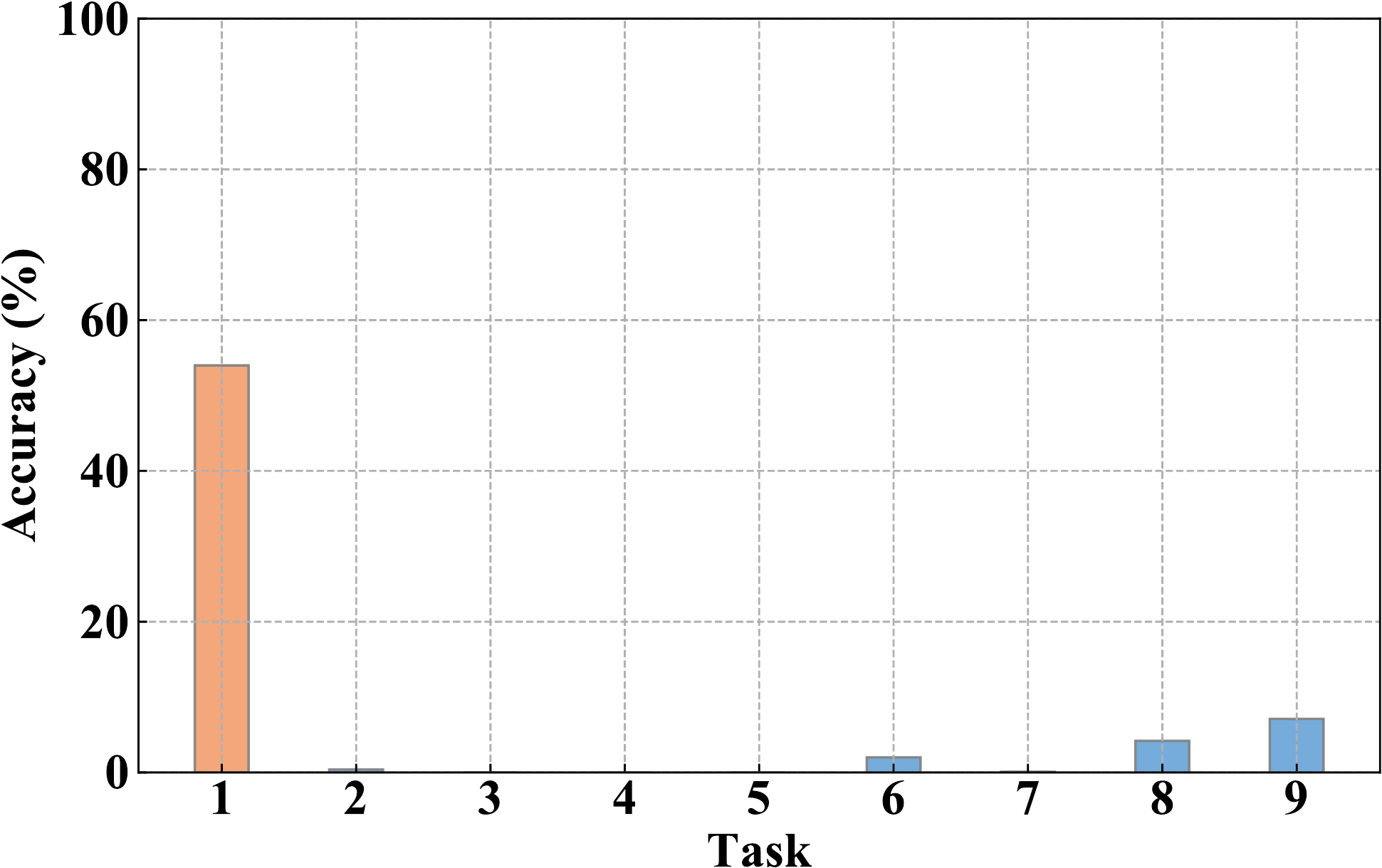}
        \caption{Results of all tasks using PLwF if only matching the function from the 1st task.}
    \end{subfigure}
   \begin{subfigure}{0.3\textwidth}
       \includegraphics[width=1.8in]{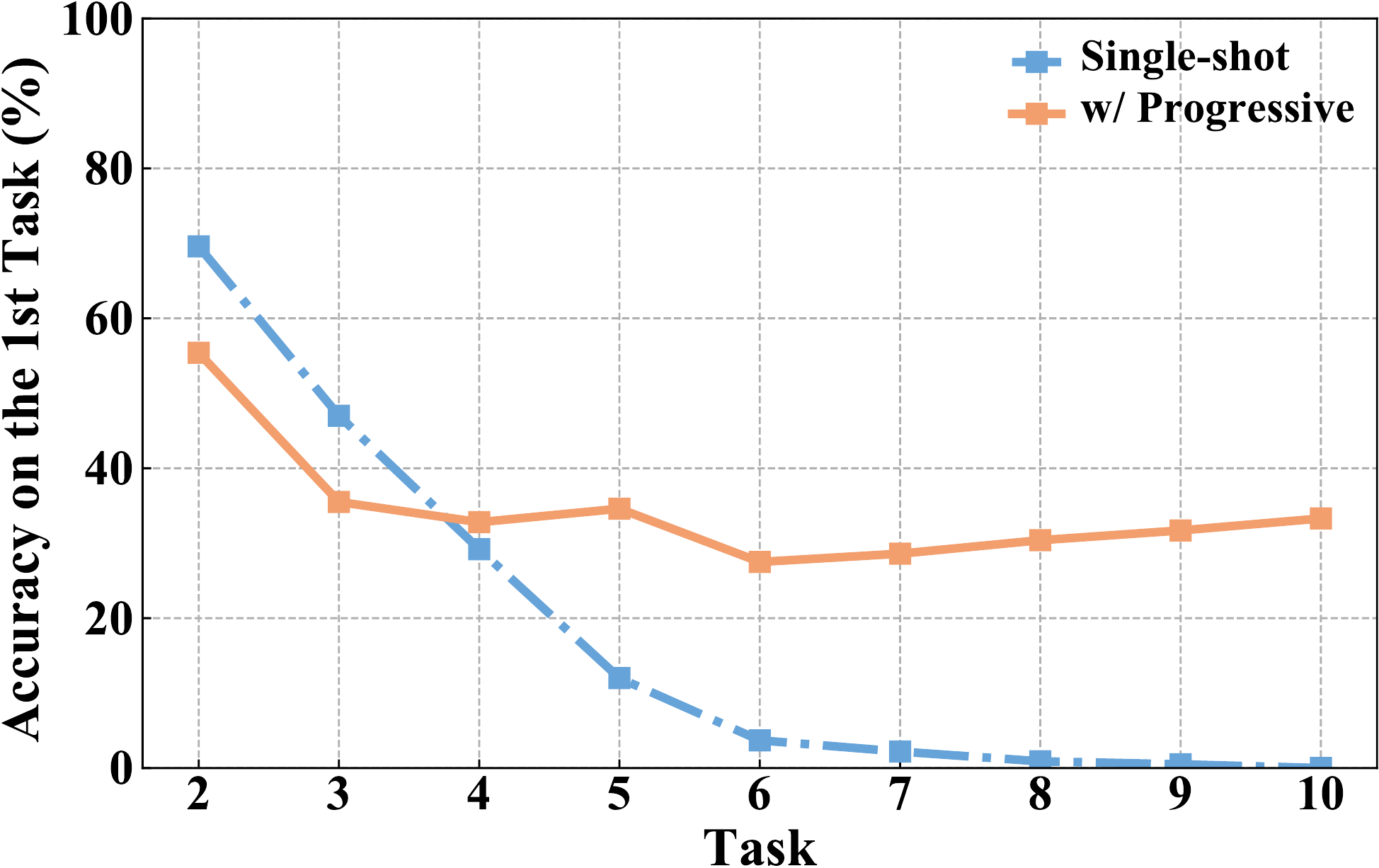}
        \caption{Performance variation on the 2nd task if only matching the function from the 2nd task.}
    \end{subfigure}
    \hspace{2cm}
   \begin{subfigure}{0.3\textwidth}
       \includegraphics[width=1.8in]{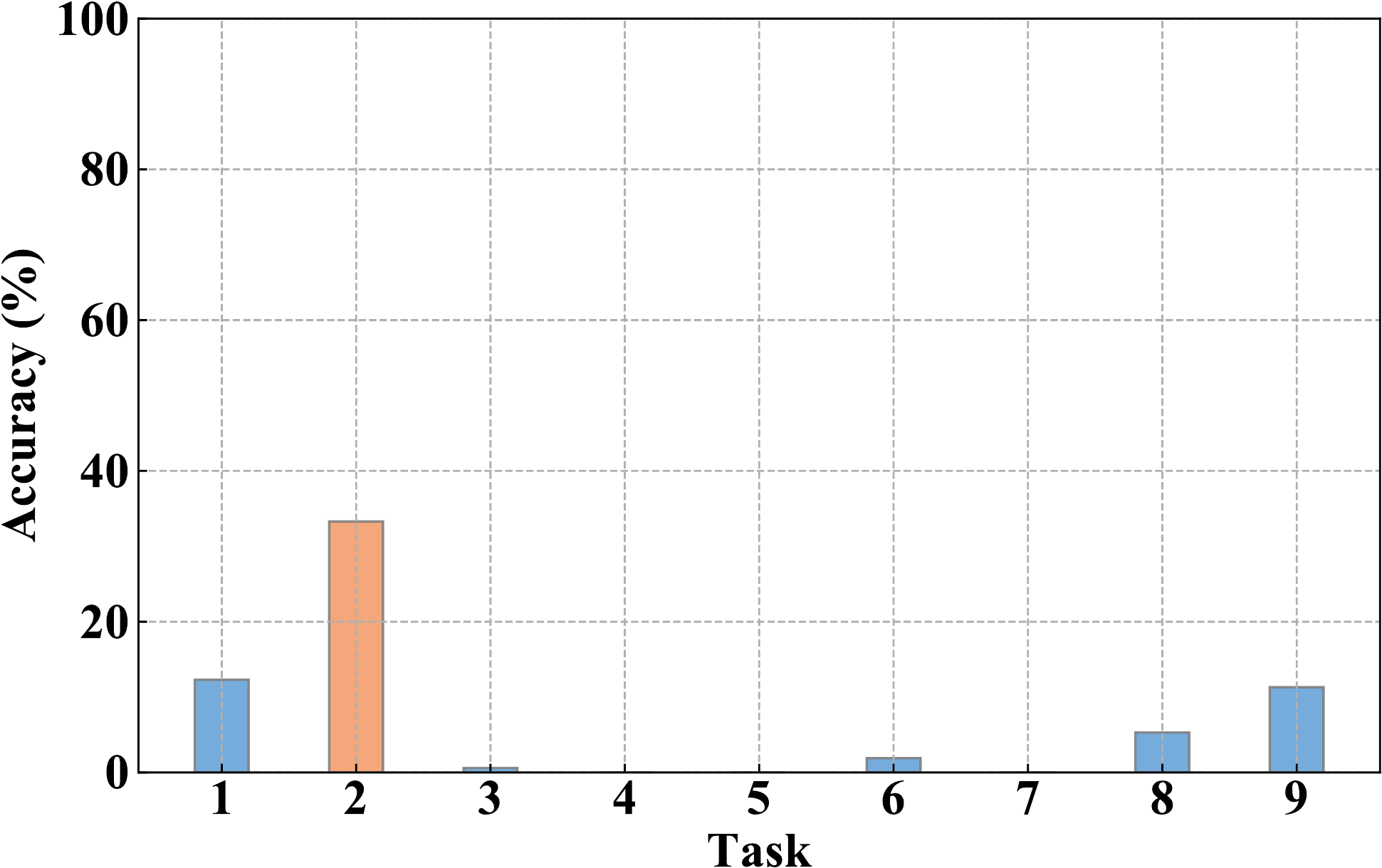}
        \caption{Results of all tasks using PLwF when matching the function from the 2nd task.}
    \end{subfigure}
   \begin{subfigure}{0.3\textwidth}
       \includegraphics[width=1.8in]{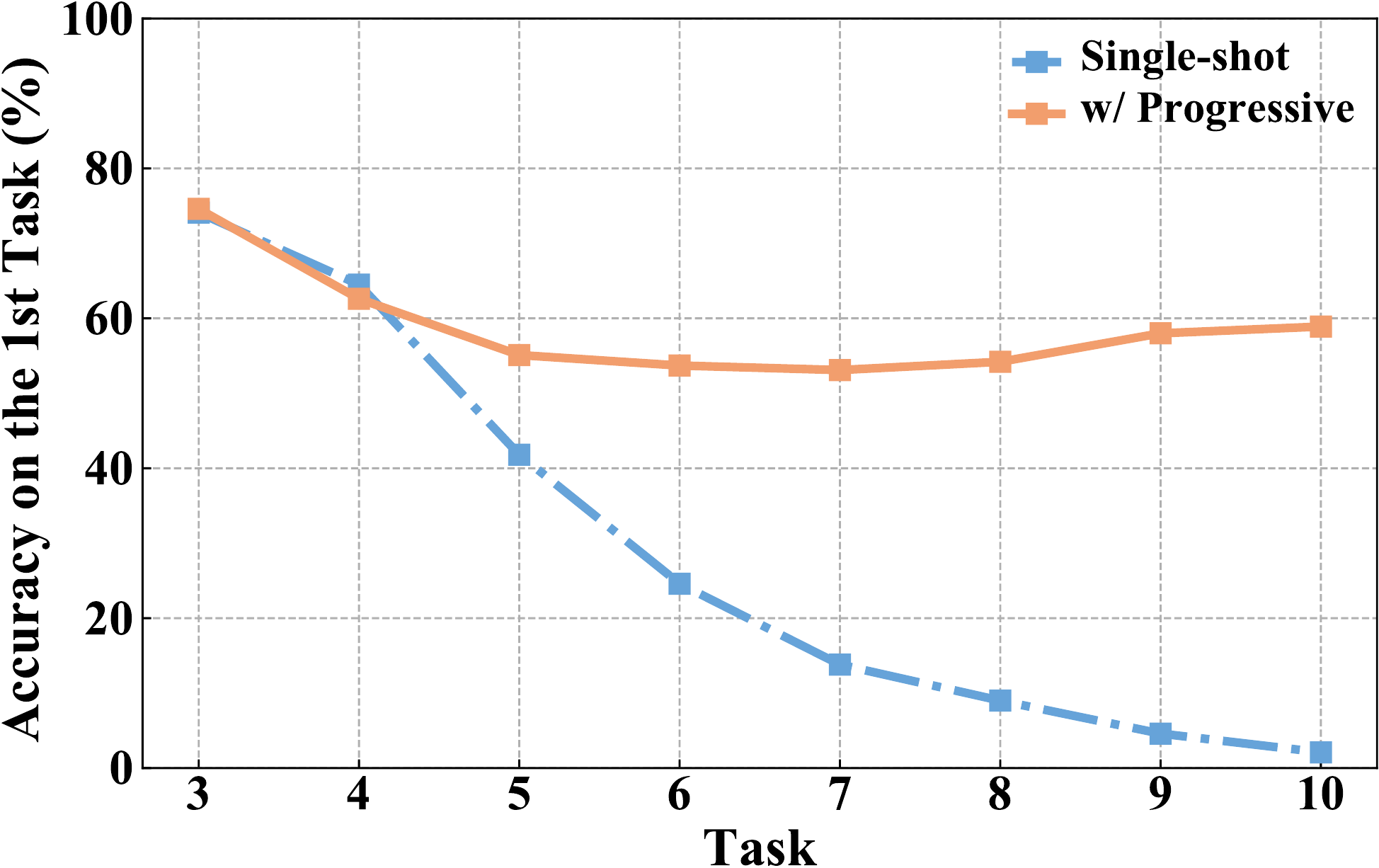}
        \caption{Performance variation on the 3rd task if only matching the function from the 3rd task.}
    \end{subfigure}
    \hspace{2cm}
   \begin{subfigure}{0.3\textwidth}
       \includegraphics[width=1.8in]{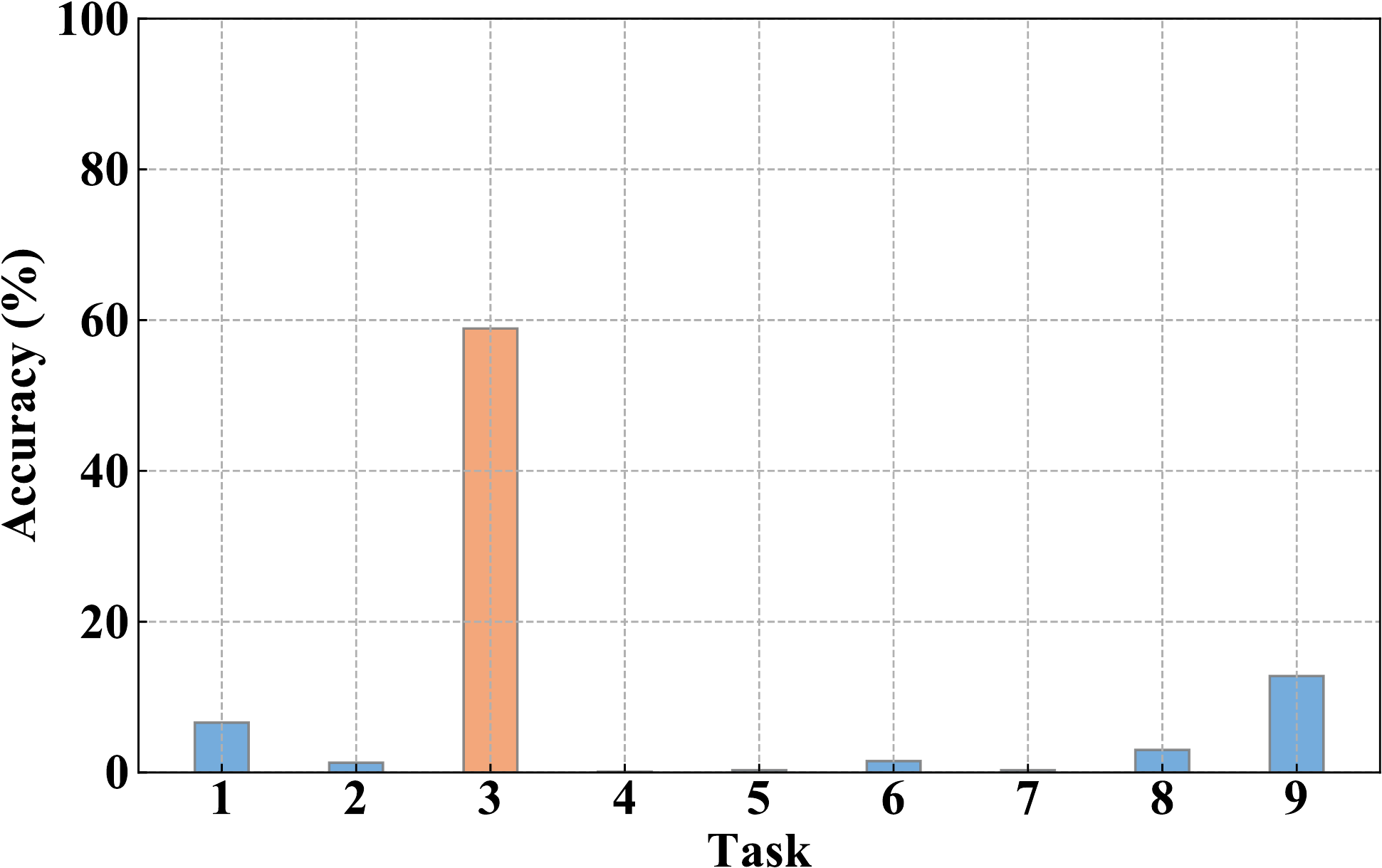}
        \caption{Results of all tasks using PLwF when matching the function from the 3rd task.}
    \end{subfigure}
   \begin{subfigure}{0.3\textwidth}
       \includegraphics[width=1.8in]{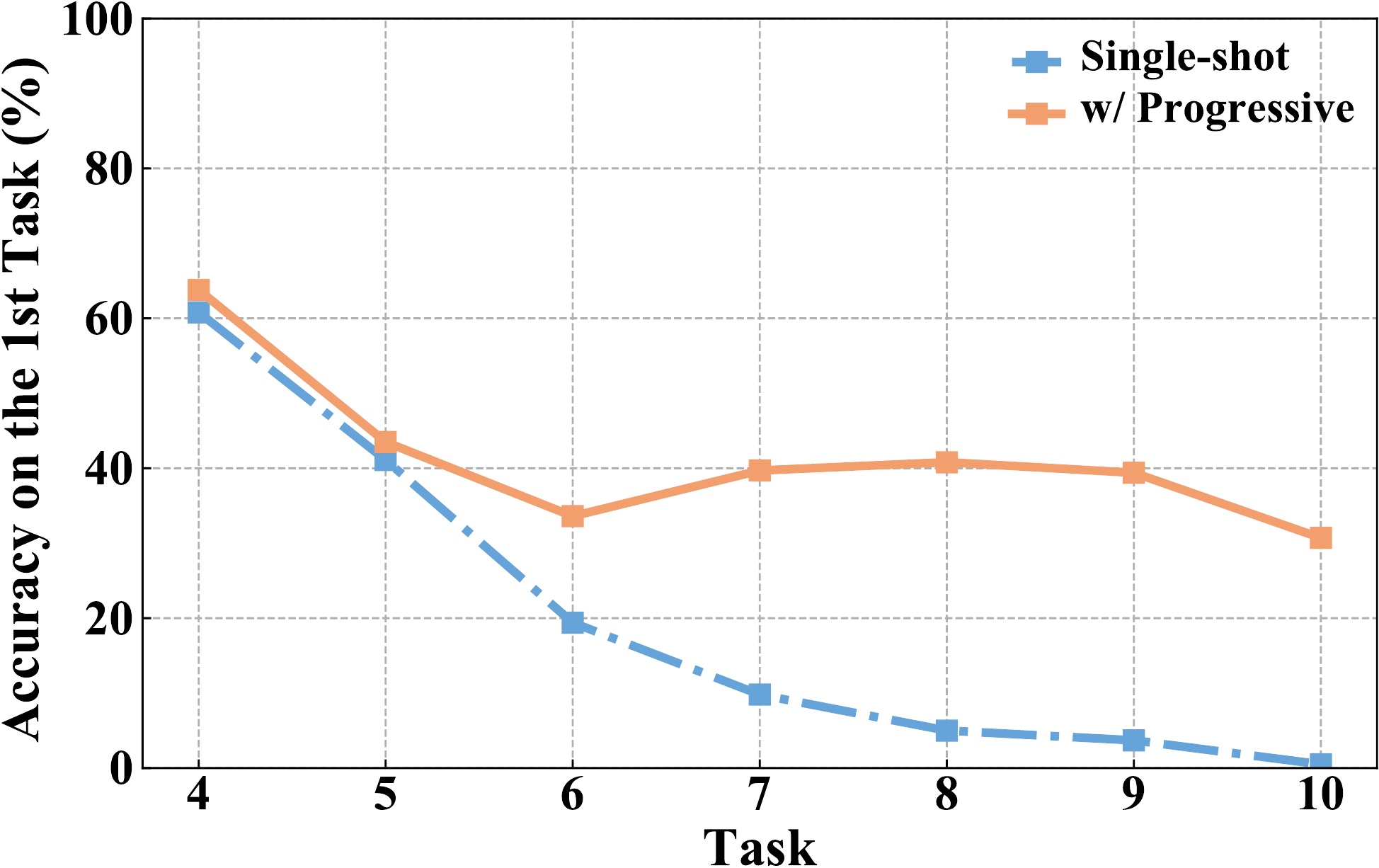}
        \caption{Performance variation on the 4th task if only matching the function from the 4th task.}
    \end{subfigure}
    \hspace{2cm}
   \begin{subfigure}{0.3\textwidth}
       \includegraphics[width=1.8in]{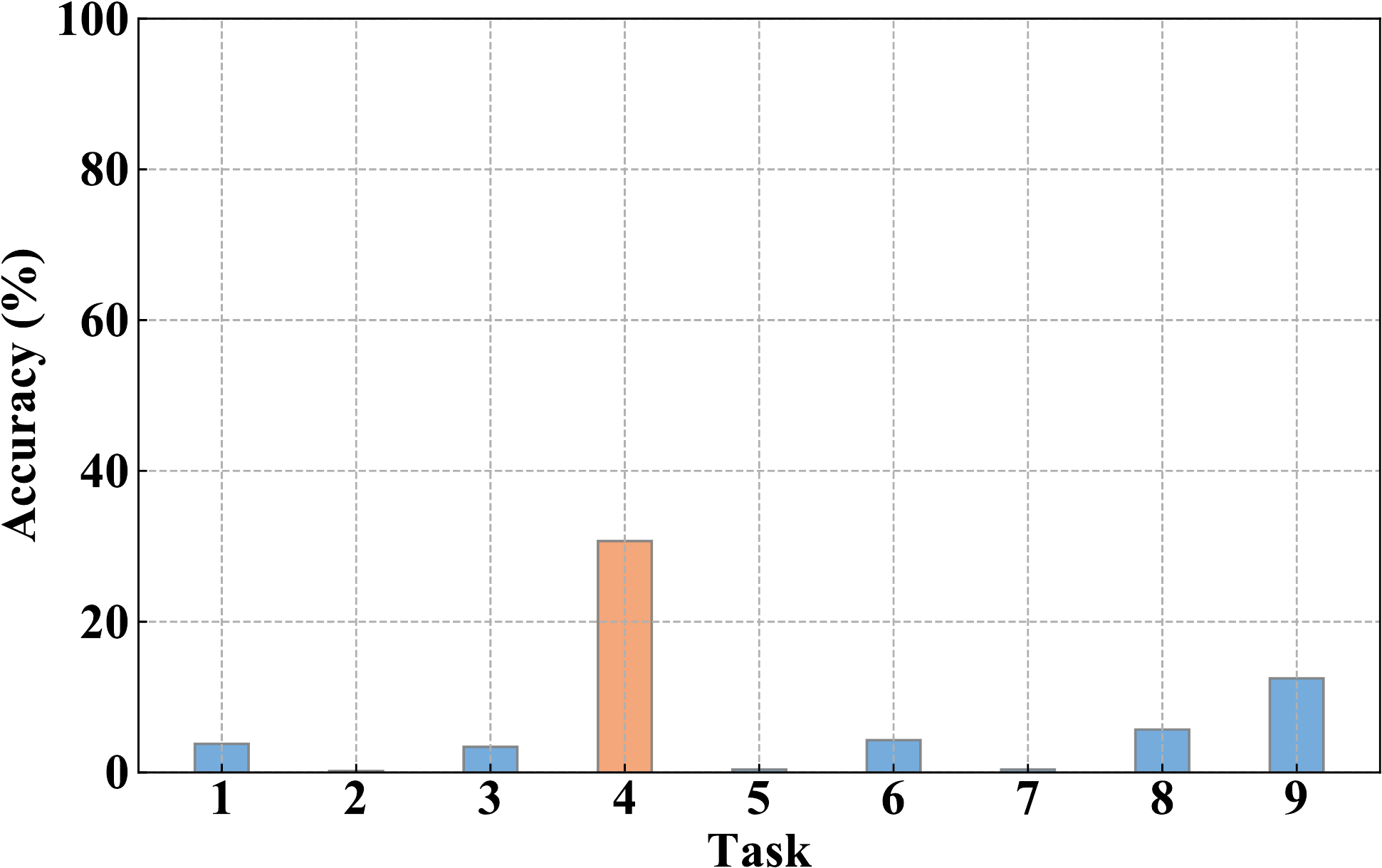}
        \caption{Results of all tasks using PLwF when matching the function from the 4th task.}
    \end{subfigure}
   \begin{subfigure}{0.3\textwidth}
       \includegraphics[width=1.8in]{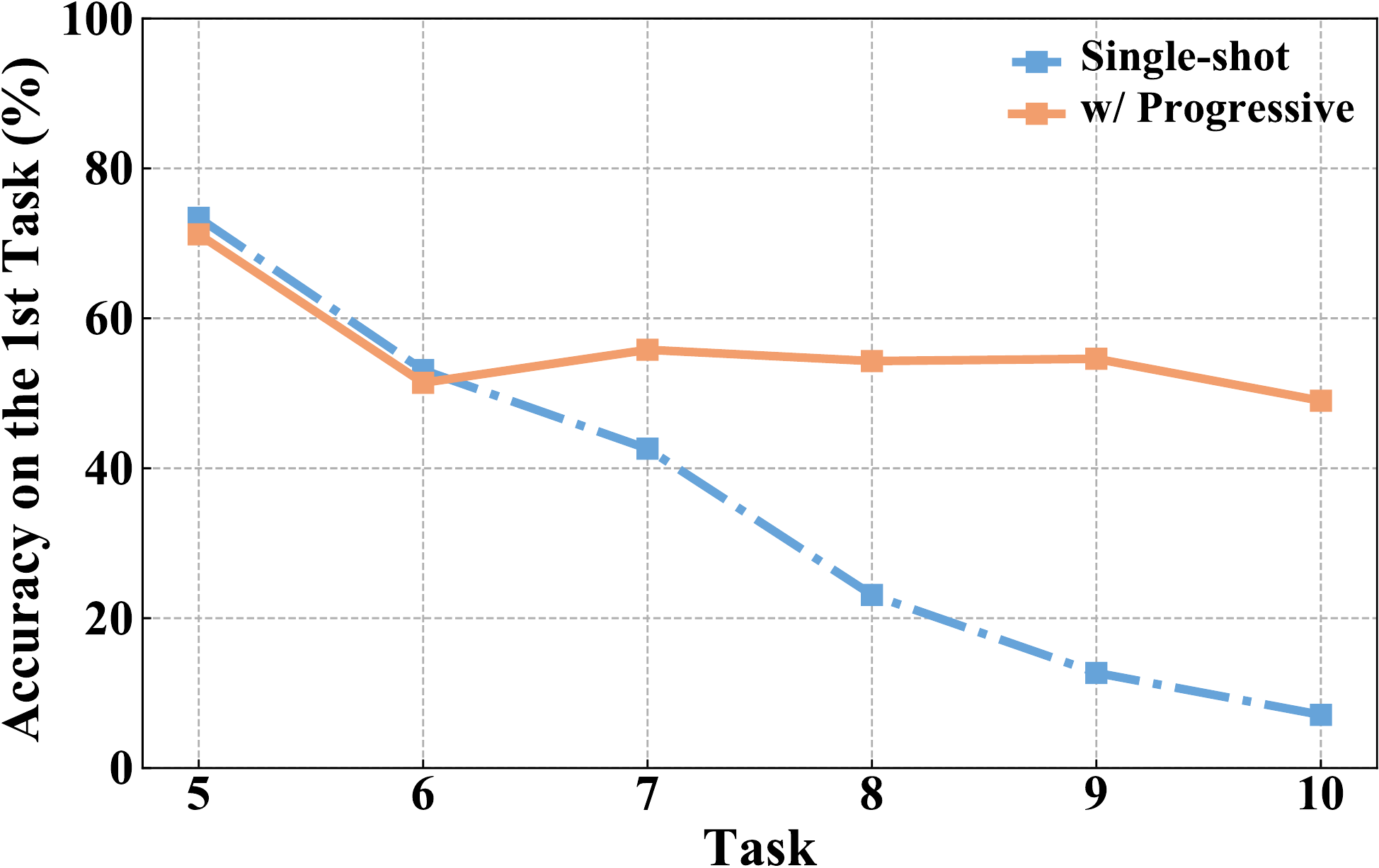}
        \caption{Performance variation on the 5th task if only matching the function from the 5th task.}
    \end{subfigure}
    \hspace{2cm}
   \begin{subfigure}{0.3\textwidth}
       \includegraphics[width=1.8in]{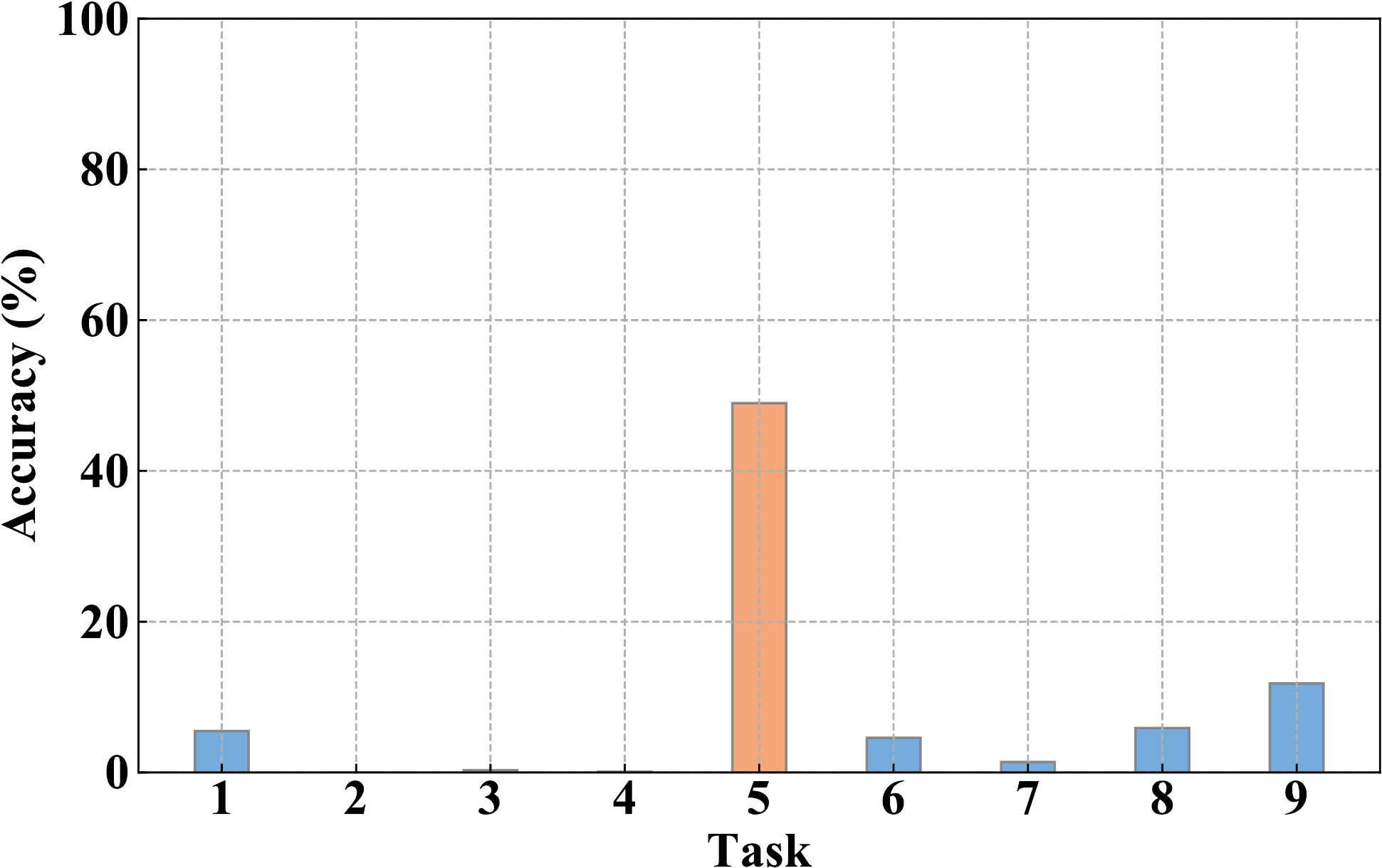}
        \caption{Results of all tasks using PLwF when matching the function from the 5th task.}
    \end{subfigure}
\caption{The effects of function from different distances in PLwF when trained over 10 tasks on CIFAR-100.}
\label{fig:1st_single}  
\end{figure*}

\subsection*{A.2 More intuitive explanations of PLwF}
\label{A2}

As shown in Figure \ref{fig:1st_single}, to examine the benefit of adapting previous functions, we observe the effect of matching functions from different distances in PLwF. We specifically use only one function from the early 5 tasks as a matching function to perform CL. 
Figure \ref{fig:1st_single} reveals the following observations: \textbf{(i.)} In Figure \ref{fig:1st_single} (a, c, e, g, i), when one function from early task is applied as a matching function, the accumulation of forgetting at that task decreases significantly as learning continues (Orange line). This indicates that deep learner recalls the knowledge learned in that early tasks, which makes current function more trustworthy. \textbf{(ii.)} Compared to other tasks whose corresponding matching functions are not adopted, the task whose corresponding matching function is adopted is significantly less forgotten (Orange column in Figure \ref{fig:1st_single} (b, d, f, h, j)). This indicates that a function from early tasks without knowledge fading is essential.

In summary, these experiments provide more support for the reliability of the functions from early tasks. 
Building on these, we conjecture that: \emph{the absence of a function at a specific stage makes CL biased against this old task}. In contrast, provided absent functions from the early tasks, a deep learner could recall the knowledge learned.

\subsection*{A.3 More discussion about limitations}
\label{A3}
In the main paper, we discuss the limitations of the proposed method. 
As a complement, we provide detailed discussions and evidence on two method options for reducing computational complexity to ensure the applicability of PLwF.

\textbf{[Method A] Fast Knowledge Distillation~\cite{shen2021afast} (FKD).} In the following paragraphs, we aim to investigate a possible approach to boost the training efficiency of PLwF with the assistance of \textit{FKD}. 

PLwF adopts \textit{vanilla KD}~\cite{DBLP:journals/corr/HintonVD15} as a default method to transfer knowledge from previous functions to the current model. The main drawback of a vanilla KD framework is that it consumes the majority of the computational overhead on forwarding through the giant teacher model. To be more specific, the parameters of the teacher model is frozen, making repetitive forwarding on the teacher model redundant in training. While FKD, to some extent, solve the problem. FKD generates one probability vector as the soft label for each training image, then reuse them circularly for different training epochs. Efficiently reduce repetitive forward computations to speed up \textbf{2$\sim$5× without compromising accuracy}. 
For example, Table~\ref{tab:FKD} reveals that by employing the same hyper-parameters and teacher network (Resnet-50), FKD~\cite{shen2021afast} achieves similar results to a baseline KD method~\cite{DBLP:journals/corr/abs-2009-08453} while greatly accelerating the training. (Results are borrowed from \cite{shen2021afast}.)



\begin{table}[h]
\small
\centering
\caption{Results on FKD with ResNet-50. $\heartsuit$ represents the training using \textit{cosine lr} and 1.5x epochs.}
\label{speed_fkd}
\begin{tabular}{@{}ccccc@{}}
\toprule
Method   & Network & Top-1 & Top-5 & Speed-up      \\ \midrule
Baseline KD~\cite{DBLP:journals/corr/abs-2009-08453}   & ResNet-50  & 80.67 & 95.09 & 1.0           \\
w/ FKD    & ResNet-50  & 80.70 & 95.13 & \textbf{0.3x} \\
w/ $\heartsuit$ FKD & ResNet-50  & \textbf{80.91} & \textbf{95.39} & \textbf{0.5x} \\ \bottomrule
\end{tabular}
\label{tab:FKD}
\end{table}

With regard to the combination of PLwF and FKD, since PLwF densely introduces previous functions into the current stage of incremental learning,
we could largely speed up the training due to the large number of functions (\textit{i.e.} teacher models). We expect that contributions will follow.


\textbf{[Method B] Prune, then Distill~\cite{DBLP:journals/corr/abs-2109-14960}.} 
Another intuitive solution to reduce computational overhead is to boost the inference speed of model functions. \textbf{Prune, then Distill} provides a good support for this solution. We analyze the feasibility in terms of both inference speed and accuracy. 

\textit{Inference speed \& Accuracy:} 
The desirable result of Method B is to reduce the computational overhead while ensuring uncompromised accuracy. 
In Table~\ref{pruning_acc},  FKD~\cite{DBLP:journals/corr/abs-2109-14960} has demonstrated pruning models would not negatively impact the inference speed and accuracy.
Furthermore, FKD~\cite{DBLP:journals/corr/abs-2109-14960} even helps student models achieve better performance. (Results in Table~\ref{pruning_acc} are borrowed from ~\cite{DBLP:journals/corr/abs-2109-14960}.

\begin{table}[h]
\centering
\footnotesize
\caption{The effect of Prune, then Distill. Teacher
``None" indicates the student is trained without a teacher, while the pruning ratio ``None" means the distillation from the unpruned teacher.}
\label{pruning_acc}
\begin{tabular}{@{}ccc|cc@{}}
\toprule
Teacher  & Pruning ratio & Accuracy & Student  & Accuracy              \\ \midrule
None     & -             & -        & ResNet18 & 57.75 ± 0.24 \\ \midrule
ResNet18 & None          & 57.75    & ResNet18 & 57.97 ± 0.10 \\
ResNet18 & 36\%          & 57.66    & ResNet18 & 59.39 ± 0.21 \\
ResNet18 & 59\%          & 57.58    & ResNet18 & 58.99 ± 0.26 \\
ResNet18 & 79\%          & 57.32    & ResNet18 & 59.33 ± 0.18 \\ \bottomrule
\end{tabular}
\end{table}

In addition, we could utilize more superior pruning methods to boost the performance.
For example, \cite{DBLP:journals/corr/abs-2202-02643} proposed that a randomly pruned subnetwork of ResNet can outperform a dense ResNet.
To sum up, in our case, the problem of high computational overhead would be decently resolved by \textbf{Prune, then Distill.}

\subsection*{A.4 Discussion the relaxed PLwF}
\label{A4}
\begin{figure*}
\begin{minipage}[c]{0.5\textwidth}
\centering
\includegraphics[width=2.2in]{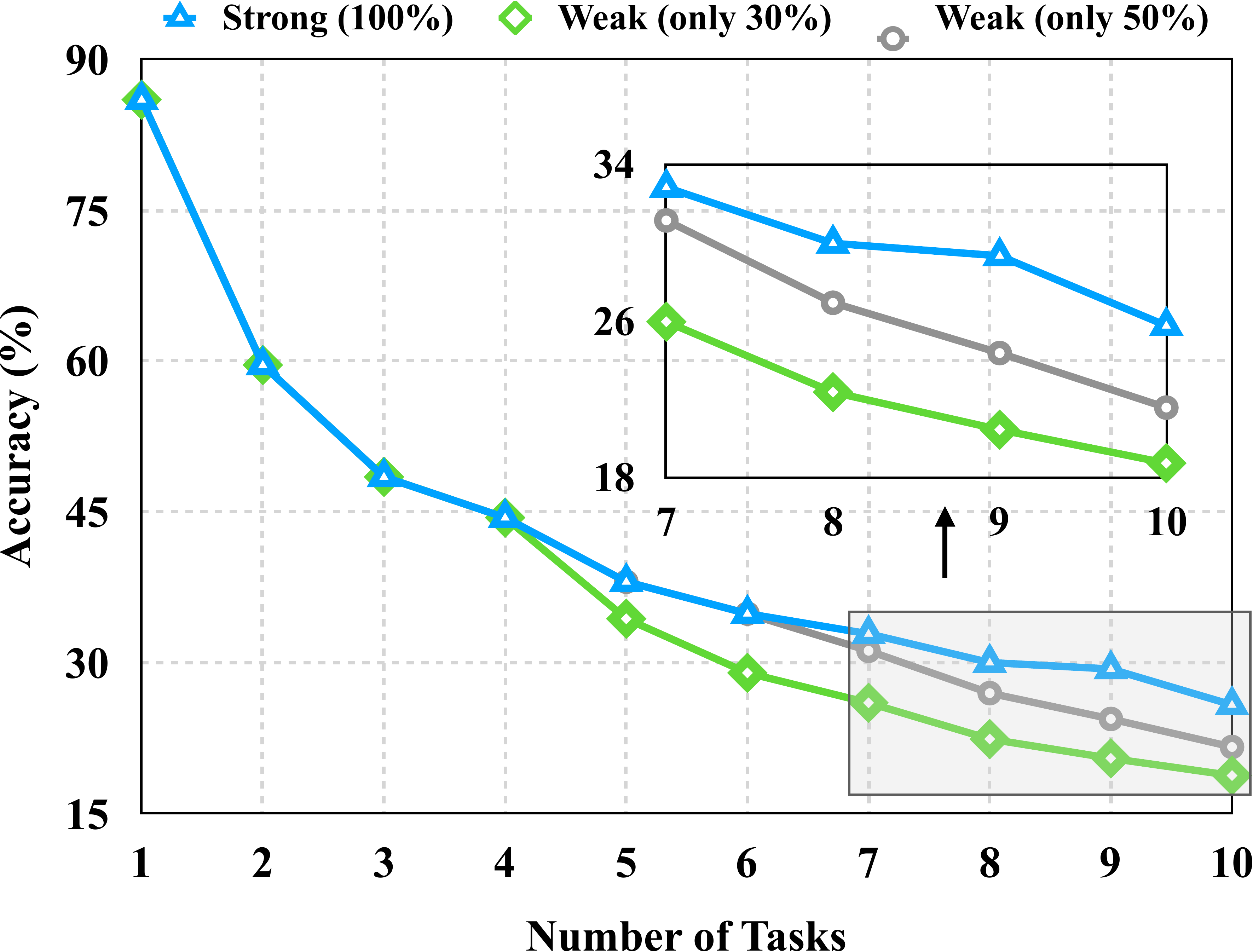}
\caption{The first 30\%/50\% functions from $\mathcal{F}$}
\label{fig:sup_relax_only}
\end{minipage}
\begin{minipage}[c]{0.5\textwidth}
\centering
\vspace{-16mm}
\captionof{table}{The first 30\%/50\% functions from $\mathcal{F}$}
\label{tab:sup_relax_only}
\begin{tabular}{@{}c|ccc@{}}
\toprule
Equation 4 & Avg           & Last          & Cost  \\ \midrule
Strong (100\%)       & 42.94         & 25.79         & -     \\
Weak (50\%)  & 41.55 (\textcolor{red}{$\downarrow$1.39}) & 21.60 (\textcolor{red}{$\downarrow$4.00}) & \textcolor{forestgreen}{$\downarrow$50\%} \\
Weak (30\%)  & 38.94 (\textcolor{red}{$\downarrow$4.00}) & 18.76 (\textcolor{red}{$\downarrow$7.03}) & \textcolor{forestgreen}{$\downarrow$70\% }\\ \bottomrule
\end{tabular}
\end{minipage}
\end{figure*}

\begin{figure*}
\begin{minipage}[c]{0.5\textwidth}
\centering
\includegraphics[width=2.2in]{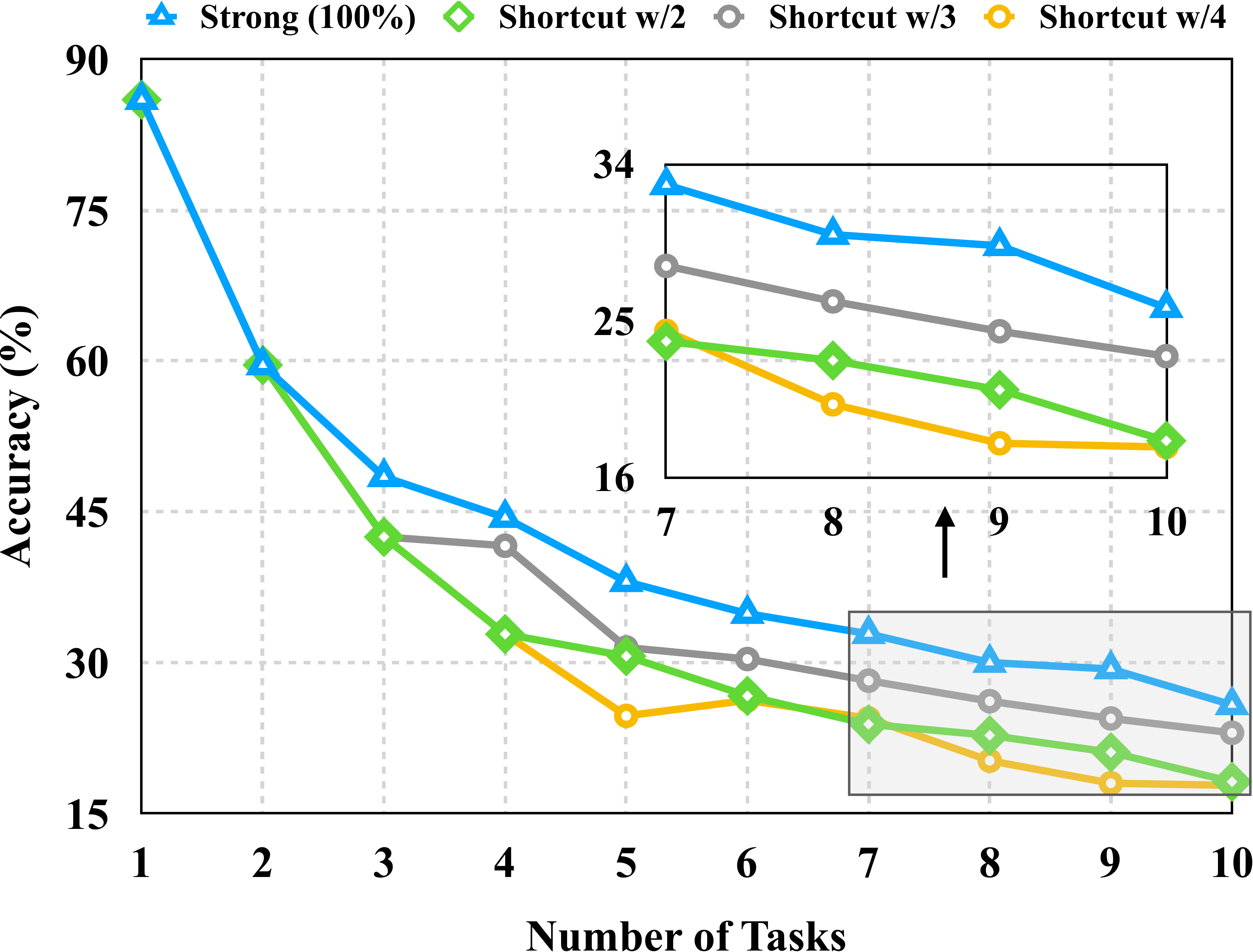}
\caption{The shortcut with interval 2, 3, 4 from $\mathcal{F}$}
\label{fig:sup_relax_shortcut}
\end{minipage}
\begin{minipage}[c]{0.5\textwidth}
\centering
\vspace{-16mm}
\captionof{table}{The shortcut with interval 2, 3, 4 from $\mathcal{F}$}
\label{tab:sup_relax_shortcut}
\begin{tabular}{@{}c|ccc@{}}
\toprule
Equation 4 & Avg           & Last          & Cost  \\ \midrule
Strong (100\%)       & 42.94         & 25.79         & -     \\
Shortcut w/2  & 39.34 (\textcolor{red}{$\downarrow$3.6}) & 23.10 (\textcolor{red}{$\downarrow$2.78}) & \textcolor{forestgreen}{$\downarrow$50\%} \\
Shortcut w/3  & 36.41 (\textcolor{red}{$\downarrow$6.53}) & 18.14 (\textcolor{red}{$\downarrow$7.65}) & \textcolor{forestgreen}{$\downarrow$70\%} \\
Shortcut w/4  & 35.24 (\textcolor{red}{$\downarrow$7.70}) & 17.78 (\textcolor{red}{$\downarrow$8.01}) & \textcolor{forestgreen}{$\downarrow$80\% }\\ \bottomrule
\end{tabular}
\end{minipage}
\end{figure*}

\begin{figure*}
\begin{minipage}[c]{0.5\textwidth}
\centering
\includegraphics[width=2.2in]{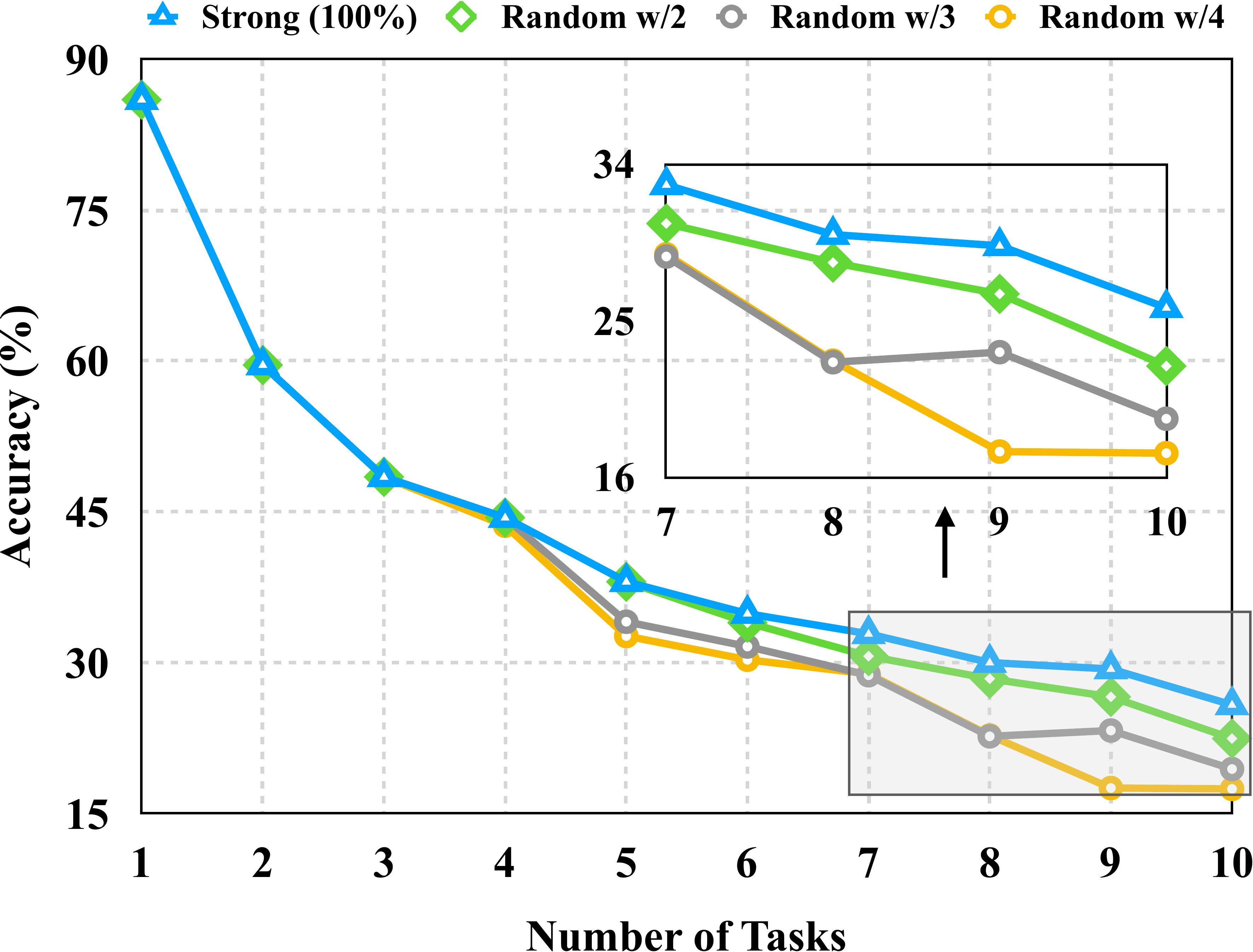}
\caption{Random selection of 2, 3, 4 functions from $\mathcal{F}$}
\label{fig:sup_relax_random}
\end{minipage}
\begin{minipage}[c]{0.5\textwidth}
\centering
\vspace{-16mm}
\captionof{table}{Random selection of 2, 3, 4 functions from $\mathcal{F}$}
\label{tab:sup_relax_random}
\begin{tabular}{@{}c|ccc@{}}
\toprule
Equation 4 & Avg           & Last          & Cost  \\ \midrule
Strong (100\%)       & 42.94         & 25.79         & -     \\
Random w/4  & 41.86 (\textcolor{red}{$\downarrow$1.08}) & 22.44 (\textcolor{red}{$\downarrow$3.35}) & \textcolor{forestgreen}{$\downarrow$60\%} \\
Random w/3  & 39.82 (\textcolor{red}{$\downarrow$3.12}) & 19.41 (\textcolor{red}{$\downarrow$6.38}) & \textcolor{forestgreen}{$\downarrow$70\%} \\
Random w/2  & 38.72 (\textcolor{red}{$\downarrow$4.22}) & 17.43 (\textcolor{red}{$\downarrow$8.36}) & \textcolor{forestgreen}{$\downarrow$80\% }\\ \bottomrule
\end{tabular}
\end{minipage}
\end{figure*}

We provide several relaxed versions of PLwF by adopting a subset of previous functions $\Tilde{\mathcal{F}}$: \textbf{(i)} \textit{Scheme 1} (Figure~\ref{fig:sup_relax_sum} and Table~\ref{tab:sup_relax_sum}): Use the ($t-1$)th function and first 30\%/50\% functions from earlier stages in Equation 4. \textbf{(ii)} \textit{Scheme 2}: (Figure~\ref{fig:sup_relax_only} and Table~\ref{tab:sup_relax_only}): Solely use first 30\%/50\% functions from earlier stages in Equation 4. \textbf{(iii)} \textit{Scheme 3}: (Figure~\ref{fig:sup_relax_shortcut} and Table~\ref{tab:sup_relax_shortcut}): Use functions sampled with a stage interval of 2, 3 and 4 in Equation 4. \textbf{(iv)} \textit{Scheme 4}: (Figure~\ref{fig:sup_relax_random} and Table~\ref{tab:sup_relax_random}): Use 2, 3 and 4 randomly selected functions in Equation 4.

All of the above schemes substantially reduce the computational overhead, which enhances the application of the proposed method. 
Among them, 
\emph{Scheme 1} considers that previous functions have different influences on the optimization of the current task, for example, more recent functions suffer from slighter forgetting, and earlier functions suffer from severer forgetting.
Therefore, we could sample more recent functions to achieve decent performance.
In contrast, 
\emph{Scheme 2} abandons more recent functions and thus, appears a less cost-effective strategy. 
While \emph{Scheme 3} and \emph{Scheme 4} are affected by different sampling strategies, thus a good sampling strategy needs to be considered in practice. 
In summary, by proper sampling functions (relaxed PLwF), we could maintain the computational overhead of PLwF to be constant, while ensuring decent performance.
Such a scheme potentially benefits the application of PLwF.

\subsection*{A.5 GPU occupation of PLwF and Relaxed PLwF}
\label{A5}
\begin{figure}[h]
    \centering
   \begin{subfigure}{0.45\textwidth}
       \centering
       \includegraphics[width=1.8in]{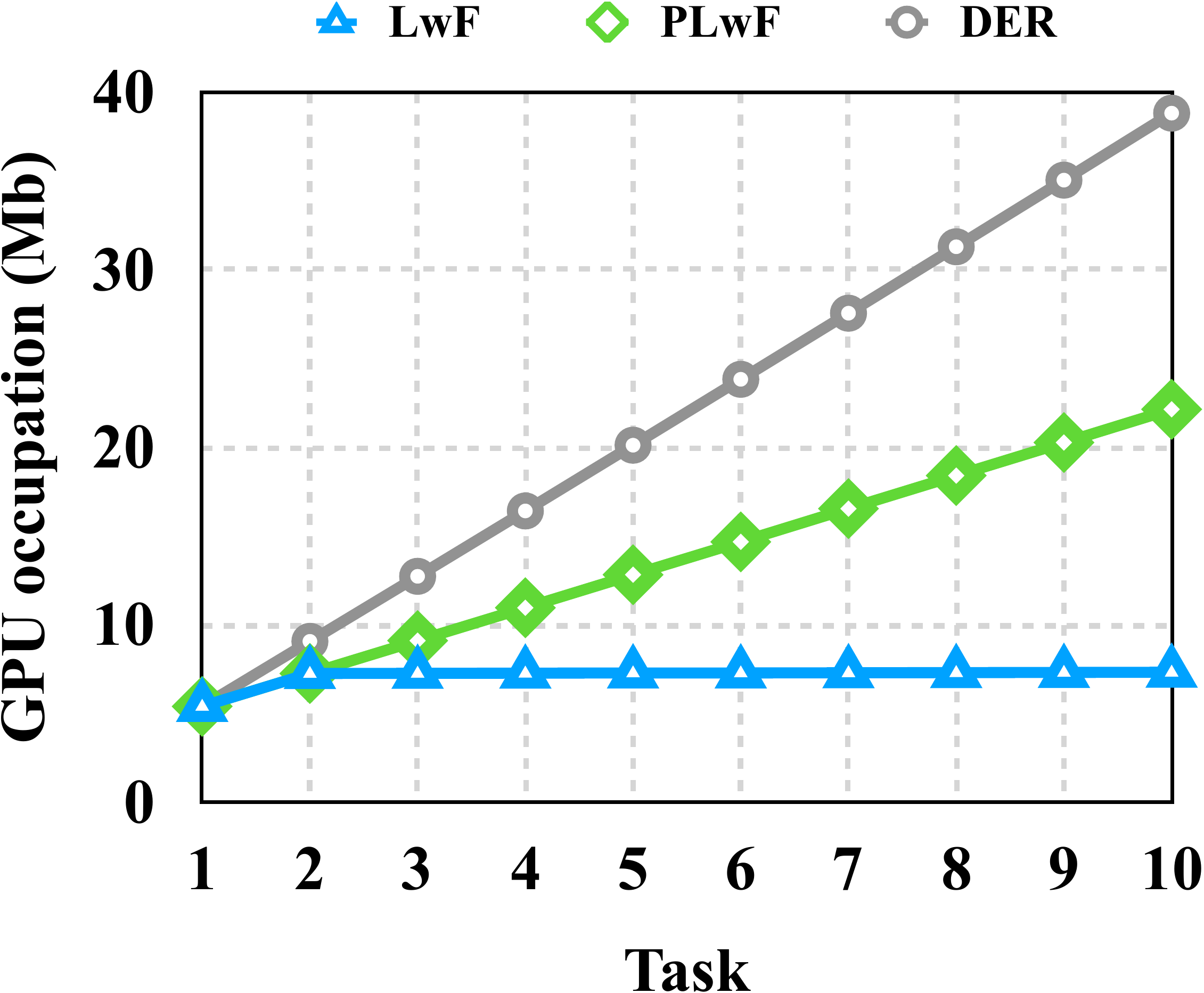}
        \caption{PLwF}
        \label{fig:gpu_non_constant}
    \end{subfigure}
   \begin{subfigure}{0.45\textwidth}
       \centering
       \includegraphics[width=1.8in]{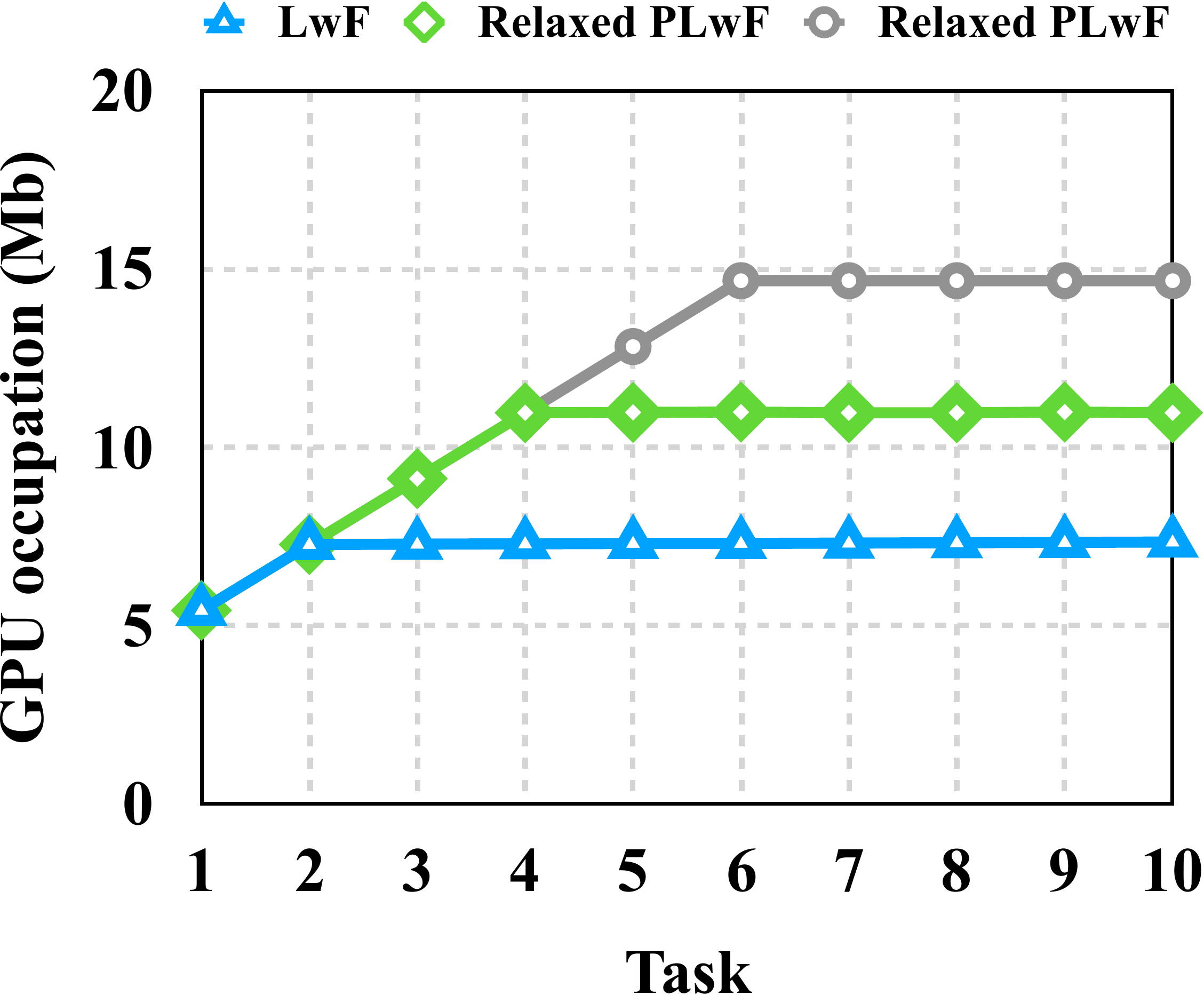}
        \caption{Relaxed PLwF}
        \label{fig:gpu_constant}
    \end{subfigure}
\caption{GPU occupation of PLwF and Relaxed PLwF on Split CIFAR-100. (b) Relaxed PLwF (green) indicates the GPU occupancy of Scheme 1 (30\%) and Relaxed PLwF (gray) indicates the GPU occupancy of Scheme 1 (50\%).}
\label{fig:gpu_occu} 
\vspace{-6mm}
\end{figure}

\begin{figure*}
\begin{minipage}[c]{0.5\textwidth}
\centering
\includegraphics[width=2.2in]{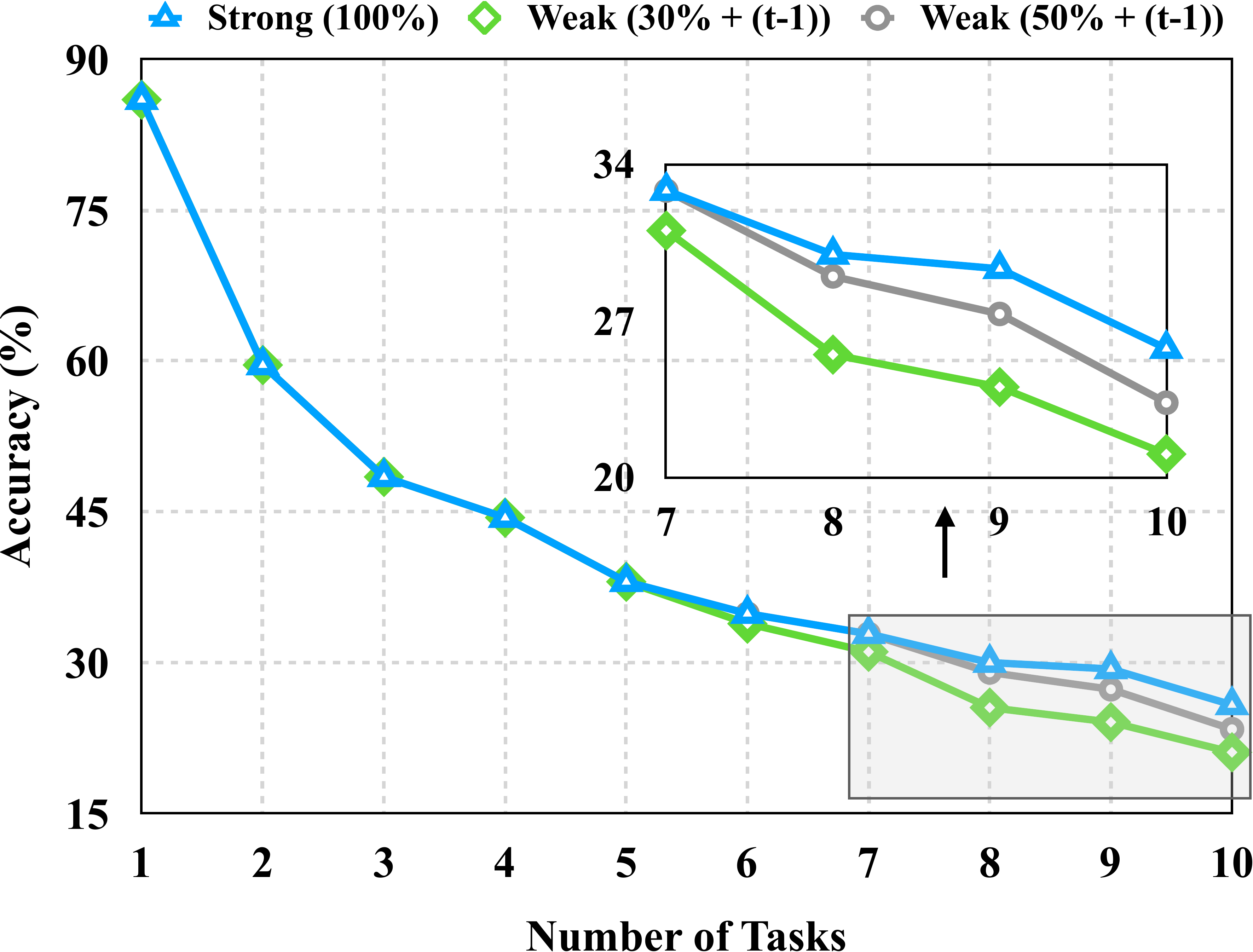}
\caption{The first 30\%/50\% and ($t-1$)th functions from $\mathcal{F}$}
\label{fig:sup_relax_sum}
\end{minipage}
\begin{minipage}[c]{0.5\textwidth}
\centering
\vspace{-16mm}
\captionof{table}{The first 30\%/50\% and ($t-1$)th functions from $\mathcal{F}$}
\label{tab:sup_relax_sum}
\begin{tabular}{@{}c|ccc@{}}
\toprule
Equation 4 & Avg           & Last          & Cost  \\ \midrule
Strong (100\%)       & 42.94         & 25.79         & -     \\
Weak (50\% + ($t-1$))  & 42.40 (\textcolor{red}{$\downarrow$0.54}) & 23.37 (\textcolor{red}{$\downarrow$2.42}) & \textcolor{forestgreen}{$\downarrow$40\%} \\
Weak (30\% + ($t-1$))  & 41.21 (\textcolor{red}{$\downarrow$1.73}) & 21.07 (\textcolor{red}{$\downarrow$4.72}) & \textcolor{forestgreen}{$\downarrow$60\% }\\ \bottomrule
\end{tabular}
\end{minipage}
\end{figure*}

In this subsection, we present the GPU occupation of PLwF and Relaxed PLwF. In Figure~\ref{fig:gpu_occu}a, we specifically compare to dynamically expandable representation method (DER~\cite{DBLP:conf/cvpr/YanX021}). We observe that the GPU occupancy and growth rate are significantly lower than DER, even though PLwF densely introduced the previous functions. Moreover, as shown in Figure~\ref{fig:gpu_occu}b, we present the detailed GPU utilization for scheme 1 (Figure~\ref{fig:sup_relax_sum} and Table~\ref{tab:sup_relax_sum} ). Figure~\ref{fig:gpu_occu}b reveals that we could maintain the computational overhead of PLwF to be constant.

\subsection*{A.6 More details about main results}
\label{A6}
\begin{figure}
    \centering
   \begin{subfigure}{0.45\textwidth}
       \centering
       \includegraphics[width=1.8in]{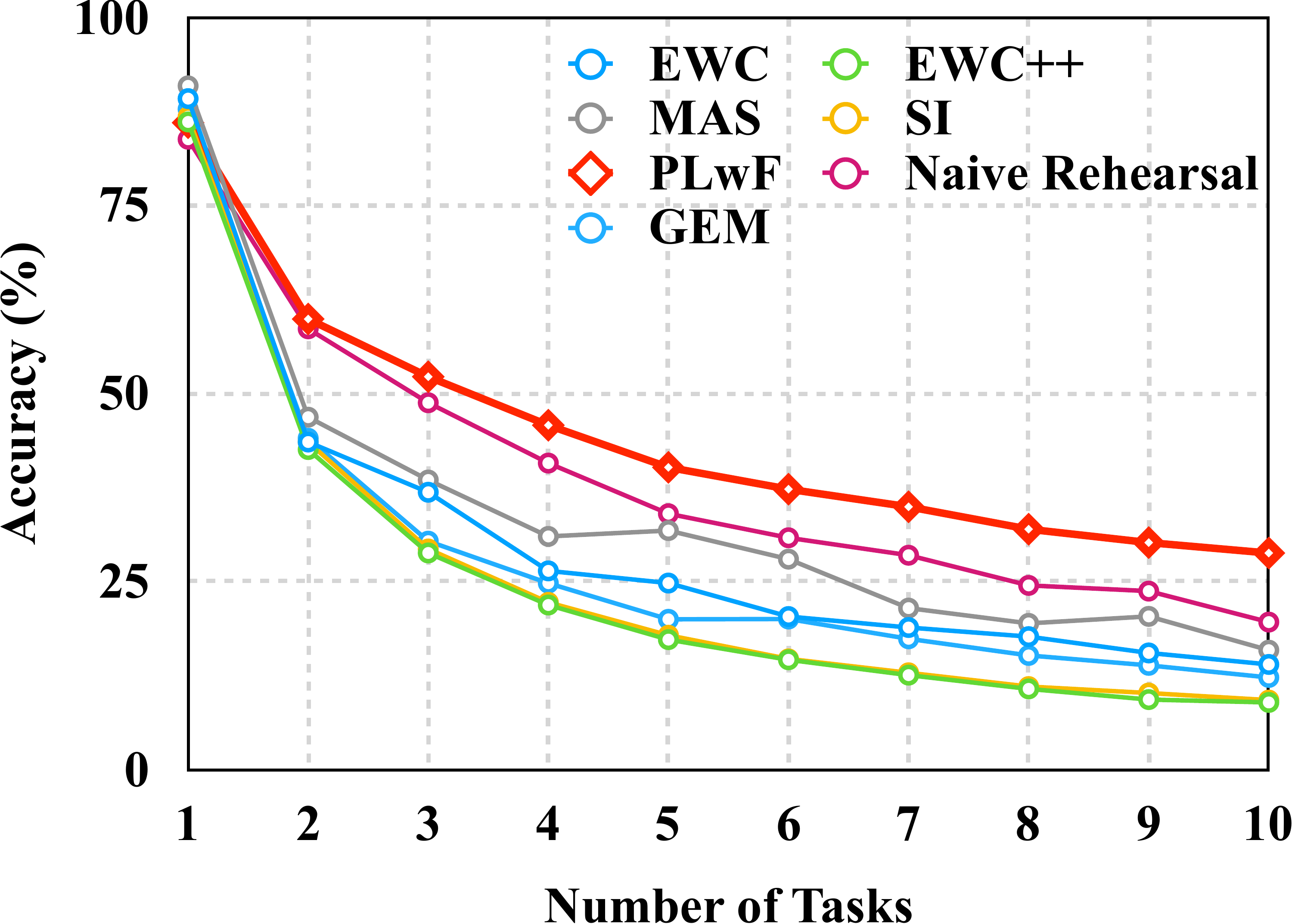}
        \caption{10-Split CIFAR-100.}
        \label{fig:act_10}
    \end{subfigure}
   \begin{subfigure}{0.45\textwidth}
       \centering
       \includegraphics[width=1.8in]{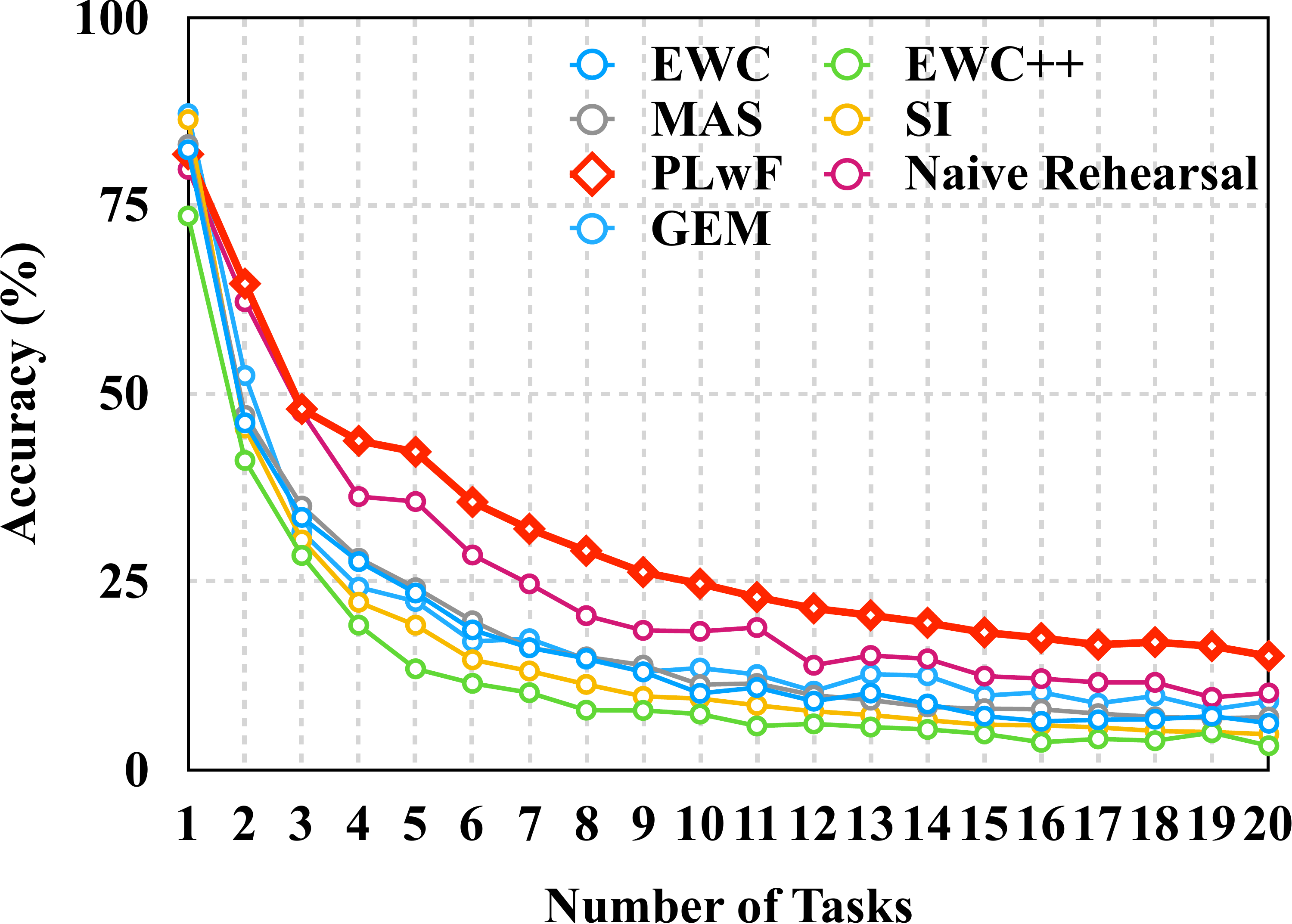}
        \caption{20-Split CIFAR-100.}
        \label{fig:act_20}
    \end{subfigure}
\caption{Incremental learning with 5 and 10 classes at a time on Split CIFAR-100.}
\label{fig:act_adv}
\vspace{-6mm}
\end{figure}

To better understand the actual advantage of the proposed method, we present the average results over several stages of learning on 10-Split CIFAR-100 and 20-Split CIFAT-100. As shown in Figure \ref{fig:act_adv}, PLwF consistently outperforms all the methods by significant margins across all settings. 

\subsection*{A.7 More discussion about the generality of credit assignment}
\label{A7}

In \textit{Generality of Credit Assignment} of the main paper, we present the results of adding credit assignment on EWC~\cite{kirkpatrick2017overcoming}, and brings tangible improvements. Further, we observe the impact of credit assignment on iCaRL~\cite{rebuffi2017icarl}. In Table \ref{table_icarl}, we present the results of adding credit assignment on iCaRL, which improves the performance of iCaRL by 1.14\% on Avg and 2.08\% on Last. In Table \ref{table_icarl_1st}, we present the performance variation of iCaRL on the first task, which shows forgetting is further controlled. 
This is a valuable phenomenon since the tug-of-way dynamics commonly exists in previous methods, and the credit assignment can recreate this conflict. 
We expect the credit assignment will inspire future work to focus on this problem and contributions will follow.

\begin{table}[h]
\centering
\caption{Credit assignment on iCaRL.}
\label{table_icarl}
\begin{tabular}{@{}ccc@{}}
\toprule
\multirow{2}{*}{Method} & \multicolumn{2}{c}{5-Split CIFAR10} \\ \cmidrule(l){2-3} 
                        & Avg           & Last          \\ \midrule
iCaRL                   & 79.21         & 69.00         \\
w/ Credit                & \textbf{80.35} (\textcolor{forestgreen}{+1.14})   & \textbf{71.08} (\textcolor{forestgreen}{+2.08})   \\ \bottomrule
\end{tabular}
\end{table}

\begin{table*}
\centering
\footnotesize
\caption{Evaluation results (\%) of order-agnostic behavior on different seeds.}
\label{order_seed}
\begin{tabular}{@{}cc|ccccccccccc|c@{}}
\toprule
Random & Method & EWC   & EWC++ & MAS   & SI    & GEM & A-GEM & ER    & FDR   & GSS   & HAL   & PODNET & Ours           \\ \midrule
\multirow{2}{*}{Order 1}     & Avg    & 29.82 & 23.68 & 33.82  & 26.08 & -   & 25.93 & 40.89 & 40.20 & 33.85 & 28.63 & 46.95  & \textbf{49.37} \\
     & Last    & 13.72 & 7.18  & 15.97 & 9.24  & -   & 9.68  & 21.27 & 22.45 & 13.50 & 12.12 & 22.25  & \textbf{29.57} \\ \midrule
\multirow{2}{*}{Order 2}      & Avg    & 31.10 & 23.85 & 32.38 & 25.90 & -   & 24.89 & 39.88 & 40.99 & 33.10 & 30.73 & 44.24  & \textbf{47.14} \\
     & Last    & 16.88 & 6.69  & 14.80 & 9.32  & -   & 9.29  & 20.02 & 23.55 & 12.88 & 13.05 & 19.80  & \textbf{27.20} \\ \midrule
\multirow{2}{*}{Order 3}      & Avg    & 31.44 & 23.30 & 31.32 & 25.70 & -   & 25.87 & 39.72 & 41.38 & 32.08 & 30.00 & 45.50  & \textbf{45.54} \\
     & Last    & 15.85 & 5.54  & 14.34 & 9.37  & -   & 8.89  & 19.62 & 25.18 & 12.53 & 13.19 & 20.83  & \textbf{25.50} \\ \bottomrule
\end{tabular}
\end{table*}

\begin{table*}
\centering
\caption{Performance variation (\%) of iCaRL on 1st task when trained over 1 task to 5 tasks on 5-Split CIFAR-10.}
\label{table_icarl_1st}
\vspace{-2mm}
\begin{tabular}{@{}lccccc@{}}
\toprule
Method   & Task 1  & Task 2    & Task 3      & Task 4    & Task 5     \\ \midrule
iCaRL    & 97.60     & 85.90       & 84.80         & 82.90       & 54.75       \\
w/ Credit & 97.60 (+0) & \textbf{87.30} (\textcolor{forestgreen}{+1.4}) & \textbf{86.05} (\textcolor{forestgreen}{+1.25}) & \textbf{83.50} (\textcolor{forestgreen}{+0.6}) & \textbf{58.95} (\textcolor{forestgreen}{+4.2}) \\ \bottomrule
\end{tabular}
\end{table*}

\subsection*{A.8 Details about the order-agnostic behaviour}
\label{A8}

\begin{figure}
    \centering
  \begin{subfigure}{0.3\textwidth}
      \includegraphics[width=1.8in]{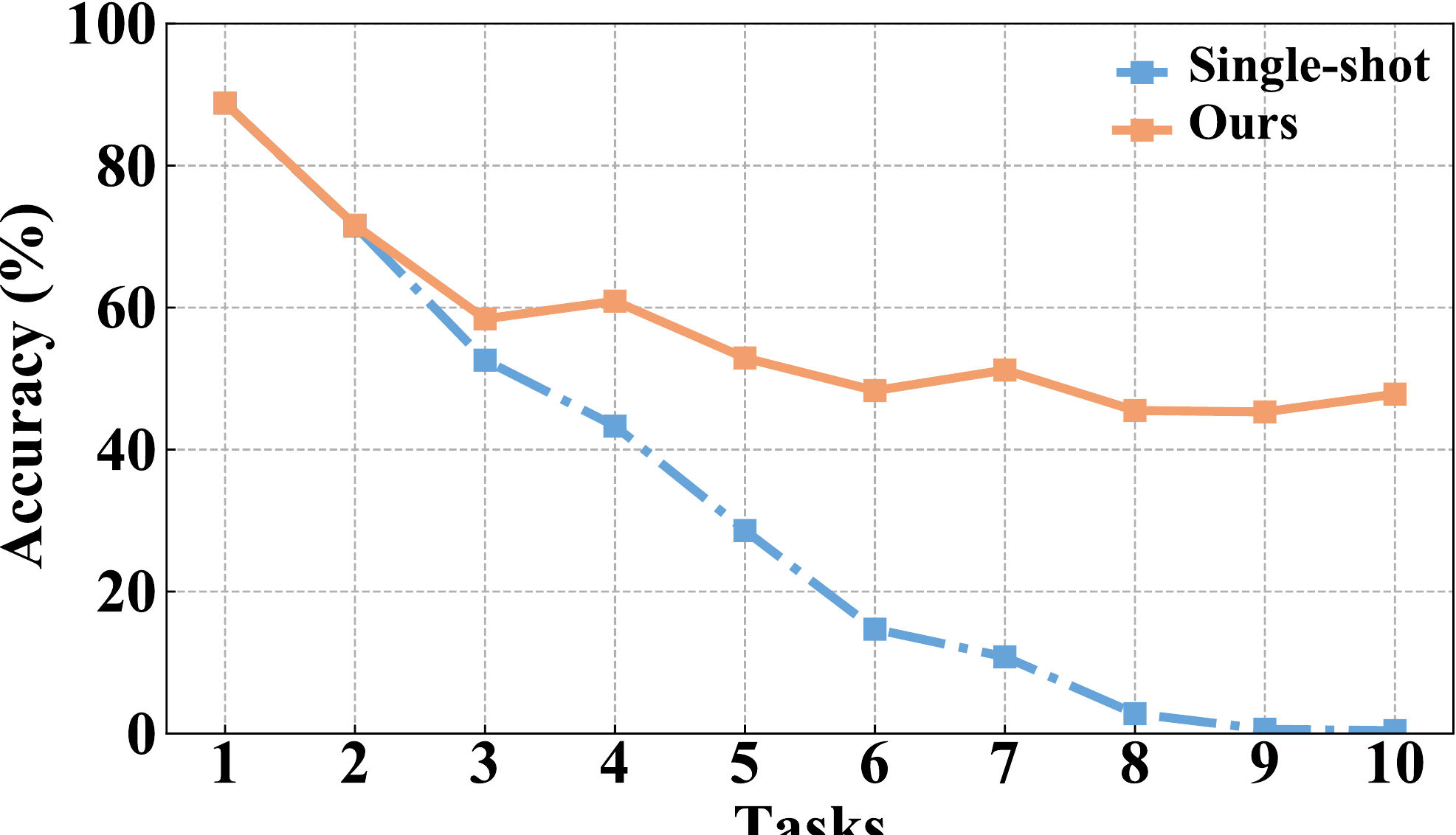}
        \caption{Random order 1}
        \label{fig:2000_seed_100}
    \end{subfigure}
  \begin{subfigure}{0.3\textwidth}
      \includegraphics[width=1.8in]{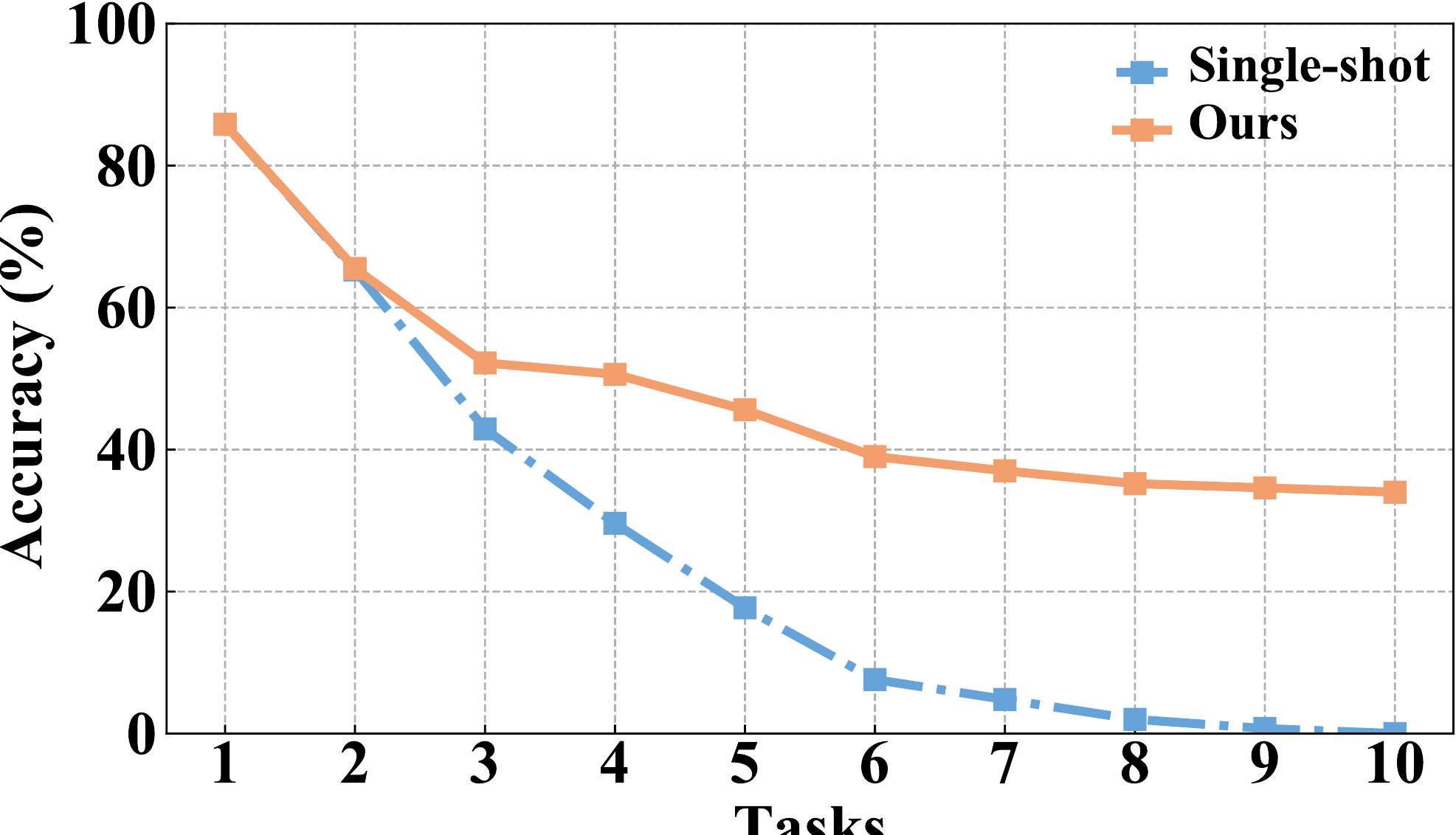}
        \caption{Random order 2}
        \label{fig:2019_seed_100}
    \end{subfigure}
  \begin{subfigure}{0.3\textwidth}
      \includegraphics[width=1.8in]{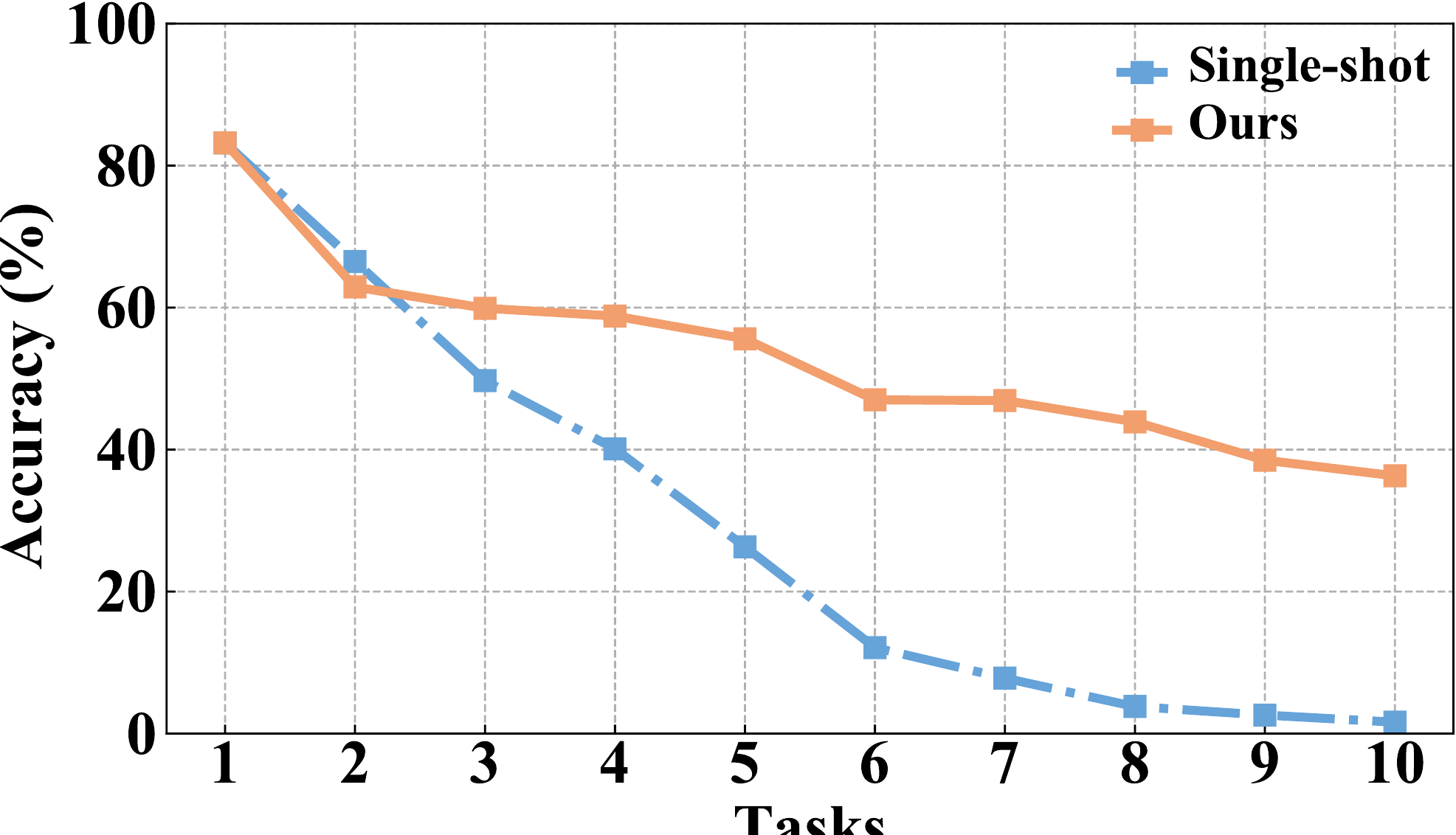}
        \caption{Random order 3}
        \label{fig:1949_seed_100}
    \end{subfigure}
\caption{Performance variation of the first task on different seeds when trained over 10 tasks on Split CIFAR-100.}
\label{fig:order_seed}  
\vspace{-7mm}
\end{figure}

To observe the influence of different class orders, we set different random seeds to split CIFAR-100 dataset, which yields a more dynamic and agnostic class distributions. 
As shown in Table \ref{order_seed}, the proposed method outperforms other methods using three different random seeds.
This further validates the robustness of our method. 
Moreover, as shown in Figure \ref{fig:order_seed}(a, b, c), in this case, our method remains effective in controlling forgetting (Orange line). 
These observations provide additional support for our method.

\subsection*{A.9 Trendy directions in raw-data-free methods}
\label{A9}
In some applications, storing raw data is not feasible due to privacy and security concerns, and this requires CL methods to maintain reasonable performance without any raw data. The regularization-based method is one line of this approach.

PLwF points out that the regularization-based method is limited by the incomplete learnable space due to the fading function credibility. 
In contrast, the rehearsal-based methods suffers from a slighter credibility crisis of the function due to direct exposure to raw data. 
PLwF provides a potential direction for regularization-based methods: 
Finding a knowledge container with the most precise and fresh knowledge over the previous tasks.
Although rehearsal-based methods generally achieve better performance than regularization-based methods, PLwF paves a way for regularization-based methods (raw-data-free methods) to thrive again.

\end{document}